\documentclass[runningheads]{llncs}

\usepackage{eccv}

\usepackage{eccvabbrv}

\usepackage{graphicx}
\usepackage{booktabs}
\usepackage{standalone}
\usepackage{amssymb}
\usepackage{graphicx}
\usepackage{multirow}
\usepackage{siunitx}
\usepackage{pgfplots}
\pgfplotsset{compat=1.18}
\usepackage{amsmath}
\usepackage{makecell}
\usepackage{pifont}
\usepackage[symbol]{footmisc}
\usepgfplotslibrary{groupplots}

\usepackage[accsupp]{axessibility}  

\usepackage{hyperref}

\usepackage{orcidlink}

\usepackage{tikz}
\usepackage{multirow}
\usepackage{multicol}
\usetikzlibrary{
  backgrounds,
  calc,
  positioning,
  quotes,
  patterns,
  spy,
  arrows.meta,
  decorations.pathmorphing,
  shapes.geometric,
  fit
}

\definecolor{tu1}{rgb}{1.0000, 0.8039, 0.0000}
\definecolor{tu11}{rgb}{1.0000, 0.8627, 0.3020}
\definecolor{tu12}{rgb}{1.0000, 0.9020, 0.4980}
\definecolor{tu13}{rgb}{1.0000, 0.9412, 0.6980}
\definecolor{tu14}{rgb}{1.0000, 0.9608, 0.8000}

\definecolor{tu2}{rgb}{0.9804, 0.4314, 0.0000}
\definecolor{tu21}{rgb}{0.9882, 0.6039, 0.3020}
\definecolor{tu22}{rgb}{0.9882, 0.7137, 0.4980}
\definecolor{tu23}{rgb}{0.9922, 0.8275, 0.6980}
\definecolor{tu24}{rgb}{0.9961, 0.8863, 0.8000}

\definecolor{tu3}{rgb}{0.6902, 0.0000, 0.2745}
\definecolor{tu31}{rgb}{0.7529, 0.2000, 0.4196}
\definecolor{tu32}{rgb}{0.8431, 0.4980, 0.6353}
\definecolor{tu33}{rgb}{0.9216, 0.7490, 0.8196}
\definecolor{tu34}{rgb}{0.9529, 0.8510, 0.8902}

\definecolor{tu4}{rgb}{0.4863, 0.8039, 0.9020}
\definecolor{tu41}{rgb}{0.6431, 0.8627, 0.9333}
\definecolor{tu42}{rgb}{0.7412, 0.9020, 0.9490}
\definecolor{tu43}{rgb}{0.8431, 0.9412, 0.9686}
\definecolor{tu44}{rgb}{0.8980, 0.9608, 0.9804}

\definecolor{tu5}{rgb}{0.0000, 0.5020, 0.7059}
\definecolor{tu51}{rgb}{0.3020, 0.6510, 0.7961}
\definecolor{tu52}{rgb}{0.5490, 0.7765, 0.8667}
\definecolor{tu53}{rgb}{0.7490, 0.8745, 0.9255}
\definecolor{tu54}{rgb}{0.8510, 0.9255, 0.9569}

\definecolor{tu6}{rgb}{0.0000, 0.3255, 0.4549}
\definecolor{tu61}{rgb}{0.2510, 0.4941, 0.5922}
\definecolor{tu62}{rgb}{0.5490, 0.6941, 0.7529}
\definecolor{tu63}{rgb}{0.7490, 0.8314, 0.8627}
\definecolor{tu64}{rgb}{0.8510, 0.8980, 0.9176}

\definecolor{tu7}{rgb}{0.0314, 0.0314, 0.0314}
\definecolor{tu71}{rgb}{0.3725, 0.3725, 0.3725}
\definecolor{tu72}{rgb}{0.5882, 0.5882, 0.5882}
\definecolor{tu73}{rgb}{0.7529, 0.7529, 0.7529}
\definecolor{tu74}{rgb}{0.8667, 0.8667, 0.8667}

\definecolor{tu8}{rgb}{0.7765, 0.9333, 0.0000}
\definecolor{tu81}{rgb}{0.8431, 0.9529, 0.3020}
\definecolor{tu82}{rgb}{0.8863, 0.9647, 0.4980}
\definecolor{tu83}{rgb}{0.9333, 0.9804, 0.6980}
\definecolor{tu84}{rgb}{0.9569, 0.9882, 0.8000}

\definecolor{tu9}{rgb}{0.5373, 0.6431, 0.0000}
\definecolor{tu91}{rgb}{0.6784, 0.7490, 0.3020}
\definecolor{tu92}{rgb}{0.7686, 0.8196, 0.4980}
\definecolor{tu93}{rgb}{0.8588, 0.8941, 0.6980}
\definecolor{tu94}{rgb}{0.9059, 0.9294, 0.8000}

\definecolor{tu10}{rgb}{0.0000, 0.4431, 0.3373}
\definecolor{tu101}{rgb}{0.3020, 0.6118, 0.5373}
\definecolor{tu102}{rgb}{0.5490, 0.7490, 0.7020}
\definecolor{tu103}{rgb}{0.7490, 0.8588, 0.8353}
\definecolor{tu104}{rgb}{0.8549, 0.9176, 0.9059}

\definecolor{tu110}{rgb}{0.8000, 0.0000, 0.6000}
\definecolor{tu111}{rgb}{0.8706, 0.3490, 0.7412}
\definecolor{tu112}{rgb}{0.9216, 0.6000, 0.8392}
\definecolor{tu113}{rgb}{0.9608, 0.8000, 0.9216}
\definecolor{tu114}{rgb}{0.9804, 0.8980, 0.9608}

\definecolor{tu120}{rgb}{0.4627, 0.0000, 0.4627}
\definecolor{tu121}{rgb}{0.5961, 0.2510, 0.5961}
\definecolor{tu122}{rgb}{0.7294, 0.4980, 0.7294}
\definecolor{tu123}{rgb}{0.8392, 0.6980, 0.8392}
\definecolor{tu124}{rgb}{0.9216, 0.8510, 0.9216}

\definecolor{tu130}{rgb}{0.4627, 0.0000, 0.3294}
\definecolor{tu131}{rgb}{0.6118, 0.3020, 0.5333}
\definecolor{tu132}{rgb}{0.7569, 0.5490, 0.6980}
\definecolor{tu133}{rgb}{0.8667, 0.7490, 0.8314}
\definecolor{tu134}{rgb}{0.9216, 0.8510, 0.9020}

\definecolor{tu140}{rgb}{0.7019, 1, 0.7019}
\definecolor{tu141}{rgb}{0.7216, 1, 0.713}
\definecolor{tu142}{rgb}{0.8039, 1, 0.7725}
\definecolor{tu143}{rgb}{0.9098, 0.9921, 0.8470}

\definecolor{tu151}{rgb}{0.1607, 0.1960, 0.1529}
\definecolor{tu152}{rgb}{0.4549, 0.4941, 0.4235}
\definecolor{tu153}{rgb}{0.70196, 0.6941, 0.6392}
\newcommand{\PROB}[0]{{\mathrm P}}         

\begin{document}

\title{Adversarial Attack and Disturbance Detection \\ by Hadamard-Coded Output Representations \\ for Object Detection and Semantic Segmentation} 

\titlerunning{Perturbation Detection by Hadamard-Coded Output Representations}

\author{Lucas Görnhardt\inst{*} \and
Timo Bartels\inst{*} \and
Niklas Schwarz \and
Tim Fingscheidt}

\authorrunning{L.~Görnhardt et al.}

\institute{Technische Universit\"at Braunschweig, Germany\\
\email{\{lucas.goernhardt, timo.bartels, n.schwarz, t.fingscheidt\}@tu-bs.de}\\}

\maketitle
\renewcommand{\thefootnote}{\fnsymbol{footnote}}
\footnotetext[1]{Marks equal contributions}

\begin{abstract}
   Conventional one-hot encodings often yield poorly calibrated models, being overconfident under attack, and letting entropy-based detection algorithms fail. Previous image classification works have demonstrated that Hadamard-coded output representations can improve adversarial robustness. However, attempts to integrate Hadamard codes into semantic segmentation fall far behind state-of-the-art models in mean intersection-over-union performance. Regarding object detection, such output encodings have not yet been investigated at all. Further, no prior art addressed intrinsic codeword inconsistencies or actually exploited intrinsic codeword redundancy. Accordingly, we first derive a novel decoding procedure for Hadamard codewords towards optimal class-wise probabilities, solving the underlying optimization problem by using the projection onto the probability simplex. Second, our optimization delivers a measure of prediction inconsistency. Third, we are the first to show how to exploit these inconsistencies for adversarial attack and disturbance detection. Fourth, we introduce \texttt{HadamardNet}, a framework employing Hadamard codes as output representations for semantic segmentation and object detection models and tasks. We conduct a comprehensive evaluation both on disturbances and adversarial attacks, \textit{achieving state-of-the-art perturbation detection performance for both tasks in only a single detection pass}, while delivering equivalent or close-by reference performance on clean data. Code is available at https://github.com/ifnspaml/HadamardPerturbationDetection.
\end{abstract}

\section{Introduction}
\label{sec:intro}
In safety-critical domains such as autonomous driving, perception models must remain reliable under image perturbations, whether caused by adversarial attacks or natural disturbances. One direction to address this challenge is the introduction of redundancy into neural networks, for example through Hadamard-coded output representations \cite{yang2015deep, NEURIPS2019_cd61a580, 9839178, NIPS2009_67974233, Hoyos2021, Hoyos2024}. Hsu et al.\ \cite{NIPS2009_67974233} introduced Ha\-da\-mard codes for large-scale image classification and reported improved accuracy. Verma et al.\ \cite{NEURIPS2019_cd61a580} demonstrated improved calibration and adversarial robustness, while Hoyos et al.\ \cite{Hoyos2021} confirmed these benefits across multiple attack types. However, these results do not directly transfer to semantic segmentation or object detection. The only Hadamard-based semantic segmentation method by Hoyos et al.\ \cite{Hoyos2024} indicates faster convergence during training but falls far behind state-of-the-art mean intersection-over-union (mIoU) performance. Furthermore, existing work does not exploit the intrinsic redundancy of Hadamard codewords and its possible usage for perturbation detection.

Perturbation detection \cite{Klingner2022a, Hendrycks2017, Smith2018, liu2020energy, Xu2018e, 10.1007/s10207-023-00735-6, maag2024detectingadversarialattackssemantic, Li2020g, Yin2021} remains an open challenge for safety-critical perception systems. A widely used baseline by Hendrycks et al.\ \cite{Hendrycks2017} employs maximum softmax probability, and Smith et al.\ \cite{Smith2018} propose predictive entropy as detection signal. Liu et al.\ \cite{liu2020energy} introduced the energy score which provides a unified metric for out-of-distribution and adversarial detection. Feature squeezing \cite{Xu2018e} compares predictions on original and quantized or smoothed inputs and Ryu et al.\ \cite{10.1007/s10207-023-00735-6} evaluate entropy differences after bit depth reduction, however, the former requires multiple forward passes and the latter degrades under strong perturbations. For semantic segmentation, Maag et al. train a detector based on entropy maps of both benign and FGSM-attacked images\ \cite{maag2024detectingadversarialattackssemantic} and achieve strong results on FGSM-related attacks but generalize poorly to structurally different perturbations such as Metzen attacks\ \cite{Metzen2017}. For object detection, perturbation detection research remains limited \cite{Nguyen2025, Thunuguntla2025}. Existing approaches either depend on model-specific assumptions as proposed by Li et al.\ \cite{Li2020g} who detect perturbations through context inconsistency or they are specifically tailored to a \texttt{Faster R-CNN}~\cite{Ren2015}. Yin et al.\ \cite{Yin2021} propose heavy additional language models exploiting multi-object relationships.

In this work, we leverage the redundancy of Hadamard-coded output representations for perturbation detection. \textit{Existing work provides no rigorous derivation for mapping predicted Hadamard codewords back to valid class probabilities while exploiting redundancy}.  This is why we first propose a novel and fully probabilistic decoding procedure. We construct a linear equation system (LES) linking the unknown class probabilities to the predicted codeword and \textit{introduce an error vector to make the typically inconsistent equation system solvable under probabilistic constraints}. The resulting optimization problem searches for the minimal error vector satisfying these constraints. Second, we show that the solution is a projection onto the probability simplex, which yields a novel decoding procedure for Hadamard codewords, delivering the error vector as an inconsistency measure of the network output. Third, based on this error vector, we propose an adversarial attack and disturbance detector that requires no adversarial training data, achieving state-of-the-art detection performance with a single detection pass with negligible computational overhead and being model- and task-agnostic. Fourth, we are first to provide a comprehensive evaluation of Hadamard output encodings for state-of-the-art semantic segmentation baselines and to extend them to object detection.

\section{Related Works}
\label{sec:relwork}
\subsubsection{Output Representations\normalfont{:}}
\label{subsec:ouputrep}
For semantic segmentation and object detection, one-hot encoding is standard due to its simplicity and performance. Multiple alternatives have been proposed \cite{NIPS2009_67974233, RODRIGUEZ201821, kong1995error, DBLP:journals/corr/cs-AI-9501101, NIPS2013_7cce53cf, akata2013label, weston2010large, NEURIPS2019_cd61a580, 9839178, Hoyos2024, Hoyos2021}, including expert knowledge-based embeddings \cite{NIPS2013_7cce53cf, akata2013label}, learned embeddings \cite{weston2010large, 9839178}, and data-independent embeddings such as error-correcting codes (ECCs) \cite{NIPS2009_67974233, RODRIGUEZ201821, kong1995error, Hoyos2024}. Dietterich et al.\ \cite{DBLP:journals/corr/cs-AI-9501101} introduce ECCs to improve multi-class performance, and Kong et al.\ \cite{kong1995error} show that ECCs reduce bias and variance across training subsets. Within ECCs, Hadamard codes have drawn particular interest \cite{yang2015deep, NEURIPS2019_cd61a580, 9839178, NIPS2009_67974233, Hoyos2021, Hoyos2024, Hoffer2018} due to their constant Hamming distance.
Our work extends this line by mathematically exploiting Hadamard code redundancy to obtain a perturbation-sensitive inconsistency measure, along with a rigorous mathematical derivation.
\subsubsection{Hadamard-Coded Output Representations\normalfont{:}}
\label{subsec:hadrepr}
Hsu et al.\ \cite{NIPS2009_67974233} first introduced Hadamard codes~\cite{Proakis1989} for large-scale image classification, showing improved accuracy and label space efficiency. Yang et al.\ \cite{yang2015deep} reported improved class discriminability in the penultimate layer. Verma et al.\ \cite{NEURIPS2019_cd61a580} evaluated activation functions for Hadamard-coded outputs and found that tanh improves calibration and adversarial robustness. Hoyos et al.\ \cite{Hoyos2021} confirmed improved robustness across FGSM \cite{Goodfellow2015}, PGD \cite{Kurakin2017}, and other attacks. Wan et al.\ \cite{9839178} further increased robustness by learning codewords starting from Hadamard codes. Hoyos et al.\ \cite{Hoyos2024} applied Hadamard codes to semantic segmentation, reporting faster convergence, however, under very short training schedules, substantially lower mIoU than standard baselines, and with inconsistencies between paper and released code. Their decoding uses a matrix multiplication and softmax but does not leverage codeword redundancy. No prior work applies Hadamard codes to object detection.
We address these gaps by, for the first time, deriving a redundancy-aware decoding method, producing both valid class probabilities and an inconsistency signal, and by extending Hadamard-coded outputs to object detection.

\subsubsection{Adversarial Attack Detection\normalfont{:}}
\label{subsec:advAttackDet}
Adversarial attack and disturbance detection mostly relies on uncertainty in the model output. Hendrycks et al.\ \cite{Hendrycks2017} propose an indicator based on maximum softmax probability and Smith et al.\ \cite{Smith2018} an entropy-based indicator. Liu et al.\ \cite{liu2020energy} introduce the energy score for attack detection. Transformation-based approaches include feature squeezing \cite{Xu2018e}, which requires multiple forward passes, and entropy difference methods by Ryu et al.\ \cite{10.1007/s10207-023-00735-6}, which degrades under stronger attacks. For semantic segmentation, Maag et al.\ \cite{maag2024detectingadversarialattackssemantic} trained a detector on FGSM \cite{Goodfellow2015} perturbations using entropy maps but generalize poorly to different attacks such as Metzen \cite{Metzen2017}. For object detection, existing methods are either architecture-specific, such as context inconsistency detection for \texttt{Faster R-CNN} by Li et al.\ \cite{Li2020g} or they employ heavy language models for multi-object reasoning~\cite{Yin2021}.
Our proposed detector does not depend on model-specific assumptions, or heavy modules, but is even task-agnostic by deriving a detection signal directly from the Hadamard-code redundancy.

\newpage
\begingroup
\renewcommand{\thefootnote}{*}
\NoHyper
\footnotetext{Note that $\check{\mathbf{H}} \cdot \check{\mathbf{H}}^\mathsf{T} \neq S\cdot \mathbf{I}$.}%
\endNoHyper
\endgroup
\section{Methods}
\label{sec:method}
Here, we describe how we employ Hadamard output encodings for perception models and utilize the redundancy for attack and disturbance detection.

\subsection{Hadamard Codes for Classification}
\label{subsec:theo}
We propose to utilize ECC-based output representations for perception models using Hadamard codes~\cite{Proakis1989}, the resulting network we denote as \texttt{HadamardNet}.
Due to its favorable mathematical properties, the Hadamard code is particularly well suited for the output encoding of perception models for classification, such as semantic segmentation or object detection. Hadamard codes provide large, equal Hamming distance between codewords, while both encoding and decoding are computationally efficient. A Hadamard code is of length $L = 2^k$ with $k \in \mathbb{N}$. The Hamming distance between each pair of codewords is $2^{k-1}$.
Hadamard codes are generated using Sylvester’s construction \cite{Sylvester01121867} of the bipolar Hadamard matrix $\check{\mathbf{H}}^{(L)}\in \{-1, 1\}^{L\times L}$, which is detailed with an example in Supplement \ref{supp:HadamardConstructionExample}.
We use $S$ columns $\check{\mathbf{H}}^{(L)}_s = (\check{H}^{(L)}_{s,\ell}) \in \{-1,1\}^{L}$ of the Hadamard matrix $\check{\mathbf{H}}^{(L)} = (\check{\mathbf{H}}^{(L)}_s)$ to obtain the unipolar representations $H^{(L)}_{s,\ell}= \frac{1}{2}(\check{H}^{(L)}_{s,\ell} + 1)$ with columns
\begin{equation}
\label{equ:hadamard_codeword_conversion}
\mathbf{H}^{(L)}_s = (H^{(L)}_{s,\ell})  \in \mathbb{B}^{L}
\end{equation}
and column index $s \in \mathcal{S}=\{1,\dots,S\}$ representing one of $S$ classes, $\ell \in \mathcal{L}= \{1, \dots, L\}$ being the (bit) index within the codeword, and $\mathbb{B}=\{0,1\}$. In this work, we will utilize these \textit{column vectors (\ref{equ:hadamard_codeword_conversion}) as target labels} for classification tasks such as semantic segmentation or object detection, which leads to our proposed \texttt{HadamardNet} architecture with a modified output representation. Since $L$ is always a power of two, it is not guaranteed that the codeword length $L$ and $S$ are equal. The codeword length must only satisfy
\begin{equation}
    L \geq S.
\end{equation}
Using (\ref{equ:hadamard_codeword_conversion}), our \textit{unipolar} Hadamard matrix for \textit{classification purposes} is given by 
\begin{equation}
\label{equ:hadamardForClassification}
\mathbf{H} =
\left(
\begin{array}{cccc}
\mathbf{H}^{(L)}_1 & \mathbf{H}^{(L)}_2 & \cdots & \mathbf{H}^{(L)}_S
\end{array}
\right) \in \mathbb{B}^{L \times S}.
\end{equation}
In analogy, the \textit{bipolar} Hadamard matrix for classification results in
\begin{equation}
\label{equ:hadamardForClassification-11}
\check{\mathbf{H}} =
\left(
\begin{array}{cccc}
\check{\mathbf{H}}^{(L)}_1 & \check{\mathbf{H}}^{(L)}_2 & \cdots & \check{\mathbf{H}}^{(L)}_S
\end{array}
\right) \in \{-1, 1\}^{L \times S}.
\end{equation}
Using $\textbf{1}^{(L\times S)}=\{1\}^{L\times S}$, the two notations (\ref{equ:hadamardForClassification}) and (\ref{equ:hadamardForClassification-11}) are related by
\begin{equation}
\label{equ:H_Hprime}
\check{\mathbf{H}} = 2\mathbf{H}-\textbf{1}^{(L\times S)}.
\end{equation}
Since the matrix $\check{\mathbf{H}}$(\ref{equ:hadamardForClassification-11}) is orthogonal, the following holds:
\begin{equation}
\label{equ:pseudo_inverse}
\check{\mathbf{H}}^\mathsf{T} \cdot \check{\mathbf{H}} = L\cdot \mathbf{I},
\end{equation}
where $\textbf{I} \in \mathbb{B}^{S\times S}$ denotes the identity matrix and $(\ )^\mathsf{T}$ is the transpose.\begingroup
\renewcommand\thefootnote{\textcolor{black}{*}}\footnotemark\endgroup

\subsection{\texttt{HadamardNet} and Hadamard Decoder}
\label{subsec:dec}

\begin{figure}[t!]
    \centering
    \input{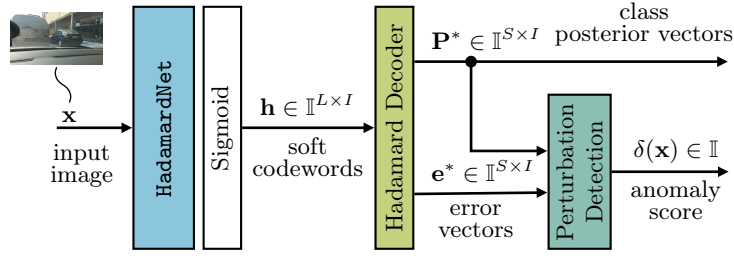}
    \caption{Inference setup using the \texttt{HadamardNet}. The Hadamard decoder is performed according to \cref{subsec:dec}. The perturbation detection is detailed in \cref{subsec:attack_detection}.}
    \label{fig:inference}
\end{figure}

Following \cref{fig:inference}, our proposed \texttt{HadamardNet} predicts a so-called \textit{soft} codeword
\begin{equation} 
\mathbf{h} =(\mathbf{h}_i) = (h_{i,\ell}) \in \mathbb{I}^{L\times I}
\end{equation} for each position, i.e.,~a pixel or an object, $i \in \mathcal{I} = \{1,\dots,I\}$, where we have $\mathbb{I}=[0,1]$ due to the sigmoid and $I \in \mathbb{N}$ is either the number of pixels or the number of objects. These predicted soft codewords $\mathbf{h}$ are then decoded (\cref{fig:inference}, green box) to deliver the optimal class posterior probabilities
\begin{equation}
\mathbf{\PROB}^* = (\mathbf{\PROB}^*_i) = (\PROB^*_{i,s}) \in \mathbb{I}^{S\times I},
\end{equation}
with $\PROB^*_{i,s} = \PROB^*_i(s|\mathbf{x})$, $i \in \mathcal{I}$, and $\mathbf{x}\in \mathbb{I}^{H\times W \times C}$ being the input image of height $H$, width $W$, and with $C=3$ color channels. For each pixel or object $i \in \mathcal{I}$, the class posteriors have to fulfill the stochastic constraints
\begin{align}
\sum_{s\in\mathcal{S}} \PROB^{*}_{i,s} & = 1
        \label{equ:nebenbedingung1}\\[4pt]
\text{and }\quad \PROB^{*}_{i,s} & \ge 0 \qquad \forall s\in\mathcal{S}.
        \label{equ:nebenbedingung2}
\end{align}
\subsubsection{Optimization Problem\normalfont{:}} The estimation of the optimal class posterior probabilities $\PROB^*_{i}(s|\mathbf{x})$ given the predicted soft codewords $\mathbf{h}=(\mathbf{h}_i)$ is now step-by-step derived and outlined in the remainder of this section.

In the unlikely case that the network predicts an error-free soft codeword $\mathbf{h}_i^*=(h^*_{i,\ell})$, which does not contain inconsistencies, the soft codeword fulfills $\mathbf{h}_i^\ast \in \mathcal{H}^*= \{ \sum_{s \in S} \mathbf{H}_s^{(L)}
\cdot P_{i,s}
\mid
\mathbf{P}_i = (P_{i,s}) \in \mathcal{P} \}$ and therefore
\begin{equation}
\label{equ:sum_pre_LGS_H2}
    h^*_{i,\ell} = \sum_{s \in \mathcal{S}} H^{(L)}_{s,\ell} \cdot \PROB_{i,s}.
\end{equation}
Here, $\mathcal{P} = \left\{ \textbf{P}_i = (P_{i,s}) \in \mathbb{I}^S \;\middle|\; \PROB_{i,s} \geq 0 \;\forall s \in \mathcal{S}, \; \sum_{s \in \mathcal{S}} \PROB_{i,s} = 1 \right\}$ is the set of all valid class-wise probability distributions satisfying the constraints (\ref{equ:nebenbedingung1}), (\ref{equ:nebenbedingung2}). In Supplement \ref{supp:classProsteriorConstraintsandGeneralOptimizationProblem}
we show a detailed derivation of (\ref{equ:sum_pre_LGS_H2}) using our probabilistic approach.
Given the unipolar Hadamard matrix for classification $\mathbf{H}$ (\ref{equ:hadamardForClassification}), and the \textit{error-free} soft codeword $\mathbf{h}^*_i$ for pixel or object $i$, we obtain a well-determined LES that we aim to solve for $\textbf{P}_i \in \mathcal{P}$ for each pixel or object $i \in \mathcal{I}$:
\begin{equation}
\label{equ:LGS2}
\mathbf{H} \cdot \textbf{P}_i = \mathbf{h}^*_i.
\end{equation}
Since the prediction $\mathbf{h}_{i} = (h_{i,\ell})$ for each bit at position $\ell$ is intended to be mutually independent, contradictions can arise and potentially an output soft codeword $\mathbf{h}_{i} \notin \mathcal{H}^*$, making the equation system unsolvable under the probabilistic constraints (\ref{equ:nebenbedingung1}), (\ref{equ:nebenbedingung2}). 
To address this, we introduce an error vector $\check{\mathbf{e}}_i = (\check{e}_{i,\ell}) \in [-1, 1]^L$ for each predicted soft codeword $\mathbf{h}_i$, with $\mathbf{h}^*_i=\mathbf{h}_i+\check{\mathbf{e}}_i$, and obtain
\begin{equation}
\label{equ:LGS}
\boxed{\mathbf{H} \cdot \textbf{P}_i = \mathbf{h}_i + \check{\mathbf{e}}_i.}
\end{equation}
This allows (\ref{equ:LGS}) to become solvable for any $\mathbf{h}_i \in \mathbb{I}^L$ subject to the constraints (\ref{equ:nebenbedingung1}), (\ref{equ:nebenbedingung2}).
Using (\ref{equ:H_Hprime}), (\ref{equ:pseudo_inverse}), and (\ref{equ:nebenbedingung1}), we can rearrange (\ref{equ:LGS}) for $\textbf{P}_i$, yielding
\begin{equation}
\label{equ:u_i}
\boxed{\mathbf{P}_i = \tilde{\mathbf{P}}_i + \mathbf{e}_i,}
\end{equation}
with $\tilde{\textbf{P}}_i = \frac{1}{L}\check{\mathbf{H}}^\mathsf{T}\check{\mathbf{h}}_i \in [-1,1]^S$, $\mathbf{e}_i = \frac{2}{L}\check{\mathbf{H}}^\mathsf{T}\check{\mathbf{e}}_i = \frac{2}{L}(\check{\mathbf{e}}^{\mathsf{T}}_i \check{\mathbf{H}})^{\mathsf{T}} \in \mathbb{R}^S$ and $\check{\mathbf{h}}_i=2{\mathbf{h}}_i-\mathbf{1}^{(L)}\in [-1,1]^L$. Note that $\tilde{\mathbf{P}}_i$ does not necessarily satisfy the constraints (\ref{equ:nebenbedingung1}), (\ref{equ:nebenbedingung2}). The rearrangement of (\ref{equ:LGS}) to obtain (\ref{equ:u_i}) is detailed in Supplement \ref{supp:OptimizationProblemGeneral}.
By introducing new unknowns $\check{\mathbf{e}}_i$, the LES (\ref{equ:LGS}) becomes under-determined, resulting in multiple feasible solutions for $\mathbf{P}_i$ and $\check{\mathbf{e}}_i$. Therefore, we employ (\ref{equ:LGS}) and search for the minimum squared $L_2$ error $\min_{\textbf{P}_i \in \mathcal{P}} \|\check{\mathbf{e}}_i\|^2_2$, which is the optimization problem we aim to solve.
\subsubsection{Solving the Optimization Problem\normalfont{:}} We solve this optimization problem by just minimizing $\|\mathbf{e}_i\|_2^2$ in (\ref{equ:u_i}), which we prove to be equivalent by showing that  $\arg\min_{\mathbf{P}_i \in \mathcal{P}} \|\mathbf{e}_i\|_2^2 
=\arg\min_{\mathbf{P}_i \in \mathcal{P}} \|\check{\mathbf{e}}_i\|_2^2$ in Supplement \ref{supp:EqualityErrorVector}.

To minimize $\|\mathbf{e}_i\|_2^2$ in (\ref{equ:u_i}), we utilize the projection $\boldsymbol{\phi}()$ onto the probability simplex by Michelot \cite{Michelot1986} and obtain the optimal class posterior probabilities delivered by the Hadamard decoder in \cref{fig:inference}:
\begin{equation}
\label{equ:ui_min}
\boxed{\mathbf{P}_i^{*}  = \boldsymbol{\phi}(\tilde{\mathbf{P}}_i) = \arg\min_{\mathbf{P}_i \in \mathcal{P}} \| \mathbf{P}_i - \tilde{\mathbf{P}}_i \|_2^2 = \arg\min_{\mathbf{P}_i \in \mathcal{P}} \| \mathbf{e}_i \|_2^2.}
\end{equation}
Please refer to Supplement \ref{supp:SimplexProjection}
for a detailed description of the simplex projection. Substituting $\mathbf{P}_i$ in (\ref{equ:u_i}) by $\mathbf{P}_i^{*}$ delivered by the simplex projection (\ref{equ:ui_min}) yields the smallest class-wise error vector
\begin{equation}
\label{equ:e_i^*}
\mathbf{e}_i^{*} = \mathbf{P}_i^{*} - \tilde{\mathbf{P}}_i \in \mathbb{R}^S.
\end{equation}
As depicted in \cref{fig:inference}, the class-wise probabilities $\mathbf{P}_i^{*}$ and error vector $\mathbf{e}_i^*$ are the two outputs of our Hadamard decoding. The error vector represents the inconsistencies of the individual binary predictions of the \texttt{HadamardNet}. Since all operations in the Hadamard decoding are differentiable, we can utilize the decoded probabilities $\mathbf{P}_i^{*}$, the error vector $\mathbf{e}_i^{*}$, as well as the predicted soft codeword $\mathbf{h}_i$ for the training.

\subsection{Adversarial Attack and Disturbance Detection}
\label{subsec:attack_detection}
As depicted in \cref{fig:inference}, our perturbation detection approach uses the output of our Hadamard decoding $\mathbf{P}^* = (\mathbf{P}^*_i)$ (\ref{equ:ui_min}) and $\mathbf{e}^* = (\mathbf{e}^*_i)$ (\ref{equ:e_i^*}) as inputs and outputs an anomaly score $\delta(\mathbf{x}) \in \mathbb{I}$ for each image $\mathbf{x}$. During inference, an image is flagged as adversarial if $\delta(\mathbf{x}) \geq \theta$ for a given threshold $\theta \geq 0$.
\subsubsection{Error Vector Anomaly Score\normalfont{:}} A simple first approach uses the $L_1$ norm of the error vector, averaged over all pixels or objects, as anomaly score
\begin{align}
\label{equ:anomaly_simple}
\delta(\mathbf{x}) & = \frac{1}{|\mathcal{I}|}\sum_{i\in \mathcal{I}} \|\mathbf{e}_i^*\|_1.
\end{align}
\subsubsection{Regression Anomaly Score\normalfont{:}} Clean images show a strong relationship between error magnitude $\lVert \mathbf{e}_i^* \rVert_1$ and class posterior probabilities $\mathbf{P}_i^*$. An analysis of this relationship is provided in Supplement \ref{supp:correlation_e_p}. 
Pixels or objects with low maximum posterior probability ($\PROB_{i,s^*}^*=\max_{s \in \mathcal{S}}\PROB_{i,s}^* \approx \tfrac{1}{S}$) typically exhibit larger inconsistencies $\|\mathbf{e}_i^*\|_1$. Thus, a detector based solely on (\ref{equ:anomaly_simple}) would tend to falsely alarm on images containing many low-confidence predictions. Under perturbations, the distribution of $\|\mathbf{e}_i^*\|_1$ changes from the clean-data distribution.
We introduce a regression model $\mathcal{F}$ (details in Supplement \ref{supp:details_F_reg}) that, for each pixel or object $i$, predicts the error magnitude $\widehat{\lVert \mathbf{e}_i^* \rVert}_1 = \mathcal{F}([\PROB_{i,s^*}^*,{H}(\mathbf{P}^*_i)]) \ge 0$ from the maximum posterior probability $\PROB_{i,s^*}^*$ and the entropy of the class posterior ${H}(\mathbf{P}^*_i)=-\sum_{s \in \mathcal{S}} \PROB_{i,s}^*\cdot\log(\PROB_{i,s}^*) \ge 0$. The per-image prediction error is then $r(\mathbf{x}) = \frac{1}{\sqrt{|\mathcal{I}|}} \cdot \sum_{i \in \mathcal{I}} L_2(\lVert \mathbf{e}_i^* \rVert_1,\widehat{\lVert \mathbf{e}_i^*\rVert}_1) \ge 0$.
Using clean training data $\mathcal{D}_{\mathrm{train}}$, we compute the mean prediction error $\mu^{\mathrm{train}} = \frac{1}{|\mathcal{D}_{\mathrm{train}}|} \cdot \sum_{\mathbf{x} \in \mathcal{D}_{\mathrm{train}}}r(\mathbf{x}) \ge 0$. The anomaly score is defined as the absolute deviation from the training mean
\begin{align}
\label{equ:anomaly_score_pred_e}
    \delta(\mathbf{x}) & = |r(\mathbf{x}) - \mu^{\mathrm{train}}|.
\end{align}
\subsubsection{Quantile Anomaly Score\normalfont{:}} Using the regression anomaly score, the regression model $\mathcal{F}$ (details in Supplement \ref{supp:details_F_reg}) predicts only the expected value of $\lVert \mathbf{e}_i^* \rVert_1$ for a pixel or object $i$. However, this formulation does not account for variability, and thus yields on average high anomaly scores in case of large standard deviations of $\lVert \mathbf{e}_i^* \rVert_1$. To explicitly incorporate standard deviation, our third approach uses a regression network $\mathcal{F}$ that predicts the conditional $85\%$ quantile of $\lVert \mathbf{e}_i^* \rVert_1$ over all pixels or objects. This means that $\mathcal{F}([\PROB_{i,s^*}^*, {H}(\mathbf{P}^*_{i})]) = E_i^{(0.85)} \ge 0$ provides, for each pixel or object $i \in \mathcal{I}$, the error vector $L_1$ norm $E_i^{(0.85)}$ for which $85\%$ of $\lVert \mathbf{e}_i^* \rVert_1$, $i \in \mathcal{I}$, are lower.
To obtain the anomaly score, we then count the fraction of pixels or objects whose $L_1$ norm of the error vector exceeds the predicted $85\%$ threshold:
\begin{align}
\label{equ:anomaly_score_reg}
\delta(\mathbf{x})
& = \frac{1}{|\mathcal{I}|}\,
 \sum_{i \in \mathcal{I} ,\, \lVert \mathbf{e}_i^* \rVert_1 > E_i^{(0.85)}}1.
\end{align}
\pagebreak

\subsection{\texttt{HadamardNet} and Attack/Disturbance Detector Training}

\subsubsection{\textbf{\texttt{HadamardNet}}\normalfont{:}}
In our final \texttt{HadamardNet}, training is based on the encoded-space loss
\begin{equation}
\label{equ:overall_loss}
   J^{\mathrm{cls}} = J^{\mathrm{ENC}}(\mathbf{h}_i, \overline{\mathbf{H}}_i),
\end{equation}
computed between the predicted soft codeword $\mathbf{h}_i$ and the target Hadamard codeword
$\overline{\mathbf{H}}_i=\mathbf{H}^{(L)}_{s=\overline{s}_i}$
(\ref{equ:hadamard_codeword_conversion}), with $\overline{s}_i$ denoting the class label of pixel or object $i$.
Details on $ J^{\mathrm{cls}}$ (51) and $J^{\mathrm{ENC}}$ (52) are given in Supplement \ref{supp:details_hadamardnet_training}.

\subsubsection{Detector\normalfont{:}} After \texttt{HadamardNet} training, the regression model $\mathcal{F}$ \textit{is trained exclusively on clean images} by minimizing $J^{\mathrm{MSE}}(\lVert \mathbf{e}_i^* \rVert_1,\widehat{\lVert \mathbf{e}_i^*\rVert}_1)$ for the regression anomaly score (\ref{equ:anomaly_score_pred_e}), and by minimizing the $85\%$ quantile loss \cite{koenker1978regression}
\begin{equation}
\label{equ:quantile_loss}
J^{(0.85)}
= \frac{1}{|\mathcal{I}|}\sum_{i \in \mathcal{I}}
\max\bigl(0.85 \,(\lVert \mathbf{e}_i^* \rVert_1 - E_i^{(0.85)}) ,\, 
        0.15 \,(E_i^{(0.85)}-\lVert \mathbf{e}_i^* \rVert_1) \bigr)
\end{equation}
for the quantile anomaly score (\ref{equ:anomaly_score_reg}).

\section{Experimental Setup and Metrics}
\label{sec:exp_setup}

\subsubsection{Semantic Segmentation\normalfont{:}}
\label{subsec:train_semantic}
For semantic segmentation, \texttt{SegFormer MiT-B0} \cite{Xie2022segformer} is used in most of our studies, due to its fast training and strong performance. We additionally report results for \texttt{DeepLabv3+} \cite{Chen2018a}. We employ Cityscapes \cite{Cordts2016} training ($\mathcal{D}^\mathrm{CS}_\mathrm{train}$) and validation ($\mathcal{D}^\mathrm{CS}_\mathrm{val}$) sets, BDD100K \cite{Yu2019} training ($\mathcal{D}^\mathrm{BDD,Seg}_\mathrm{train}$) and validation ($\mathcal{D}^\mathrm{BDD,Seg}_\mathrm{val}$) sets, and ADE20K \cite{Zhou2017} training ($\mathcal{D}^\mathrm{ADE20K}_\mathrm{train}$) and validation ($\mathcal{D}^\mathrm{ADE20K}_\mathrm{val}$) sets. Segmentation quality is measured using mean intersection over union (mIoU). The \texttt{mmsegmentation} framework \cite{mmseg2020} is used to train all models using default hyperparameters, except for the loss, where we use exclusively $J^{\mathrm{cls}}$ (\ref{equ:overall_loss}). Supplement \ref{supp:details_hyperparameters} 
provides further training details.

\subsubsection{Object Detection\normalfont{:}}
\label{subsec:train_object_det}
For object detection, we consider \texttt{Faster R-CNN}~\cite{Ren2015} as a standard two-stage CNN-based detector widely used in the adversarial attack and detection literature \cite{Chow2020, Li2020g, Yin2021}. We further extend our approach to \texttt{DETR}~\cite{Carion2020}, an end-to-end single-stage transformer-based detector, demonstrating that our method is architecture-agnostic. We employ the combined Pascal VOC 2007 and 2012 trainval \cite{everingham2010pascal,Everingham2015} ($\mathcal{D}_\mathrm{train}^{\mathrm{VOC}}$) and Pascal VOC 2007 test ($\mathcal{D}_\mathrm{val}^{\mathrm{VOC}}$) sets, BDD100K~\cite{Yu2019} train ($\mathcal{D}_{\mathrm{train}}^{\mathrm{BDD,OD}}$) and test ($\mathcal{D}_{\mathrm{test}}^{\mathrm{BDD,OD}}$) sets, and COCO~\cite{Lin2014microsoft} train ($\mathcal{D}_{\mathrm{train}}^{\mathrm{COCO}}$) and validation ($\mathcal{D}_{\mathrm{val}}^{\mathrm{COCO}}$) sets. Object detection quality is measured using the COCO average precision (AP) metric \cite{Lin2014microsoft}. Specifically, AP denotes  the mean average precision averaged over multiple IoU thresholds from 0.50 to 0.95 in steps of 0.05. The models are trained by minimizing
\begin{equation}
\label{equ:object_detection_loss}
J^{\mathrm{OD}} = \lambda^{\mathrm{reg}} J^{\mathrm{reg}} + \lambda^{\mathrm{cls}} J^{\mathrm{cls}},
\end{equation}
where $\lambda^{\mathrm{reg}},\lambda^{\mathrm{cls}} > 0$ weigh the bounding box regression loss $J^{\mathrm{reg}}$ and classification loss $J^{\mathrm{cls}}$ (\ref{equ:overall_loss}). Since Hadamard encodings are applied only to the classification branch, we use the standard $J^{\mathrm{reg}}$ from \cite{Ren2015,Carion2020} for \texttt{Faster R-CNN} and \texttt{DETR}. For classification, we employ \texttt{HadamardNet} as in the segmentation experiments, with $J^{\mathrm{cls}}$ in (\ref{equ:object_detection_loss}) as defined in (\ref{equ:overall_loss}). Unless noted otherwise, all hyperparameters follow the defaults of the \texttt{mmdetection} framework \cite{mmdetection}. Supplement  \ref{supp:details_hyperparameters}
provides further training details. For both, semantic segmentation and object detection, we report 95\% confidence intervals obtained over three seeds.

\subsubsection{Perturbation Detection and Attacks\normalfont{:}}
\label{subsec:train_regression}
The perturbation detection regression network $\mathcal{F}$ is trained in a second stage after the segmentation or object detection model have been trained. For this purpose, we use the same training set as in the first stage, which again only contains clean data. We train $\mathcal{F}$ for $20$ epochs using the AdamW \cite{Loshchilov2019} optimizer, more details in Supplement \ref{supp:details_hyperparameters}.

For semantic segmentation, we employ FGSM \cite{Goodfellow2015}, PGD \cite{Madry2018}, and Metzen \cite{Metzen2017} attacks. For object detection, we employ TOG \cite{Chow2020} untargeted, fabrication, vanishing and mislabeling attacks. Additionally, for both, we employ Gaussian noise and salt and pepper disturbances. For all experiments evaluating perturbation detection performance, the corresponding perturbation strength $\epsilon$ \cite{Klingner2020} is explicitly stated, which is the only hyperparameter used for Gaussian noise, salt and pepper noise, and the FGSM attack. The PGD, TOG, and Metzen attacks use additional hyperparameters. For PGD and TOG, we use $10$ iterations, each with a step size of $\alpha_\epsilon = \tfrac{\epsilon}{4}$. For the Metzen attack, we use $60$ iterations, a fixed step size of $\alpha^\mathrm{Metzen}_\epsilon = \tfrac{1}{255}$, and match all remaining hyperparameters to those of Metzen et al.\ \cite{Metzen2017}. Details on our definition of $\epsilon$ and the generation of the attacks and disturbances can be found in Supplement \ref{supp:attack_strength}.

Perturbation detection performance is evaluated using the true positive rate (TPR), false positive rate (FPR), and the area under the receiver operating characteristic curve (AuROC). In addition, to assess generalization across different perturbations and strengths, we report \textit{global} AuROC. To compute these global metrics, we aggregate detection outputs either across all perturbation strengths for a fixed perturbation, or across all perturbations and strengths, and evaluate AuROC on the resulting combined set. \textit{This protocol enforces a single operating threshold across all perturbations and strengths, reflecting practical deployment conditions where the perturbation configuration is unknown.}

\section{Experimental Evaluation and Discussion}
\label{sec:experiments}

\subsubsection{Semantic Segmentation\normalfont{:}}
\label{sec:semanticSegmentation}
Note that in Supplement \ref{supp:SemanticSegmentationAblation},
we conduct an ablation study on the variants and combinations of the classification loss  $J^{\mathrm{cls}}$ (\ref{equ:overall_loss}) components and adopt $J^{\mathrm{cls}} = J^{\mathrm{ENC}}$ with $J^{\mathrm{ENC}} = J^{\mathrm{MSE}}$ \textit{for all following experiments both in the semantic segmentation and object detection task.}

In \cref{tab:anomaly_score_func}, we compare the perturbation detection performance of all three variants of our proposed method described in \cref{subsec:attack_detection} with state-of-the-art detection approaches for semantic segmentation across multiple perturbation types and strengths $\epsilon$. All methods employ a \texttt{SegFormer MiT-B0} \cite{Xie2022segformer} trained on $\mathcal{D}^\mathrm{CS}_\mathrm{train}$ and evaluated on perturbed samples from $\mathcal{D}^\mathrm{CS}_\mathrm{val}$. We report the perturbation detection accuracy measured by global AuROC (\%) per perturbation across all strengths and across all perturbations and strengths. Perturbed mIoU and detection performance per strength $\epsilon$ are provided in Supplement \ref{supp:SemanticSegmentationPerturbation}.
We observe that the three-pass detector by Xu et al.\ \cite{Xu2018e} perform best in three of five perturbations, however, naturally requiring three times the computational complexity of the single-pass detectors.
\begin{table}[t!]
\centering
\caption{\textbf{Perturbation detection for semantic segmentation} on $\mathcal{D}^\mathrm{CS}_\mathrm{val}$ using \texttt{SegFormer MiT-B0}. Global AuROC (best in bold) per perturbation across all strengths $\epsilon\!\in\!\{1,2,4,8,16\}$ and across all perturbations and strengths in \%.}
\label{tab:anomaly_score_func}
\setlength{\tabcolsep}{0.1pt}        
\begin{tabular}{l|c|ccccc|c|c}
\toprule[.9pt]
\multirow{2}{*}{Method} &
\multirow{2}{*}{\begin{tabular}{c}Optim.\\on\end{tabular}} &
\multicolumn{5}{c|}{Global AuROC (\%)} &
\multirow{2}{*}{\begin{tabular}{c}Global\\AuROC\end{tabular}} &
\multirow{2}{*}{\begin{tabular}{c}FLOPs\\(G)\end{tabular}} \\

& & FGSM & PGD & Metzen & Gauss. & S\&P & \\

\midrule
\multicolumn{9}{c}{Multi-pass detector} \\
\midrule
Xu et al.\ \cite{Xu2018e} & - &
$97.1^{\pm 0.3}$ & $98.2^{\pm 0.4}$ & $78.0^{\pm 1.0}$ &
$64.2^{\pm 1.3}$ & $65.4^{\pm 0.8}$ &
$80.7^{\pm 0.6}$ & $366$ \\
\midrule
\multicolumn{9}{c}{Single-pass detector} \\
\midrule
Entropy \cite{Smith2018} & - &
$95.7^{\pm 0.5}$ & $92.2^{\pm 1.1}$ & $45.0^{\pm 1.6}$ &
$62.9^{\pm 0.6}$ & $62.1^{\pm 0.5}$ &
$71.6^{\pm 0.2}$ & $122$ \\
Max posterior \cite{Hendrycks2017} & - &
$95.5^{\pm 0.4}$ & $91.7^{\pm 1.2}$ & $45.2^{\pm 1.8}$ &
$62.8^{\pm 0.1}$ & $61.6^{\pm 0.5}$ &
$71.4^{\pm 0.2}$ & $122$ \\
Maag et al.\ \cite{maag2024detectingadversarialattackssemantic} & FGSM &
$\mathbf{98.2}^{\pm 0.0}$ & $94.9^{\pm 2.7}$ & $48.0^{\pm 2.9}$ &
$67.6^{\pm 0.6}$ & $64.3^{\pm 0.2}$ &
$74.8^{\pm 0.9}$ & $122$ \\
Ours: Error (\ref{equ:anomaly_simple}) & - &
$95.7^{\pm 0.8}$ & $95.2^{\pm 0.8}$ & $68.4^{\pm 1.9}$ &
$65.0^{\pm 0.8}$ & $60.6^{\pm 0.9}$ &
$77.1^{\pm 0.7}$ & $123$ \\
Ours: Reg. (\ref{equ:anomaly_score_pred_e}) & - &
$89.6^{\pm 2.8}$ & $94.1^{\pm 0.9}$ & $72.7^{\pm 3.3}$ &
$59.9^{\pm 1.4}$ & $55.9^{\pm 1.0}$ &
$74.6^{\pm 1.7}$ & $124$ \\
Ours: Quantile (\ref{equ:anomaly_score_reg}) & - &
$95.9^{\pm 0.7}$ & $\mathbf{96.1}^{\pm 0.5}$ &
$\mathbf{75.9}^{\pm 2.1}$ &  $\mathbf{68.5}^{\pm 1.1}$ &
$\mathbf{64.4}^{\pm 1.0}$ &
$\mathbf{80.2}^{\pm 0.9}$ & $124$ \\
\bottomrule[.9pt]
\end{tabular}
\end{table}
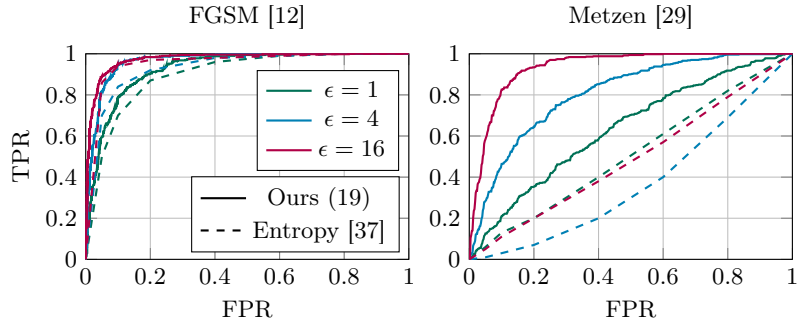
\begin{figure}[t!]
    \centering
    \begin{tikzpicture}
\begin{groupplot}[
    group style={
        group size=2 by 1,
        horizontal sep=0.8cm
    },
    width=0.48\textwidth,
    height=4.3cm,
    xmin=0, xmax=1,
    ymin=0, ymax=1,
    xlabel={FPR},
    ylabel={TPR},
    ylabel style={yshift=-0mm},
    grid=both
]

\nextgroupplot[
    title={FGSM \cite{Goodfellow2015}},
    legend style={at={(0.97,0.43)}, anchor=south east, font=\small}
]

\addplot[thick,tu10] table [col sep=comma, x=fpr, y=tpr] {figures/roc_Quantile_anomaly_score_fgsm_ours_eps1.csv};
\addplot[thick,tu10,dashed,forget plot] table {
0    0
0.05 0.5
0.10 0.7
0.20 0.87
0.40 0.96
0.60 0.99
0.80 1.00
1.00 1.00
};

\addplot[thick,tu5] table [col sep=comma, x=fpr, y=tpr] {figures/roc_Quantile_anomaly_score_fgsm_ours_eps4.csv};
\addplot[thick,tu5,dashed,forget plot] table {
0    0
0.05 0.70
0.10 0.84
0.20 0.92
0.40 0.98
0.60 0.99
0.80 1.00
1.00 1.00
};

\addplot[thick,tu3] table [col sep=comma, x=fpr, y=tpr] {figures/roc_Quantile_anomaly_score_fgsm_ours_eps16.csv};
\addplot[thick,tu3,dashed,forget plot] table {
0    0
0.05 0.86
0.10 0.94
0.20 0.97
0.40 0.98
0.60 0.99
0.80 1.00
1.00 1.00
};

\addlegendentry{$\epsilon = 1$}
\addlegendentry{$\epsilon = 4$}
\addlegendentry{$\epsilon = 16$}

\coordinate (fgsmLegendPos) at (rel axis cs:0.97,0.03);

\nextgroupplot[
    title={Metzen \cite{Metzen2017}},
    ylabel={}
]

\addplot[thick,tu10] table [col sep=comma, x=fpr, y=tpr] {figures/roc_Quantile_anomaly_score_metzen_ours_eps1.csv};
\addplot[thick,tu10,dashed,forget plot] table {
0    0
0.05 0.06
0.10 0.13
0.20 0.20
0.40 0.40
0.60 0.61
0.80 0.82
1.00 1.00
};

\addplot[thick,tu5] table [col sep=comma, x=fpr, y=tpr] {figures/roc_Quantile_anomaly_score_metzen_ours_eps4.csv};
\addplot[thick,tu5,dashed,forget plot] table {
0    0
0.05 0.01
0.10 0.03
0.20 0.07
0.40 0.20
0.60 0.40
0.80 0.69
1.00 1.00
};

\addplot[thick,tu3] table [col sep=comma, x=fpr, y=tpr] {figures/roc_Quantile_anomaly_score_metzen_ours_eps16.csv};
\addplot[thick,tu3,dashed,forget plot] table {
0    0
0.05 0.05
0.10 0.11
0.20 0.20
0.40 0.38
0.60 0.57
0.80 0.79
1.00 1.00
};

\end{groupplot}

\begin{axis}[
    hide axis,
    scale only axis,
    width=0pt,
    height=0pt,
    at={(0,0)},
    anchor=south west,
    xmin=0, xmax=1,
    ymin=0, ymax=1,
    legend to name=methodlegend3,
    legend style={draw=black,fill=white,font=\small}
]
\addlegendimage{black,thick}
\addlegendentry{Ours (\ref{equ:anomaly_score_reg})}
\addlegendimage{black,thick,dashed}
\addlegendentry{Entropy \cite{Smith2018}}
\end{axis}

\node at (fgsmLegendPos) [anchor=south east, xshift=3.5pt, yshift=-3pt]
    {\pgfplotslegendfromname{methodlegend3}};

\end{tikzpicture}
    \caption{\textbf{ROC for our proposed quantile detection (\ref{equ:anomaly_score_reg}) for semantic segmentation}, left for the FGSM~\cite{Goodfellow2015} attack, right for the Metzen~\cite{Metzen2017} attack at perturbation strengths $\epsilon \in \{1, 4, 16\}$. All results are obtained using a \texttt{SegFormer MiT-B0} model \cite{Xie2022segformer} trained on $\mathcal{D}^\mathrm{CS}_\mathrm{train}$ and evaluated on perturbed images from $\mathcal{D}^\mathrm{CS}_\mathrm{val}$.}
    \label{fig:ROC_seg}
\end{figure}
In the single-pass regime, Maag et al.\ \cite{maag2024detectingadversarialattackssemantic} excel in FGSM (where they optimized for). The entropy \cite{Smith2018} and the max posterior \cite{Hendrycks2017} detectors are lacking behind, so do Maag et al.\ for attacks other than FGSM. Among our methods, our quantile detector is ahead, delivering also the best global AuROC (80.2\%), which is statistically equivalent to the multi-pass performance (80.7\%) of complex Xu et al.\ \cite{Xu2018e}. The second-best of our approaches is the quite balanced error anomaly score (\ref{equ:anomaly_simple}) method (77.1\% global AuROC) which for each single perturbation delivers a global AuROC of more than 60\%.

In \cref{fig:ROC_seg}, we display receiver operating characteristic (ROC) curves for FGSM~\cite{Goodfellow2015} (left) and Metzen~\cite{Metzen2017} (right) attacks on semantic segmentation models for various attack strengths $\epsilon$ for our quantile regression approach (\ref{equ:anomaly_score_reg}) and the entropy baseline \cite{Smith2018}. The results are obtained using the \texttt{SegFormer MiT-B0} model \cite{Xie2022segformer}, which is trained on $\mathcal{D}^\mathrm{CS}_\mathrm{train}$ and evaluated on attacked images from $\mathcal{D}^\mathrm{CS}_\mathrm{val}$. The ROC curves are obtained by varying the decision threshold $\theta$ applied to the anomaly score $\delta(\mathbf{x})$ (\ref{equ:anomaly_score_reg}). Expectedly, we observe that ROC curves get better the stronger the attacks are. Our quantile method excels the entropy approach for both attacks, for Metzen even with high significance.
Further ROC curves are provided in Supplement \ref{supp:ROC_Sem_Seg}.

In \cref{tab:onehotVsHadamard}, we compare the clean mIoU and global AuROC across all perturbations, model size, and computational complexity in terms of GFLOPs of one-hot and Hadamard output encodings for two model architectures and datasets. For perturbation detection with one-hot models, we use the entropy \cite{Smith2018} and max posterior \cite{Hendrycks2017} baselines, while for our \texttt{HadamardNet} we report our three perturbation detectors (\ref{equ:anomaly_simple}), (\ref{equ:anomaly_score_pred_e}), (\ref{equ:anomaly_score_reg}).
\begin{table}[t!]
  \centering
  \caption{{One-hot vs.\ Hadamard output encodings on \textbf{semantic segmentation} models} on $\mathcal{D}^\mathrm{CS}_\mathrm{val}$ and $\mathcal{D}^\mathrm{BDD,Seg}_\mathrm{val}$. Global AuROC (best in bold, second-best underlined) across all perturbations and strengths $\epsilon\!\in\!\{1,2,4,8,16\}$ and mIoU in \%.}
  \label{tab:onehotVsHadamard}
    
\renewcommand{\arraystretch}{0.9}
\setlength{\tabcolsep}{1pt}
\extrarowheight=\aboverulesep
\addtolength{\extrarowheight}{\belowrulesep}
\aboverulesep=1pt
\belowrulesep=-2pt
\begin{tabular}{@{}ll|ccccrr@{}}
\toprule[.9pt]
\multirow{3}{*}{Model} &
Output &
\multicolumn{2}{c}{Cityscapes (CS)} &
\multicolumn{2}{c}{BDD} &
\multirow{3}{*}{\makecell{Size\\(M)}} &
\multirow{3}{*}{\makecell{FLOPs\\(G)}} \\[-5pt]
&representation / &
\multicolumn{1}{c}{mIoU} &
\multicolumn{1}{c}{AuROC} &
\multicolumn{1}{c}{mIoU} &
\multicolumn{1}{c}{AuROC} &
&
\\[-5pt]
&detector &
clean &
attack &
clean &
attack &
&
\\
\midrule

\multirow{5}{*}{\centering\makecell[l]{\texttt{DeepLabv3+}\\ \cite{Chen2018a}}}
& One-hot/entropy~\cite{Smith2018}
& ${80.2}^{\pm 0.2}$ 
& $72.5^{\pm 0.1}$ 
& ${59.8}^{\pm 1.5}$ 
& $59.2^{\pm 3.2}$ 
& $43.6$ 
& $1413$ \\[-3pt]

& One-hot/max post.\ \cite{Hendrycks2017}
& ${80.2}^{\pm 0.2}$ 
& $72.2^{\pm 1.4}$ 
& $59.8^{\pm 1.5}$
& $60.3^{\pm 2.9}$ 
& $43.6$ 
& $1413$ \\[-3pt]

& Hadamard/(\ref{equ:anomaly_simple}) (ours)
& ${80.1}^{\pm 0.1}$ 
& $80.4^{\pm 1.1}$ 
& ${57.2}^{\pm 0.5}$ 
& $\textbf{70.9}^{\pm 2.8}$ 
& $43.6$ 
& $1414$ \\[-3pt]

& Hadamard/(\ref{equ:anomaly_score_pred_e}) (ours)
& ${80.1}^{\pm 0.1}$
& $\mathbf{81.0}^{\pm 0.4}$ 
& ${57.2}^{\pm 0.5}$ 
& ${63.4}^{\pm 6.6}$ 
& $43.6$ 
& $1415$ \\[-3pt]

& Hadamard/(\ref{equ:anomaly_score_reg}) (ours)
& ${80.1}^{\pm 0.1}$ 
& $\underline{80.7}^{\pm 0.2}$ 
& ${57.2}^{\pm 0.5}$ 
& $\underline{64.8}^{\pm 4.6}$ 
& $43.6$ 
& $1415$ \\

\midrule

\multirow{5}{*}{\centering\makecell[l]{\texttt{SegFormer-B0}\\ \cite{Xie2022segformer}}}
& One-hot/entropy~\cite{Smith2018}
& $76.4^{\pm 0.2}$ 
& $72.1^{\pm 1.4}$ 
& ${59.5}^{\pm 0.3}$ 
& $62.0^{\pm 3.0}$  
& $3.7$ 
& $122$ \\[-3pt]

& One-hot/max post.\ \cite{Hendrycks2017}
& $76.4^{\pm 0.2}$ 
& $72.1^{\pm 1.4}$ 
& $59.5^{\pm 0.3}$
& $62.8^{\pm 2.7}$ 
& $3.7$ 
& $122$ \\[-3pt]

& Hadamard/(\ref{equ:anomaly_simple}) (ours)
& ${76.5}^{\pm 0.3}$
& $\underline{77.1}^{\pm 0.7}$
& ${57.6}^{\pm 1.5}$
& $\textbf{70.3}^{\pm 0.6}$
& $3.7$
& $123$ \\[-3pt]

& Hadamard/(\ref{equ:anomaly_score_pred_e}) (ours)
& ${76.5}^{\pm 0.3}$ 
& $74.6^{\pm 1.7}$  
& ${57.6}^{\pm 1.5}$ 
& $64.5^{\pm 1.5}$ 
& $3.7$ 
& $124$ \\[-3pt]

& Hadamard/(\ref{equ:anomaly_score_reg}) (ours)
& ${76.5}^{\pm 0.3}$
& $\textbf{80.2}^{\pm 0.9}$
& ${57.6}^{\pm 1.5}$
& $\underline{67.2}^{\pm 0.9}$ 
& $3.7$
& $124$ \\

\bottomrule[.9pt]
\end{tabular}

\end{table}
%
On clean data, we reach among all methods statistically equivalent semantic segmentation mIoU on $\mathcal{D}^\mathrm{CS}_\mathrm{val}$, and close-by performance on $\mathcal{D}^\mathrm{BDD,Seg}_\mathrm{val}$. \textit{Concerning global AuROC of the detectors, the \texttt{HadamardNet} with our detector (\ref{equ:anomaly_simple}) is significantly better than one-hot encodings with any of the baseline detectors \cite{Smith2018}, \cite{Hendrycks2017} for both models and datasets.} This holds also for our quantile detection (\ref{equ:anomaly_score_reg}) on $\mathcal{D}^\mathrm{CS}_\mathrm{val}$. A comparison to the (very poor-performing) the so-far only Hadamard semantic segmentation method by Hoyos et al.\ \cite{Hoyos2024} is provided in Supplement \ref{supp:comparison_hoyos}.
More results on $\mathcal{D}^\mathrm{BDD,Seg}_\mathrm{val}$ are reported in Supplement \ref{supp:additional_results_bdd_seg} 
and on $\mathcal{D}^\mathrm{ADE20K}_\mathrm{val}$ in Supplement \ref{supp:large_vocabulary_datasets}.

\begin{table}[t]
    \centering
    \caption{Global AuROC (\%) for semantic segmentation on $\mathcal{D}^\mathrm{CS}_\mathrm{val}$, comparing entropy-based one-hot detection \cite{Smith2018}, inconsistency-based detection in the one-hot output space, and inconsistency-based detection in the Hadamard output space. All models use \texttt{DeepLabv3+}. Best results in bold.}
    \label{tab:global_auroc_cs_semseg_oh_vs_had}
    \setlength{\tabcolsep}{3pt}
    \renewcommand{\arraystretch}{0.5}
    \begin{tabular}{ccc}
        \toprule
         One-hot+entropy~\cite{Smith2018} & One-hot+(\ref{equ:anomaly_simple}) & Hadamard+(\ref{equ:anomaly_simple}) \\
        \midrule
        $72.5^{\pm0.1}$ & $76.9^{\pm0.3}$ & $\textbf{80.4}^{\pm1.1}$\\
        \bottomrule
    \end{tabular}
\end{table}
In \cref{tab:global_auroc_cs_semseg_oh_vs_had}, we further investigate whether the performance gain stems only from allowing prediction inconsistencies or whether the Hadamard output encoding itself contributes to perturbation detection. To this end, we train a one-hot model to directly predict $\tilde{\mathbf{P}}_i$ using sigmoid instead of softmax activation, while keeping the remaining setup unchanged, therefore allowing the model to have prediction inconsistencies. We then apply the same simplex projection (\ref{equ:ui_min}), error vector computation (\ref{equ:e_i^*}), and perturbation detector (\ref{equ:anomaly_simple}) as for the \texttt{HadamardNet}. We denote this setup as one-hot+(\ref{equ:anomaly_simple}). Results are reported on $\mathcal{D}^\mathrm{CS}_\mathrm{val}$ using a \texttt{DeepLabv3+}.
Compared to the one-hot entropy baseline \cite{Smith2018}, using the prediction inconsistency in the one-hot space already improves the global AuROC by $4.4\%$ absolute. This shows that the general principle of exploiting inconsistencies between predictions and their projection is already beneficial. However, replacing the one-hot output encoding with Hadamard output encoding further improves the global AuROC by another $3.5\%$ absolute, reaching $80.4\%$. \textit{This shows that both, allowing prediction inconsistencies and structuring them through Hadamard encodings, improve perturbation detection.}

\subsubsection{Object Detection\normalfont{:}}
\label{sec:objectDetection}
In Supplement \ref{supp:ObjectDetectionAblation}, 
we conduct an ablation study on $\mathcal{D}^\mathrm{VOC}_\mathrm{val}$ for weight $\lambda^{\mathrm{cls}}$ of the classification loss $J^{\mathrm{cls}}$ (\ref{equ:overall_loss}) as part of $J^{\mathrm{OD}}$ (\ref{equ:object_detection_loss}), and fix the optimum for all further object detection experiments on all datasets.

In \cref{tab:anomaly_score_func_od}, we compare the perturbation detection performance of all three variants of our proposed method in \cref{subsec:attack_detection} with state-of-the-art detection approaches for object detection across multiple perturbation types and strengths $\epsilon$. All methods employ a \texttt{Faster R-CNN} \cite{Ren2015} model trained on $\mathcal{D}^\mathrm{VOC}_\mathrm{train}$ and evaluated on perturbed samples from $\mathcal{D}^\mathrm{VOC}_\mathrm{val}$. We report the perturbation detection accuracy measured by global AuROC (\%) per perturbation across all strengths and across all perturbations and strengths.
\begin{table}[t!]
\centering
\caption{\textbf{Perturbation detection for object detection} on $\mathcal{D}^\mathrm{VOC}_\mathrm{val}$ using \texttt{Faster R-CNN}. Global AuROC (best in bold) per perturbation (Untargeted, Fabrication, Vanishing, Mislabeling, Gaussian, S\&P) across all perturbation strengths $\epsilon\!\in\!\{0.25, 0.5,1,2,4,8,16\}$ and across all perturbations and strengths in \%.}
\label{tab:anomaly_score_func_od}
\setlength{\tabcolsep}{0.6pt}        
\renewcommand{\arraystretch}{0.92} 
\resizebox{1\linewidth}{!}{
\begin{tabular}{l|cccccc|c|>{\centering\arraybackslash}p{1.1cm}}
\toprule[.9pt]
\multirow{2}{*}{Method} &
\multicolumn{6}{c|}{Global AuROC (\%)} &
\multirow{2}{*}{\begin{tabular}{c}Global\\AuROC\end{tabular}} &
\multirow{2}{*}{\begin{tabular}{c}FLOPs\\(G)\end{tabular}}\\

& Untar. & Fabric. & Vanish. & Mislab. & Gauss & S\&P & \\
\midrule
\multicolumn{9}{c}{Multi-pass detector} \\
\midrule
Xu et al.\ \cite{Xu2018e} & 
$81.0^{\pm 3.9}$ &
$82.1^{\pm 3.9}$ &
$83.5^{\pm 2.6}$ &
$80.9^{\pm 3.9}$ &
$34.6^{\pm 0.7}$ &
$63.4^{\pm 2.4}$ &
$69.8^{\pm 2.6}$ & $645$  \\
\midrule
\multicolumn{9}{c}{Single-pass detector} \\
\midrule
Entropy \cite{Smith2018} &$84.8^{\pm 0.9}$ &
$87.5^{\pm 0.9}$ &
$8.2^{\pm 0.3}$ &
$84.8^{\pm 1.0}$ &
$49.8^{\pm 0.2}$ &
$46.1^{\pm 2.0}$ &
$59.6^{\pm 0.1}$ & $215$ \\
Max posterior \cite{Hendrycks2017} & $92.1^{\pm 0.5}$ &
$94.0^{\pm 0.4}$ &
$5.4^{\pm 0.1}$ &
$92.2^{\pm 0.6}$ &
$48.4^{\pm 0.2}$ &
$40.8^{\pm 1.5}$ &
$61.2^{\pm 0.1}$ &$215$ \\
Ours: Error (\ref{equ:anomaly_simple}) &$92.3^{\pm 1.3}$ &
$93.5^{\pm 1.3}$ &
$9.7^{\pm 0.7}$ &
$92.4^{\pm 1.4}$ &
$49.7^{\pm 0.8}$ &
$47.0^{\pm 0.5}$ &
$63.3^{\pm 0.5}$  &$215$  \\
Ours: Reg. (\ref{equ:anomaly_score_pred_e}) &$\mathbf{93.9}^{\pm 0.9}$ &
$\mathbf{94.7}^{\pm 0.8}$ &
$\mathbf{70.6}^{\pm 1.1}$ &
$\mathbf{93.8}^{\pm 0.9}$ &
$\mathbf{50.2}^{\pm 0.5}$ &
$\mathbf{51.2}^{\pm 2.1}$ &
$\mathbf{74.4}^{\pm 0.4}$ &$215$ \\

Ours: Quantile (\ref{equ:anomaly_score_reg}) &$64.1^{\pm 19}$ &
$64.3^{\pm 18}$ &
$38.5^{\pm 17}$ &
$64.5^{\pm 18}$ &
$47.5^{\pm 0.2}$ &
$49.1^{\pm 4.8}$ &
$54.3^{\pm 17}$&$215$  \\
\bottomrule[.9pt]
\end{tabular}

}
\end{table}
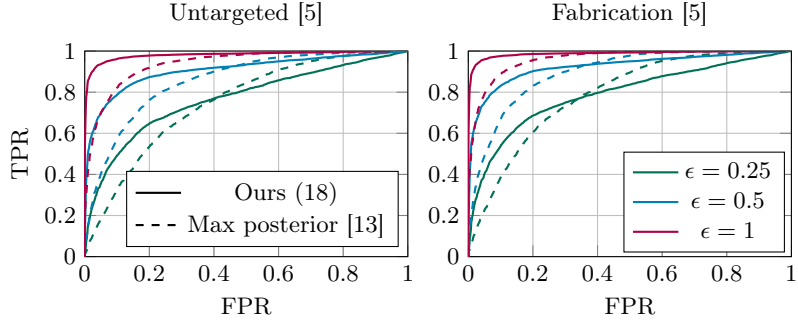
\begin{figure}[t!]
    \centering
    \begin{tikzpicture}
\begin{groupplot}[
    group style={
        group size=2 by 1,
        horizontal sep=0.8cm
    },
    width=0.48\textwidth,
    height=4.3cm,
    xmin=0, xmax=1,
    ymin=0, ymax=1,
    xlabel={FPR},
    ylabel={TPR},
    ylabel style={yshift=-0mm},
    grid=both
]

\nextgroupplot[
    title={Untargeted \cite{Chow2020}},
    legend style={at={(0.97,0.40)}, anchor=south east, font=\small}
]

\addplot[thick,tu10]
table[
  col sep=comma,
  x=FPR,
  y=TPR,
  each nth point=10,      
] {figures/data/untargeted/nonlinear_eps0_25_roc_data.csv};

\addplot[thick,tu10,dashed,forget plot]
table[
  col sep=comma,
  x=FPR,
  y=TPR,
  each nth point=10,      
] {figures/data/untargeted/maxposterior_eps0_25_roc_data.csv};

\addplot[thick,tu5]
table[
  col sep=comma,
  x=FPR,
  y=TPR,
  each nth point=10,      
] {figures/data/untargeted/nonlinear_eps0_5_roc_data.csv};

\addplot[thick,tu5,dashed,forget plot]
table[
  col sep=comma,
  x=FPR,
  y=TPR,
  each nth point=10,      
] {figures/data/untargeted/maxposterior_eps0_5_roc_data.csv};

\addplot[thick,tu3]
table[
  col sep=comma,
  x=FPR,
  y=TPR,
  each nth point=10,      
] {figures/data/untargeted/nonlinear_eps1_roc_data.csv};

\addplot[thick,tu3,dashed,forget plot]
table[
  col sep=comma,
  x=FPR,
  y=TPR,
  each nth point=10,      
] {figures/data/untargeted/maxposterior_eps1roc_data.csv};

\nextgroupplot[
    title={Fabrication \cite{Chow2020}},
    ylabel={},
    legend style={
        at={(axis description cs:0.98,0.02)},
        anchor=south east
    }
]

\addplot[thick,tu10]
table[
  col sep=comma,
  x=FPR,
  y=TPR,
  each nth point=10,      
] {figures/data/fabrication/nl_0_25_roc_data.csv};

\addplot[thick,tu10,dashed,forget plot]
table[
  col sep=comma,
  x=FPR,
  y=TPR,
  each nth point=10,      
] {figures/data/fabrication/mp_0_25_roc_data.csv};

\addplot[thick,tu5]
table[
  col sep=comma,
  x=FPR,
  y=TPR,
  each nth point=10,      
] {figures/data/fabrication/nl_0_5_roc_data.csv};

\addplot[thick,tu5,dashed,forget plot]
table[
  col sep=comma,
  x=FPR,
  y=TPR,
  each nth point=10,      
] {figures/data/fabrication/mp_0_5_roc_data.csv};

\addplot[thick,tu3]
table[
  col sep=comma,
  x=FPR,
  y=TPR,
  each nth point=10,      
] {figures/data/fabrication/nl_1_roc_data.csv};

\addplot[thick,tu3,dashed,forget plot]
table[
  col sep=comma,
  x=FPR,
  y=TPR,
  each nth point=10,      
] {figures/data/fabrication/mp_1_roc_data.csv};
\addlegendentry{$\epsilon = 0.25$}
\addlegendentry{$\epsilon = 0.5$}
\addlegendentry{$\epsilon = 1$}

\end{groupplot}

\begin{axis}[
    hide axis,
    scale only axis,
    width=0pt,
    height=0pt,
    at={(0,0)},
    anchor=south west,
    xmin=0, xmax=1,
    ymin=0, ymax=1,
    legend to name=methodlegend2,
    legend style={draw=black,fill=white,font=\small}
]
\addlegendimage{black,thick}
\addlegendentry{Ours (\ref{equ:anomaly_score_pred_e})}
\addlegendimage{black,thick,dashed}
\addlegendentry{Max posterior~\cite{Hendrycks2017}}
\end{axis}

\node at (fgsmLegendPos) [anchor=south east, xshift=3.5pt, yshift=-2pt]
    {\pgfplotslegendfromname{methodlegend2}};

\end{tikzpicture}
    \caption{\textbf{ROC of our proposed regression perturbance detection (\ref{equ:anomaly_score_pred_e}) for object detection}, untargeted~\cite{Chow2020} attack (left), fabrication~\cite{Chow2020} attack (right) at specific perturbation strengths $\epsilon \in \{0.25, 0.5, 1\}$. All results are obtained using a \texttt{Faster R-CNN} \cite{Ren2015} model trained on $\mathcal{D}^\mathrm{VOC}_\mathrm{train}$ and evaluated on perturbed images from $\mathcal{D}^\mathrm{VOC}_\mathrm{val}$.}
\label{fig:ROC_dec}
\end{figure}
We observe that Xu et al.'s multi-pass approach \cite{Xu2018e} shows imbalanced performance across perturbations, being strong under the vanishing attack, but particularly poor under Gaussian noise.
The max posterior baseline \cite{Hendrycks2017} is the overall best among the single-pass baselines. Expectedly, our quantile method (\ref{equ:anomaly_score_reg}) performs poorly due to the small number of objects compared to the high number of pixels in semantic segmentation. \textit{Our regression detector (\ref{equ:anomaly_score_pred_e}), however, is best among all single-pass methods in each perturbation. With a global AuROC of 74.4\% it is overall best, significantly excelling even Xu et al.'s multi-pass approach (69.8\%).}
The AP and detection performance detailed for each strength $\epsilon$ is provided in Supplement \ref{supp:ObjectDetectionRobustness}.

In \cref{fig:ROC_dec}, we display ROC curves for the untargeted~\cite{Chow2020} (left) and fabrication~\cite{Chow2020} (right) adversarial attacks for different attack strengths~$\epsilon$ employing our regression approach (\ref{equ:anomaly_score_pred_e}) and the 
(strong) max posterior baseline \cite{Hendrycks2017}. The results are obtained using the \texttt{Faster R-CNN}~\cite{Ren2015} model, which is trained on $\mathcal{D}^\mathrm{VOC}_\mathrm{train}$ and evaluated on attacked images from $\mathcal{D}^\mathrm{VOC}_\mathrm{val}$. The ROC curves are obtained by varying the decision threshold $\theta$ applied to the anomaly score $\delta(\mathbf{x})$ (\ref{equ:anomaly_score_pred_e}). 
Again, we observe that ROC curves improve with increasing attack strength. For reasonably small FPRs, our regression approach (\ref{equ:anomaly_score_pred_e}) delivers better ROC curves than the strongest baseline max posterior \cite{Hendrycks2017}. Note that further ROC curves are provided in Supplement \ref{supp:ROC_OD}.

In \cref{tab:onehotVsHadamard_det}, we report the clean AP and global AuROC across all perturbations, model size, and computational complexity in terms of GFLOPs of one-hot and our proposed Hadamard output encodings across two object detection architectures and datasets. For perturbation detection with one-hot models, we use the entropy \cite{Smith2018} and max posterior \cite{Hendrycks2017} baselines, while for the \texttt{HadamardNet} we report our three detectors (\ref{equ:anomaly_simple}), (\ref{equ:anomaly_score_pred_e}), (\ref{equ:anomaly_score_reg}).
Models are trained on either $\mathcal{D}^\mathrm{VOC}_\mathrm{train}$ or $\mathcal{D}^\mathrm{BDD,OD}_\mathrm{train}$ and evaluated on $\mathcal{D}^\mathrm{VOC}_\mathrm{val}$ or $\mathcal{D}^\mathrm{BDD,OD}_\mathrm{test}$, respectively. On clean data, we observe close-by AP performance on $\mathcal{D}^\mathrm{BDD,OD}_\mathrm{test}$ for \texttt{Faster R-CNN}, while for any other configuration all methods reach a statistically equivalent AP. Among the baselines, for \texttt{Faster R-CNN} the max posterior perturbation detector \cite{Hendrycks2017} delivers the better global AuROC than the entropy one \cite{Smith2018}, while this is vice versa for \texttt{DETR}. \textit{Overall, our regression perturbation detector (\ref{equ:anomaly_score_pred_e}) in conjunction with the \texttt{HadamardNet} shows the best global AuROC results for both network topologies and on both datasets.} Again, the second-best perturbation detector (\ref{equ:anomaly_simple}) shows good and balanced performance and confirms its well generalizing applicability to both semantic segmentation and object detection on various datasets along with various network topologies.
Detailed object detection results on $\mathcal{D}^\mathrm{BDD,OD}_\mathrm{test}$ are provided in Supplement \ref{supp:additional_results_bdd_od}
and on $\mathcal{D}^\mathrm{COCO}_\mathrm{val}$ in Supplement \ref{supp:large_vocabulary_datasets}.
\begin{table}[t!]
  \centering
  \caption{{One-hot vs.\ Hadamard output encodings on \textbf{object detection} models} on $\mathcal{D}^\mathrm{VOC}_\mathrm{val}$ and $\mathcal{D}^\mathrm{BDD,OD}_\mathrm{test}$. Global AuROC (best in bold, second-best underlined) across all perturbations and strengths $\epsilon\!\in\!\{0.25,0.5,1,2,4,8,16\}$ and AP in (\%).}
  \label{tab:onehotVsHadamard_det}
    
\renewcommand{\arraystretch}{0.9}

\extrarowheight=\aboverulesep
    \addtolength{\extrarowheight}{\belowrulesep}
    \aboverulesep=1pt
    \belowrulesep=-2pt

\begin{tabular}{@{}ll|ccccrr@{}}
\toprule[.9pt]
\multirow{3}{*}{Model} &
Output &
\multicolumn{2}{c}{VOC} &
\multicolumn{2}{c}{BDD} &
\multirow{3}{*}{\makecell{Size\\(M)}} &
\multirow{3}{*}{\makecell{FLOPs\\(G)}} \\[-5pt]
& representation / &
\multicolumn{1}{c}{AP} &
\multicolumn{1}{c}{AuROC} &
\multicolumn{1}{c}{AP} &
\multicolumn{1}{c}{AuROC} &
&
\\[-5pt]
&detector &
clean &
attack &
clean &
attack &
&
\\
\midrule
\multirow{5}{*}{\centering\makecell[l]{\texttt{Faster} \\ \texttt{R-CNN} \cite{Ren2015}}}
& One-hot/entropy~\cite{Smith2018}     &$41.0^{\pm0.4}$  &$59.6^{\pm0.1}$ &$32.8^{\pm0.2}$ &$53.7^{\pm2.3}$  &$41.4$   &$215$   \\[-3pt]
    & One-hot/max post.~\cite{Hendrycks2017}     &$41.0^{\pm0.4}$  &$61.2^{\pm0.1}$ &$32.8^{\pm0.2}$ &${60.5}^{\pm0.4}$  &$41.4$   &$215$   \\[-3pt]
    &{Hadamard/(\ref{equ:anomaly_simple}) (ours)} 
& {$40.4^{\pm0.5}$}  
& {$\underline{63.3}^{\pm0.5}$} 
& {$32.3^{\pm0.3}$} 
& {$\underline{63.2}^{\pm0.4}$}  
& {$41.4$}   
& {$215$}   \\[-3pt]

& Hadamard/(\ref{equ:anomaly_score_pred_e}) (ours) 
& $40.4^{\pm0.5}$  
& $\mathbf{74.4}^{\pm0.4}$ 
& $32.3^{\pm0.3}$ 
& $\mathbf{65.0}^{\pm2.9}$  
& $41.4$   
& $215$   \\[-3pt]


& {Hadamard/(\ref{equ:anomaly_score_reg}) (ours)} 
& {$40.4^{\pm0.5}$}  
& {$39.2^{\pm17}$} 
& {$32.3^{\pm0.3}$} 
& {$57.5^{\pm23}$}  
& {$41.4$}   
& {$215$}   \\
\midrule

\multirow{5}{*}{\centering\makecell[l]{\texttt{DETR} \\ \cite{Carion2020}}}
& One-hot/entropy\ \cite{Smith2018}  
& $47.5^{\pm0.5}$   
& ${66.0}^{\pm3.6}$ 
& $26.7^{\pm0.7}$ 
& $57.2^{\pm2.8}$   
& $41.6$   
& $102$   \\[-3pt]
& One-hot/max post.\ \cite{Hendrycks2017}  
& $47.5^{\pm0.5}$   
& $60.2^{\pm4.0}$ 
& $26.7^{\pm0.7}$ 
& $55.2^{\pm2.4}$   
& $41.6$   
& $102$   \\[-3pt]

& {Hadamard/(\ref{equ:anomaly_simple}) (ours) }
& {$46.9^{\pm0.5}$}   
& {$\mathbf{68.2}^{\pm 3.6}$} 
& {$25.8^{\pm 0.9}$} 
& {$58.3^{\pm4.5}$}  
& {$41.6$}   
& {$102$} \\[-3pt]

& Hadamard/(\ref{equ:anomaly_score_pred_e}) (ours) 
& $46.9^{\pm0.5}$   
& $\underline{68.1}^{\pm 1.9}$ 
& $25.8^{\pm 0.9}$ 
& $\underline{58.4}^{\pm1.3}$  
& $41.6$   
& $102$ \\[-3pt]


& {Hadamard/(\ref{equ:anomaly_score_reg}) (ours) }
& {$46.9^{\pm0.5}$}   
& {$39.2^{\pm6.9}$} 
& {$25.8^{\pm 0.9}$} 
& {$\mathbf{63.6}^{\pm5.8}$}  
& {$41.6$}   
& {$102$} \\

\bottomrule[.9pt]
\end{tabular}

\end{table}

\section{Conclusions}
\label{sec:Conclusion}
Hadamard codes are advantageous output representations for semantic segmentation and object detection tasks, as we have shown in this work that \textit{their redundancy allows for a single-pass state-of-the-art error detection}. 
This is achieved while we preserve an overall equal or close-by task performance w.r.t.\ mIoU and AP metrics on clean data.
We reported results on four network topologies and five datasets, and support our claim by extensive ablations. 
\pagebreak

\section*{Acknowledgments}
The research leading to these results is funded by the German Federal Ministry for Economic Affairs and Energy within the project “Safe AI Engineering – Sicherheitsargumentation befähigendes AI Engineering über den gesamten Lebenszyklus einer KI-Funktion". The authors would like to thank the consortium for the successful cooperation.

%
%
\bibliographystyle{splncs04}
\bibliography{main, sources}

\begin{thebibliography}{10}
\providecommand{\url}[1]{\texttt{#1}}
\providecommand{\urlprefix}{URL }
\providecommand{\doi}[1]{https://doi.org/#1}

\bibitem{akata2013label}
Akata, Z., Perronnin, F., Harchaoui, Z., Schmid, C.: {Label-Embedding for Attribute-Based Classification}. In: Proc. of CVPR. pp. 819--826. Portland, OR, USA (Jun 2013)

\bibitem{Carion2020}
Carion, N., Massa, F., Synnaeve, G., Usunier, N., Kirillov, A., Zagoruyko, S.: {End-to-End Object Detection with Transformers}. In: Proc. of ECCV. pp. 213--229. Glasgow, United Kingdom (Nov 2020)

\bibitem{mmdetection}
Chen, K., Wang, J., Pang, J., Cao, Y., Xiong, Y., Li, X., Sun, S., Feng, W., Liu, Z., Xu, J., Zhang, Z., Cheng, D., Zhu, C., Cheng, T., Zhao, Q., Li, B., Lu, X., Zhu, R., Wu, Y., Dai, J., Wang, J., Shi, J., Ouyang, W., Loy, C.C., Lin, D.: {MMDetection: Open MMLab Detection Toolbox and Benchmark}. arXiv preprint arXiv:1906.07155  (2019)

\bibitem{Chen2018a}
Chen, L.C., Zhu, Y., Papandreou, G., Schroff, F., Adam, H.: {Encoder-Decoder With Atrous Separable Convolution for Semantic Image Segmentation}. In: Proc. of ECCV. pp. 801--818. Munich, Germany (Sep 2018)

\bibitem{Chow2020}
Chow, K.H., Liu, L., Loper, M., Bae, J., Gursoy, M.E., Truex, S., Wei, W., Wu, Y.: {Adversarial Objectness Gradient Attacks in Real-time Object Detection Systems}. In: Proc.\ of TPS-ISA. pp. 263--272. Atlanta, GA, USA (Oct 2020)

\bibitem{mmseg2020}
Contributors, M.: {MMSegmentation}: Openmmlab semantic segmentation toolbox and benchmark. \url{https://github.com/open-mmlab/mmsegmentation} (2020), accessed: 2026-06-25

\bibitem{Cordts2016}
Cordts, M., Omran, M., Ramos, S., Rehfeld, T., Enzweiler, M., Benenson, R., Franke, U., Roth, S., Schiele, B.: {The Cityscapes Dataset for Semantic Urban Scene Understanding}. In: Proc. of CVPR. pp. 3213--3223. Las Vegas, NV, USA (Jun 2016)

\bibitem{DBLP:journals/corr/cs-AI-9501101}
Dietterich, T.G., Bakiri, G.: {Solving Multiclass Learning Problems via Error-Correcting Output Codes}. J. Artif. Intell. Res.  \textbf{2}(1),  263--286 (Jan 1995)

\bibitem{everingham2010pascal}
Everingham, M., Van~Gool, L., Williams, C.K.I., Winn, J., Zisserman, A.: {The PASCAL Visual Object Classes (VOC) Challenge}. International Journal of Computer Vision  \textbf{88}(2),  303--338 (Sep 2010)

\bibitem{Everingham2015}
Everingham, M., Van~Gool, L., Williams, C.K.I., Winn, J., Zisserman, A.: {The Pascal Visual Object Classes Challenge: A Retrospective}. International Journal of Computer Vision (IJCV)  \textbf{111}(1),  98--136 (Jan 2015)

\bibitem{NIPS2013_7cce53cf}
Frome, A., Corrado, G.S., Shlens, J., Bengio, S., Dean, J., Ranzato, M., Mikolov, T.: {DeViSE: A Deep Visual-Semantic Embedding Model}. In: Proc. of NeurIPS. pp. 2121--2129. Lake Tahoe, NV, USA (Dec 2013)

\bibitem{Goodfellow2015}
Goodfellow, I., Shlens, J., Szegedy, C.: {Explaining and Harnessing Adversarial Examples}. In: Proc. of ICLR. pp. 1--10. San Diego, CA, USA (May 2015)

\bibitem{Hendrycks2017}
Hendrycks, D., Gimpel, K.: {A Baseline for Detecting Misclassified and Out-of-Distribution Examples in Neural Networks}. In: Proc. \mbox{of ICLR}. pp. 1--12. Toulon, France (Apr 2017)

\bibitem{Hoffer2018}
Hoffer, E., Hubara, I., Soudry, D.: {Fix Your Classifier: The Marginal Value of Training the Last Weight Layer}. In: Proc. of ICLR - Workshops. pp. 1--11. Vancouver, Canada (Apr 2018)

\bibitem{Hoyos2024}
Hoyos, A., Rivera, M.: {Hadamard Layer for Improve Semantic Segmentation}. IEEE Access  \textbf{12},  194367--194377 (2024)

\bibitem{Hoyos2021}
Hoyos, A., Ruiz, U., Chavez, E.: {Hadamard’s Defense Against Adversarial Examples}. IEEE Access  \textbf{9},  118324--118333 (2021)

\bibitem{NIPS2009_67974233}
Hsu, D.J., Kakade, S.M., Langford, J., Zhang, T.: {Multi-Label Prediction via Compressed Sensing}. In: Proc. of NeurIPS. pp. 772--780. Vancouver, BC, Canada (Dec 2009)

\bibitem{Klingner2020}
Klingner, M., B\"{a}r, A., Fing\-scheidt, T.: {Improved Noise and Attack Robustness for Semantic Segmentation by Using Multi-Task Training with Self-Supervised Depth Estimation}. In: Proc. of CVPR - Workshops. pp. 1299--1309. Seattle, WA, USA (Jun 2020)

\bibitem{Klingner2022a}
Klingner, M., Kumar, V.R., Yogamani, S., Bär, A., Fingscheidt, T.: {Detecting Adversarial Perturbations in Multi-Task Perception}. In: Proc. of IROS. pp. 13050--13057. Kyoto, Japan (Oct 2022)

\bibitem{koenker1978regression}
Koenker, R., Bassett, G.: {Regression Quantiles}. Econometrica  \textbf{46}(1),  33--50 (1978)

\bibitem{kong1995error}
Kong, E.B., Dietterich, T.G.: {Error-Correcting Output Coding Corrects Bias and Variance}. In: Proc. of ICML. pp. 313--321. Tahoe City, CA, USA (Jul 1995)

\bibitem{Kurakin2017}
Kurakin, A., Goodfellow, I., Bengio, S.: {Adversarial Examples in the Physical World}. In: Proc. of ICLR - Workshops. pp. 1--14. Toulon, France (Apr 2017)

\bibitem{Li2020g}
Li, S., Zhu, S., Paul, S., Roy-Chowdhury, A., Song, C., Krishnamurthy, S., Swami, A., Chan, K.S.: {Connecting the Dots: Detecting Adversarial Perturbations Using Context Inconsistency}. In: Proc. of ECCV. pp. 396--413. Glasgow, United Kingdom (Nov 2020)

\bibitem{Lin2014microsoft}
Lin, T.Y., Maire, M., Belongie, S., Hays, J., Perona, P., Ramanan, D., Doll{\'a}r, P., Zitnick, C.L.: {Microsoft COCO: Common Objects in Context}. In: Proc. of ECCV. pp. 740--755. Zurich, Switzerland (Sep 2014)

\bibitem{liu2020energy}
Liu, W., Wang, X., Owens, J., Li, Y.: {Energy-based Out-of-Distribution and Adversarial Detection}. In: Proc.\ of NeurIPS. pp. 21464--21475. virtual (Dec 2020)

\bibitem{Loshchilov2019}
Loshchilov, I., Hutter, F.: {Decoupled Weight Decay Regularization}. In: Proc.\ of ICLR. pp. 1--18. New Orleans, LA, USA (May 2019)

\bibitem{maag2024detectingadversarialattackssemantic}
Maag, K., Resner, R., Fischer, A.: {Detecting Adversarial Attacks in Semantic Segmentation via Uncertainty Estimation: A Deep Analysis}. arXiv:2408.10021  (Aug 2024)

\bibitem{Madry2018}
Madry, A., Makelov, A., Schmidt, L., Tsipras, D., Vladu, A.: {Towards Deep Learning Models Resistant to Adversarial Attacks}. In: Proc. of ICLR. pp. 1--28. Vancouver, BC, Canada (Apr 2018)

\bibitem{Metzen2017}
Metzen, J.H., Kumar, M.C., Brox, T., Fischer, V.: {Universal Adversarial Perturbations Against Semantic Image Segmentation}. In: Proc. of ICCV. pp. 2774--2783. Venice, Italy (Oct 2017)

\bibitem{Michelot1986}
Michelot, C.: {A Finite Algorithm for Finding the Projection of a Point onto the Canonical Simplex of $\alpha^n$}. Journal of Optimization Theory and Applications  \textbf{50}(1),  195--200 (1986)

\bibitem{Nguyen2025}
Nguyen, K.N.T., Zhang, W., Lu, K., Wu, Y.H., Zheng, X., Li~Tan, H., Zhen, L.: {A Survey and Evaluation of Adversarial Attacks in Object Detection}. IEEE Transactions on Neural Networks and Learning Systems  \textbf{36}(9),  15706--15722 (2025)

\bibitem{Proakis1989}
Proakis, J.G.: {Digital Communications}. McGraw-Hill Book Company (1989)

\bibitem{Ren2015}
Ren, S., He, K., Girshick, R., Sun, J.: {Faster R-CNN: Towards Real-Time Object Detection With Region Proposal Networks}. In: Proc. \mbox{of NIPS}. pp. 91--99. Montr\' {e}al, QC, Canada (Dec 2015)

\bibitem{RODRIGUEZ201821}
Rodríguez, P., Bautista, M.A., Gonzàlez, J., Escalera, S.: {Beyond One-Hot Encoding: Lower Dimensional Target Embedding}. Image and Vision Computing  \textbf{75},  21--31 (2018)

\bibitem{Ronneberger2015}
Ronneberger, O., Fischer, P., Brox, T.: {U-Net: Convolutional Networks for Biomedical Image Segmentation}. In: Proc. of MICCAI. pp. 234--241. Munich, Germany (Oct 2015)

\bibitem{10.1007/s10207-023-00735-6}
Ryu, G., Choi, D.: {Detection of Adversarial Attacks Based on Differences in Image Entropy}. International Journal of Information Security  \textbf{23}(1),  299--314 (Feb 2024)

\bibitem{Smith2018}
Smith, L., Gal, Y.: {Understanding Measures of Uncertainty for Adversarial Example Detection}. arXiv preprint:1803.08533  (Mar 2018)

\bibitem{Sylvester01121867}
Sylvester, J.J.: {LX. Thoughts on Inverse Orthogonal Matrices, Simultaneous Sign Successions, and Tessellated Pavements in Two or More Colours, With Applications to Newton's Rule, Ornamental Tile-Work, and the Theory of Numbers}. The London, Edinburgh, and Dublin Philosophical Magazine and Journal of Science  \textbf{34}(232),  461--475 (1867)

\bibitem{Thunuguntla2025}
Thunuguntla, A., Tadepalli, P., Raffa, G.: {Defenses Against Adversarial Attacks on Object Detection: Methods and Future Directions}. Information  \textbf{16}(11) (2025)

\bibitem{NEURIPS2019_cd61a580}
Verma, G., Swami, A.: {Error Correcting Output Codes Improve Probability Estimation and Adversarial Robustness of Deep Neural Networks}. In: Proc. of NeurIPS. pp. 8646--8656. Vancouver, BC, Canada (Dec 2019)

\bibitem{9839178}
Wan, L., Alpcan, T., Viterbo, E., Kuijper, M.: {Efficient Error-Correcting Output Codes for Adversarial Learning Robustness}. In: Proc. of ICC. pp. 2345--2350. Seoul, South Korea (May 2022)

\bibitem{weston2010large}
Weston, J., Bengio, S., Usunier, N.: {Large Scale Image Annotation: Learning to Rank with Joint Word-Image Embeddings}. Machine learning  \textbf{81}(1),  21--35 (2010)

\bibitem{Xie2022segformer}
Xie, E., Wang, W., Yu, Z., Anandkumar, A., Alvarez, J.M., Luo, P.: {SegFormer: Simple and Efficient Design for Semantic Segmentation with Transformers}. In: Proc. of NeurIPS. pp. 12077--12090. virtual (Dec 2021)

\bibitem{Xu2018e}
Xu, W., Evans, D., Qi, Y.: {Feature Squeezing: Detecting Adversarial Examples in Deep Neural Networks}. In: Proc. of Network and Distributed Systems Security Symposium. San Diego, CA, USA (Feb 2018)

\bibitem{yang2015deep}
Yang, S., Luo, P., Loy, C.C., Shum, K.W., Tang, X.: {Deep Representation Learning with Target Coding}. In: Proc.\ of AAAI. pp. 3848--3854. Austin, TX, USA (Jan 2015)

\bibitem{Yin2021}
Yin, M., Li, S., Cai, Z., Song, C., Asif, M.S., Roy-Chowdhury, A.K., Krishnamurthy, S.V.: {Exploiting Multi-Object Relationships for Detecting Adversarial Attacks in Complex Scenes}. In: Proc. of ICCV. pp. 7858--7867. Montréal, QC, Canada (Oct 2021)

\bibitem{Yu2019}
Yu, F., Chen, H., Wang, X., Xian, W., Chen, Y., Liu, F., Madhavan, V., Darrell, T.: {BDD100K: A Diverse Driving Dataset for Heterogeneous Multitask Learning}. In: Proc. of CVPR. pp. 1--14. Seattle, WA, USA (Jun 2020)

\bibitem{Zhou2017}
Zhou, B., Zhao, H., Puig, X., Fidler, S., Barriuso, A., Torralba, A.: {Scene Parsing through ADE20K Dataset}. In: Proc. of CVPR. pp. 633--641. Honolulu, HI, USA (Jul 2017)

\end{thebibliography}
\newpage
\appendix
\renewcommand*{\theHsection}{appendix.\Alph{section}}
\section{Hadamard Code Construction and Example}
\label{supp:HadamardConstructionExample}
Hadamard codes are generated using Sylvester’s construction \cite{Sylvester01121867} of the bipolar Hadamard matrix
\begin{equation} \check{\mathbf{H}}^{(L)} = \left( {\begin{array}{cc} \check{\mathbf{H}}^{(L/2)} & \check{\mathbf{H}}^{(L/2)} \\ \check{\mathbf{H}}^{(L/2)} & -\check{\mathbf{H}}^{(L/2)} \\ \end{array} } \right) \in \{-1, 1\}^{L\times L}, \end{equation}
with the base case $\check{\mathbf{H}}^{(1)} = 1$.
Accordingly, a Hadamard matrix contains binary elements (bits). An example of the Hadamard matrix for $L = 4$ in bipolar notation is given as
\begin{equation} \check{\mathbf{H}}^{(4)} = \left( {\begin{array}{rrrr} 1 & 1 & 1 & 1 \\ 1 & -1 & 1 & -1 \\ 1 & 1 & -1 & -1 \\ 1 & -1 & -1 & 1 \\ \end{array} } \right) = (\check{\mathbf{H}}^{(4)}_1 \ \check{\mathbf{H}}^{(4)}_2 \ \check{\mathbf{H}}^{(4)}_3 \ \check{\mathbf{H}}^{(4)}_4). \end{equation}
Each column corresponds to a Hadamard codeword $\check{\mathbf{H}}^{(L)}_s$ representing class $s \in \mathcal{S}=\{1,\dots,S\}$ with $S=4$ classes, where each codeword differs in exactly $L/2$ bits from any other codeword.

\section{Class Posterior Constraints and General Optimization Problem}
\label{supp:classProsteriorConstraintsandGeneralOptimizationProblem}
In this section, using a solely probabilistic approach, we show that an error-free soft codeword $\mathbf{h}_i^*$ is always a linear combination of legacy codewords $\textbf{H}^{(L)}_s$ (\ref{equ:hadamard_codeword_conversion}) with $s \in \mathcal{S}$, where the coefficients $\PROB_{i,s}$ fulfill the constraints (\ref{equ:nebenbedingung1}) and (\ref{equ:nebenbedingung2}).

In the error-free soft codeword $\mathbf{h}_i^* = (h^*_{i,\ell})$, at each bit position $\ell$, the introduced \texttt{HadamardNet} performs a binary prediction between a positive class subset 
\begin{equation}
\label{equ:positive_subset}
\mathcal{S}^{+}_\ell = \{s \in \mathcal{S}|H^{(L)}_{s,\ell} = 1\},
\end{equation}
that includes all classes for which the value at bit position $\ell$ in the ground truth codeword is $H^{(L)}_{s,l} = 1$, and a negative subset
\begin{equation}
\label{equ:negative_subset}
\mathcal{S}^{-}_\ell = \{s \in \mathcal{S}|H^{(L)}_{s,\ell} = 0\},
\end{equation}
that includes all classes for which the value at bit position $\ell$ is $H^{(L)}_{s,\ell}  = 0$.
Both subsets, $\mathcal{S}^{+}_\ell$ and $\mathcal{S}^{-}_\ell$, are disjoint for any bit position $\ell$, yielding
\begin{equation}
\mathcal{S}^{-}_\ell \cap \mathcal{S}^{+}_\ell = \emptyset,
\end{equation}
and
\begin{equation}
\label{equ:disjoint_and}
\mathcal{S}^{-}_\ell \cup \mathcal{S}^{+}_\ell = \mathcal{S}.
\end{equation}

Using (\ref{equ:positive_subset}) and (\ref{equ:negative_subset}), we can now define the following constraints for the desired class posterior probabilities $\mathbf{P}_i$ of the \texttt{HadamardNet}. According to the addition rule for mutually exclusive events, the probabilities of all classes in the positive subset $s \in \mathcal{S}^{+}_\ell$ for the bit position $\ell$ have to sum up to the predicted error-free soft codeword entry
\begin{equation}
\label{equ:LGS_sum_pos}
h^*_{i,\ell} = \PROB_i(s \in \mathcal{S}_\ell^+|\mathbf{x}) = \sum_{s \in \mathcal{S}^{+}_\ell} \PROB_{i,s}
\end{equation}
at that bit position $\ell$. The same holds for the negative subset, where the probabilities of all classes in the negative subset $s \in \mathcal{S}^{-}_\ell$ for bit position $\ell$ have to sum up to the reversed predicted probability
\begin{equation}
\label{equ:LGS_sum_neg}
1 - h^*_{i,\ell} = \PROB_i(s \in \mathcal{S}_\ell^-|\mathbf{x}) = \sum_{s \in \mathcal{S}^{-}_\ell} \PROB_{i,s}
\end{equation}
at that bit position $\ell$. 
By adding a zero to (\ref{equ:LGS_sum_pos}), we obtain
\begin{equation}
\label{equ:LGS_sum_pos_zero}
h^*_{i,\ell} = \sum_{s \in \mathcal{S}^{+}_\ell} \PROB_{i,s} = 1 \cdot \sum_{s \in \mathcal{S}^{+}_\ell} \PROB_{i,s} + 0 \cdot \sum_{s \in \mathcal{S}^{-}_\ell} \PROB_{i,s}.
\end{equation}
Employing (\ref{equ:positive_subset}) and (\ref{equ:negative_subset}), (\ref{equ:LGS_sum_pos_zero}) becomes
\begin{equation}
\label{equ:sum_pre_LGS}
    h^*_{i,\ell} =  \sum_{s \in \mathcal{S}^{+}_\ell} H^{(L)}_{s,\ell} \cdot \PROB_{i,s} + \sum_{s \in \mathcal{S}^{-}_\ell} H^{(L)}_{s,\ell} \cdot \PROB_{i,s}.
\end{equation}
Finally, using (\ref{equ:disjoint_and}) and (\ref{equ:sum_pre_LGS}), yields
\begin{equation}
\label{equ:sum_pre_LGS_H}
    h^*_{i,\ell} = \sum_{s \in \mathcal{S}} H^{(L)}_{s,\ell} \cdot \PROB_{i,s},
\end{equation}
which proves that an error-free soft-codeword $\mathbf{h}_i^*$ is always a linear combination 
\begin{equation}
    \mathbf{h}_i^* = \sum_{s\in \mathcal{S}} \mathbf{H}^{(L)}_s\cdot \PROB_{i,s} = \mathbf{H} \cdot\mathbf{P}_i
\end{equation}
of legacy codewords $\textbf{H}^{(L)}_s$ with coefficients $\PROB_{i,s}$ fulfilling the constraints (\ref{equ:nebenbedingung1}) and (\ref{equ:nebenbedingung2}).

\section{Rearrangement of LES}
\label{supp:OptimizationProblemGeneral}
Here, we provide a step-by-step rearrangement for $\textbf{P}_i$ of our linear equation system (LES) (\ref{equ:LGS}).
Using (\ref{equ:LGS}) and (\ref{equ:H_Hprime}), we obtain
\begin{equation}
\frac{1}{2}(\check{\mathbf{H}} +\textbf{1}^{(L\times S)}) \cdot \textbf{P}_i = \textbf{h}_i + \check{\mathbf{e}}_i,
\end{equation}
\begin{equation}
\label{equ:5in20}
\frac{1}{2}(\check{\mathbf{H}} \cdot \textbf{P}_i+\textbf{1}^{(L\times S)}\cdot \textbf{P}_i) = \textbf{h}_i + \check{\mathbf{e}}_i.
\end{equation}
Using the constraint (\ref{equ:nebenbedingung1}), we can simplify (\ref{equ:5in20}) as follows
\begin{equation}
\frac{1}{2}(\check{\mathbf{H}} \cdot \textbf{P}_i + \textbf{1}^{(L)}) = \textbf{h}_i + \check{\mathbf{e}}_i.
\end{equation}
Multiplication by $2$ and substraction of $\mathbf{1}^{(L)}$ yields
\begin{equation}
\check{\mathbf{H}} \cdot \textbf{P}_i = 2\textbf{h}_i - \textbf{1}^{(L)} + 2\check{\mathbf{e}}_i.
\end{equation}
In analogy to (\ref{equ:H_Hprime}), we have $\check{\mathbf{h}}_i = 2\textbf{h}_i - \textbf{1}^{(L)} \in [-1,1]^L$, and accordingly we obtain
\begin{equation}
\label{equ:-11lgs}
\check{\mathbf{H}} \cdot \textbf{P}_i = \check{\mathbf{h}}_i+ 2\check{\mathbf{e}}_i,
\end{equation}
which is similar to the original linear equation system in (\ref{equ:LGS}), but now uses the Hadamard representation $\check{\mathbf{H}}$ (\ref{equ:hadamardForClassification-11}) with bipolar entries.
Multiplying both sides in (\ref{equ:-11lgs}) by $1/L\cdot\check{\mathbf{H}}^\mathsf{T}$, according to (\ref{equ:pseudo_inverse}), we obtain
\begin{equation}
\textbf{P}_i = \frac{1}{L}\check{\mathbf{H}}^\mathsf{T}\check{\mathbf{h}}_i+ \frac{2}{L}\check{\mathbf{H}}^\mathsf{T}\check{\mathbf{e}}_i.
\end{equation}
Defining $\tilde{\textbf{P}}_i = \frac{1}{L}\check{\mathbf{H}}^\mathsf{T}\check{\mathbf{h}}_i \in [-1,1]^S$ and $\mathbf{e}_i = \frac{2}{L}\check{\mathbf{H}}^\mathsf{T}\check{\mathbf{e}}_i = \frac{2}{L}(\check{\mathbf{e}}^{\mathsf{T}}_i \check{\mathbf{H}})^{\mathsf{T}} \in \mathbb{R}^S$, we obtain
\begin{equation}
\label{equ:u_i2}
\mathbf{P}_i = \tilde{\mathbf{P}}_i + \mathbf{e}_i,
\end{equation}
which is the rearranged system of linear equations (\ref{equ:LGS}) we aim to solve.

\section{Equality of Minimizing Error Vectors}
\label{supp:EqualityErrorVector}
In this section, we show that $\arg\min_{\mathbf{P}_i \in \mathcal{P}} \|\mathbf{e}_i\|_2^2 
=\arg\min_{\mathbf{P}_i \in \mathcal{P}} \|\check{\mathbf{e}}_i\|_2^2$ and therefore minimizing $||\check{\mathbf{e}}_i||_2^2$ in (\ref{equ:LGS}) is equivalent to minimizing $||\mathbf{e}_i||_2^2$ in (\ref{equ:u_i}).
We use (\ref{equ:-11lgs}) to obtain
\begin{align}
\label{equ:norm_check_e_i}
    \|\check{\mathbf{e}}_i\|_2^2 = \frac{1}{4}\|\check{\mathbf{H}}\mathbf{P}_i-\check{\mathbf{h}}_i\|_2^2&=\frac{1}{4}(\mathbf{P}_i^\mathsf{T}\check{\mathbf{H}}^\mathsf{T}\check{\mathbf{H}}\mathbf{P}_i-2\mathbf{P}_i^\mathsf{T}\check{\mathbf{H}}^\mathsf{T}\check{\mathbf{h}}_i+\check{\mathbf{h}}_i^\mathsf{T}\check{\mathbf{h}}_i) \notag\\
    &=\frac{L}{4}\|\mathbf{P}_i\|_2^2-\frac{L}{2}\mathbf{P}_i^\mathsf{T}\tilde{\mathbf{P}}_i+\frac{1}{4}\check{\mathbf{h}}_i^\mathsf{T}\check{\mathbf{h}}_i.
\end{align}
Using (\ref{equ:u_i}), we obtain the norm
\begin{equation}
\label{equ:norm_e_i}
    \|\mathbf{e}_i\|_2^2 = \|\mathbf{P}_i - \tilde{\mathbf{P}}_i\|_2^2=\|\mathbf{P}_i\|_2^2-2\mathbf{P}_i^\mathsf{T}\tilde{\mathbf{P}}_i + \|\tilde{\mathbf{P}}_i\|_2^2.
\end{equation}
Using (\ref{equ:norm_check_e_i}) and (\ref{equ:norm_e_i}), and exploiting that both $\tilde{\mathbf{P}}_i$ and $\check{\mathbf{h}}_i$ are constant w.r.t.\ minimization over $\mathbf{P}_i$, we find
\begin{flalign}
\arg\min_{\mathbf{P}_i \in \mathcal{P}} \|\mathbf{e}_i\|_2^2  \notag
&=\arg\min_{\mathbf{P}_i \in \mathcal{P}}(\|\mathbf{P}_i\|_2^2-2\mathbf{P}_i^\mathsf{T}\tilde{\mathbf{P}}_i + \|\tilde{\mathbf{P}}_i\|_2^2) \notag \\
&=\arg\min_{\mathbf{P}_i \in \mathcal{P}}(\|\mathbf{P}_i\|_2^2-2\mathbf{P}_i^\mathsf{T}\tilde{\mathbf{P}}_i + \frac{1}{L}\check{\mathbf{h}}_i^\mathsf{T}\check{\mathbf{h}}_i) \notag \\
&=\arg\min_{\mathbf{P}_i \in \mathcal{P}}(\frac{L}{4}\|\mathbf{P}_i\|_2^2-\frac{L}{2}\mathbf{P}_i^\mathsf{T}\tilde{\mathbf{P}}_i+\frac{1}{4}\check{\mathbf{h}}_i^\mathsf{T}\check{\mathbf{h}}_i) \notag \\
&=\arg\min_{\mathbf{P}_i \in \mathcal{P}} \|\check{\mathbf{e}}_i\|_2^2
\end{flalign}
Accordingly, minimizing $||\mathbf{e}_i||_2^2$ in (\ref{equ:u_i}) also minimizes $||\check{\mathbf{e}}_i||_2^2$ in (\ref{equ:LGS}) and therefore solves our optimization problem.

\section{Solution and Error of the Optimization Problem}
\label{supp:SimplexProjection}
In the following, we describe the projection onto the probability simplex
\begin{equation}
    \textbf{P}^*_i=\boldsymbol{\phi}(\tilde{\textbf{P}}_i)
\end{equation}
by Michelot \cite{Michelot1986} in detail.
First, given the vector $\tilde{\mathbf{P}}_{i}$, we create a sorted vector $\tilde{\mathbf{P}}^{\prime}_i = (\tilde{\PROB}_{i,s=\pi(s^\prime)}) \in  [-1,1]^{S}$, with $s,s^\prime \in \mathcal{S}$, where the sorting permutation $\pi$ satisfies 
\begin{equation}
    \quad \tilde{\PROB}_{i,\pi(1)} \geq \tilde{\PROB}_{i,\pi(2)} \geq \dots \geq \tilde{\PROB}_{i,\pi(S)}.
\end{equation}
Next, we compute an index
\begin{equation}
    \rho = \max \bigg[  k \in \mathcal{S} \mid \tilde{\PROB}_{i,\pi(k)} - \frac{1}{k} (\sum_{j\in \mathcal{K}} \tilde{\PROB}_{i,\pi(j)}-1 ) > 0 \bigg],
\end{equation}
with $1 \leq \rho \leq S$ and $\mathcal{K} = \{1,\dots, k\}$.
Then, we obtain a threshold
\begin{equation}
    \theta = \frac{1}{\rho} ( \sum_{m \in \mathcal{M}} \tilde{\PROB}_{i,\pi(m)}-1 ),
\end{equation}
with $\mathcal{M} = \{1,\dots,\rho\}$.
Finally, the projection of $\tilde{\mathbf{P}}_{i}$ onto the simplex $\mathbf{P}_i^{*}$ is
\begin{equation}
    \PROB_{i,s}^{*} = \max\left(\tilde{\PROB}_{i,s} - \theta, 0\right), \quad \forall s \in \mathcal{S}.
\end{equation}
Substituting $\mathbf{P}_i^{*} = \mathbf{P}_i$ in (\ref{equ:u_i}) yields the smallest class-wise error vector
\begin{equation}
\label{equ:e_i^*2}
\mathbf{e}_i^{*} = \mathbf{P}_i^{*} - \tilde{\mathbf{P}}_i \in \mathbb{R}^S\;.
\end{equation}

\section{Relationship of Posterior Probabilities and Error Vector}
\label{supp:correlation_e_p}
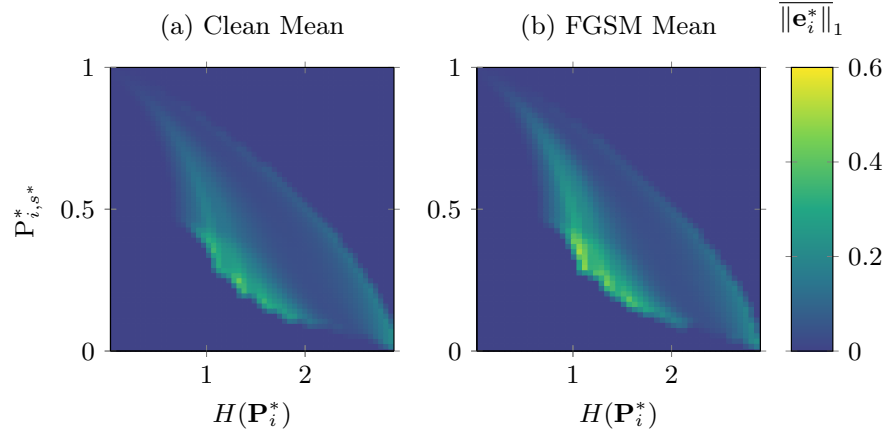
\begin{figure}[t!]
    \centering
    \resizebox{1.0\linewidth}{!}{
\pgfplotsset{
    colormap={light viridis}{
        rgb255(0cm)=(64,67,135)
        rgb255(1cm)=(42,120,142)
        rgb255(2cm)=(34,167,132)
        rgb255(3cm)=(122,209,81)
        rgb255(4cm)=(253,231,37)
    }
}

\begin{tikzpicture}

\begin{groupplot}[
    group style={
        group size=2 by 1,
        horizontal sep=1cm
    },
    width=5cm,
    height=5cm,
    view={0}{90},
    xmin=0.02567364, xmax=2.901121347,
    ymin=0, ymax=1,
    enlargelimits=false,
    axis on top,
    tick align=center,
    major tick length=4pt,
    colormap name=light viridis,
    point meta min=0,
    point meta max=0.6,
]

\nextgroupplot[
    title={(a) Clean Mean},
    xlabel={{${H}(\mathbf{P}^*_i)$}},
    ylabel={{${\PROB}_{i,s^*}^*$}},
]

\addplot3[
    surf,
    shader=flat,
    draw=none,
    mesh/cols=57,
    point meta=explicit,
] table[
    col sep=comma,
    x=x, y=y,
    z expr=0,
    meta=z
] {figures/heatmap_mean.csv};



\nextgroupplot[
    title={(b) FGSM Mean},
    xlabel={{${H}(\mathbf{P}^*_i)$}},
    colorbar,
    colorbar style={
        title={$\overline{\|\mathbf{e}_i^*\|}_1$},
        title style={at={(0.5,1.02)},anchor=south}
    },
]

\addplot3[
    surf,
    shader=flat,
    draw=none,
    mesh/cols=54,
    point meta=explicit,
] table[
    col sep=comma,
    x=x, y=y,
    z expr=0,
    meta=z
] {figures/heatmap_fgsm_mean.csv};

\end{groupplot}
\end{tikzpicture}
    }
    \caption{Two-dimensional histograms over entropy ${H}(\mathbf{P}^*_i)$ and maximum posterior probability ${\PROB}_{i,s^*}^*$. (a) Mean $\overline{\|\mathbf{e}_i^*\|}_1$ on clean data from $\mathcal{D}^\mathrm{CS}_\mathrm{train}$. (b) Mean $\overline{\|\mathbf{e}_i^*\|}_1$ on FGSM-attacked data from $\mathcal{D}^\mathrm{CS}_\mathrm{train}$ with $\epsilon=8$.}
\label{fig:CORR_analysis}
\end{figure}

In \cref{fig:CORR_analysis}, we illustrate the relationship between the entropy ${H}(\mathbf{P}^*_i)$, the maximum posterior probability $\PROB_{i,s^*}^*$, and the $L_1$ norm of the error vector $\lVert \mathbf{e}_i^* \rVert_1$ using a two-dimensional histogram. Panel (a) shows the error vector mean $\overline{\lVert \mathbf{e}_i^* \rVert}_1$ on clean data from  $\mathcal{D}^\mathrm{CS}_\mathrm{train}$ and panel (b) presents the corresponding mean under FGSM-attacked samples from $\mathcal{D}^\mathrm{CS}_\mathrm{train}$.
We observe that, for both clean and attacked data, $\overline{\lVert \mathbf{e}_i^* \rVert}_1$ is not uniformly distributed but varies with the entropy ${H}(\mathbf{P}^*_i)$ and the maximum posterior probability $\PROB_{i,s^*}^*$. For attacked samples, the relationship between ${H}(\mathbf{P}^*_i)$, $\PROB_{i,s^*}^*$, and $\overline{\lVert \mathbf{e}_i^* \rVert}_1$ largely remains the same, but for each bin (${H}(\mathbf{P}^*_i)$, $\PROB_{i,s^*}^*$) $\overline{\lVert \mathbf{e}_i^* \rVert}_1$ is increased. This indicates that directly applying the error vector anomaly score (\ref{equ:anomaly_simple}) would lead to many false alarms on clean images with many low-confidence pixels. Instead, a more reliable signal is the relative increase of the mean error vector $\overline{\lVert \mathbf{e}_i^* \rVert}_1$ for pixels or objects $i$ with similar entropy ${H}(\mathbf{P}^*_i)$ and maximum posterior probability $\PROB_{i,s^*}^*$ as used in (\ref{equ:anomaly_score_pred_e}) and (\ref{equ:anomaly_score_reg}).

\section{Details on Regression Model $\mathcal{F}$}
\label{supp:details_F_reg}
\begin{figure}[t!!]
    \centering
\tikzset{%
  block/.style    = {draw, thick, rectangle, fill=white,
    minimum height = 2.3cm, minimum width = 0.5cm, align=center},
  myptr/.style={-{Triangle[length=4.1pt,width=3.7pt]}, line width=1pt},
}

\begin{tikzpicture}[xshift=-3mm]
  \coordinate (origin) at (0,0);

  \node [left = of origin] (post) {$(\mathbf{P}_i^*)$};
  \node [left = of origin, yshift=-2.4cm] (error) {$(\mathbf{e}_i^*)$};


  \node [right = -0.2cm of origin, rectangle, fill=white, draw, minimum height=0.5cm, minimum width = 1.6cm, anchor = west] (entropy) {Entropy};

  \node [right = -0.2cm of origin, rectangle, fill=white, draw, minimum height=0.5cm, minimum width = 1.6cm, anchor = west, yshift=-0.8cm, align=center] (max) {Max};

  \node [right = of entropy.east, xshift = -1.05cm, anchor = south west] (h) {$(H(\mathbf{P}_i^*))$};
  \node [right = of max.east, xshift = -1.05cm, anchor = south west] (p_max) {$({P}_{i,s^*}^*)$};

  \coordinate (split_p) at ($ (post.east) + (0.35cm, 0)$);
  \filldraw (split_p) circle (2pt);  

    \draw [myptr] (post) -- (entropy);
    \draw [myptr] (split_p) |- (max);

    \node [right = -0.2cm of origin, rectangle, fill=white, draw, minimum height=0.5cm, minimum width = 1.6cm, anchor = west, align=center, yshift = -2.4cm] (L1) {$L_1$-Norm};

    \node [right = of L1.east, xshift = -1.05cm, anchor = south west] (e) {$(\|\mathbf{e}^*_i\|_1)$};

  \node [block, right = of entropy, xshift=0.6cm, yshift=-0.4cm] (lin1) {\rotatebox{90}{FC(32)}};
  \node [block, right = 0.05cm of lin1] (bn1) {\rotatebox{90}{BatchNorm}};
  \node [block, right = 0.05cm of bn1] (relu1) {\rotatebox{90}{ReLU}};

  \node [block, right = 0.05cm of relu1] (lin2) {\rotatebox{90}{FC(16)}};
  \node [block, right = 0.05cm of lin2] (bn2) {\rotatebox{90}{BatchNorm}};
  \node [block, right = 0.05cm of bn2] (relu2) {\rotatebox{90}{ReLU}};

  \node [block, right = 0.05cm of relu2] (lin3) {\rotatebox{90}{FC(1)}};
  
  \node [right = 0.75cm of lin3.east, anchor=west] (final) {};
  \node at ($(lin3.east)+(0.6cm,0.3cm)$) {$(z_i)$};

  \coordinate (lin1_top) at ($ (lin1.west) + (0.0cm, 0.4cm)$);
  \coordinate (lin1_bottom) at ($ (lin1.west) - (0.0cm, 0.4cm)$);

  Arrows: strictly left-to-right
  \draw [myptr] (entropy) -- (lin1_top);
  \draw [myptr] (max) -- (lin1_bottom);

    \node [right = 0.2cm of final, rectangle, fill=white, draw, minimum height=0.5cm,  anchor = west, align=center, yshift=-1.0cm] (anomaly) {(\ref{equ:anomaly_score_pred_e})/\\(\ref{equ:anomaly_score_reg})};

    \coordinate (an_top) at ($ (anomaly.west) + (0.0cm, 0.2cm)$);
  \coordinate (an_bottom) at ($ (anomaly.west) - (0.0cm, 0.2cm)$);
  \coordinate (an_b_bottom) at ($(an_bottom) - (0.3cm, 0cm)$);
  \coordinate (an_b_top) at ($(an_top) - (0.3cm, 0cm)$);

    \draw [myptr] (error) -- (L1);
  \draw [myptr] (L1) -| (an_b_bottom) -- (an_bottom);
  \draw [myptr] (lin3.east) -| (an_b_top) -- (an_top);

  \node [right = 0.05cm of anomaly.east, anchor=west, yshift=0.25cm] (anomaly_score) {$\delta(\mathbf{x})$};

  \draw [myptr] (anomaly) -- ($(anomaly.east) + (0.7cm,0)$);

\coordinate (bg_west)   at ($(entropy.west)+(-6mm,0)$);                 
\coordinate (bg_east)   at ($(anomaly.east)+(1.5mm,0)$);           
\coordinate (bg_top)    at ($(lin1.north)+(0,8mm)$);                    
\coordinate (bg_bottom) at ($(anomaly.south)+(0,-10mm)$);                

\begin{scope}[on background layer]
  \path[fill=tu102, draw=none]
    ($(bg_west |- bg_top)$) rectangle ($(bg_east |- bg_bottom)$);
\end{scope}

\node[
    anchor=north west,
    inner sep=2mm
] at ($(bg_west |- bg_top) + (-1mm, 0)$) {Perturbation Detection};

\node[font=\small, anchor=south] (reg_title)
  at ($(lin1.north)!0.5!(lin3.north)+(0mm,0.3mm)$)
  {Regression Model $\mathcal{F}$};

\coordinate (fc_west)  at ($(lin1.west)+(-3mm,0)$);
\coordinate (fc_east)  at ($(lin3.east)+(3mm,0)$);
\coordinate (fc_top)   at ($(reg_title.north)+(0,0.5mm)$);   
\coordinate (fc_bottom)at ($(lin1.south)+(0,-3mm)$);

\begin{scope}[on background layer]
  \path[fill=tu104]
    ($(fc_west |- fc_top)$) rectangle ($(fc_east |- fc_bottom)$);
\end{scope}

\end{tikzpicture}
    \caption{\textbf{Block diagram} for perturbation detection in \cref{fig:inference} and regression model $\mathcal{F}$. For the regression anomaly score $\delta(\mathbf{x})$ (\ref{equ:anomaly_score_pred_e}) we have $z_i=\widehat{\lVert \mathbf{e}_i^* \rVert}_1$ and for the quantile anomaly score $\delta(\mathbf{x})$ (\ref{equ:anomaly_score_reg}) we have $z_i = E_i^{(0.85)}$.}
    \label{fig:regression_model}
\end{figure}
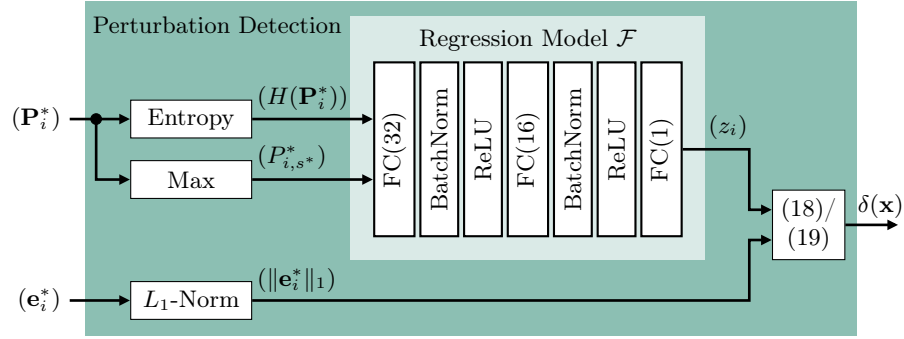

In \cref{fig:regression_model}, we illustrate the architecture of the regression model $\mathcal{F}$ used for perturbation detection with both the regression anomaly score~(\ref{equ:anomaly_score_pred_e}) and the quantile anomaly score~(\ref{equ:anomaly_score_reg}). The inputs to $\mathcal{F}$ are the entropy of the predicted posterior distribution $H(\mathbf{P}^*_i)=-\sum_{s \in \mathcal{S}} \PROB_{i,s}^*\cdot\log(\PROB_{i,s}^*) \ge 0$ and the maximum posterior probability $P^*_{i,s^*}$ for each pixel or object $i \in \mathcal{I}$. These two features are processed by two consecutive fully connected (FC) layers with 32 and 16 neurons, respectively, each followed by batch normalization and ReLU activation. A final fully connected layer produces the output $z_i \in \mathbb{R}$.
For the regression anomaly score $\delta(\mathbf{x})$ (\ref{equ:anomaly_score_pred_e}), the output is $z_i = \widehat{\|\mathbf{e}^*_i\|}_1$, representing the predicted $L_1$ norm of the error vector. In contrast, for the quantile anomaly score $\delta(\mathbf{x})$ (\ref{equ:anomaly_score_reg}), the output is $z_i = E_i^{(0.85)}$ corresponding to the conditional $85\%$ quantile of $\|\mathbf{e}_i^*\|_1$ over all pixels or objects. The predicted value $z_i$ from the regression network $\mathcal{F}$ and the corresponding $L_1$ norm $\|\mathbf{e}_i^*\|_1$ obtained from \texttt{HadamardNet} are then used in~(\ref{equ:anomaly_score_pred_e}) or~(\ref{equ:anomaly_score_reg}) to compute the image-level anomaly score $\delta(\mathbf{x}) \ge 0$.

\section{Details on \texttt{HadamardNet} Training}
\label{supp:details_hadamardnet_training}
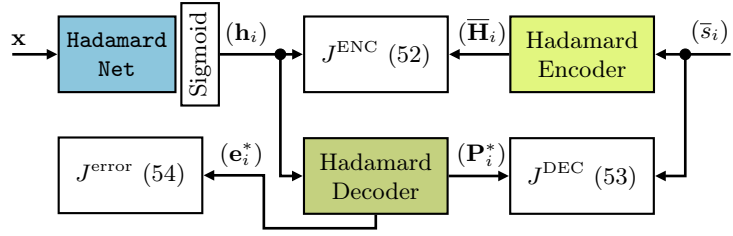
\begin{figure}[t!]
    \centering
    \tikzset{%
  block/.style    = {draw, thick, rectangle, fill=white, minimum height = 1.8em,
    minimum width = 3.5em},
  T_box/.style = {draw, thick, fill=white, rectangle, minimum height = 1.8em, minimum width = 1.8em},
  sum/.style      = {draw, circle,fill=white, inner sep=-2pt, node distance = 1.5cm}, 
  mult/.style      = {draw, circle, fill=white, inner sep= -0.5pt, node distance = 1.5cm}, 
  input/.style    = {coordinate}, 
  output/.style   = {coordinate}, 
  myptr/.style={-{Triangle[length=4.1pt,width=3.7pt]}, line width=1pt},
  myptr_dotted/.style={-{Triangle[length=4.1pt,width=3.7pt]}, line width=1pt, dashed},
}
\begin{tikzpicture}
  \coordinate (origin) at (0,0) {} {} {} {};
  
 \node [right = 0.3cm of origin, yshift=0.2 cm, anchor = east] (input)  {$\mathbf{x}$};

 \node [block, fill=tu52, draw, right = of origin, xshift=-0.4cm, anchor=west, align=center, minimum height=1cm] (Hadamardnet) {\texttt{Hadamard} \\ \texttt{Net}};
 \node [block, rotate=90, fill=white, draw, right = of Hadamardnet, yshift=0.95cm, xshift=0.0cm, anchor=north, align=center, minimum height = 0cm] (sigmoid) {Sigmoid};
 
 \node [block, fill=white, draw, right = of sigmoid.south, xshift = 0.1cm, anchor=west, align=center, minimum height = 1cm, minimum width = 1.9cm] (enc_loss) {$J^{\mathrm{ENC}}$~(\ref{equ:enc_loss_supp})};
 
  \node [block, fill=tu92, draw, right = of sigmoid, xshift = 0.36cm, yshift = -2.3cm, anchor=west, align=center, minimum height = 1cm, minimum width = 1.9cm] (had_dec) {Hadamard \\ Decoder};
  
 \node [block, fill=tu82, draw, right = of enc_loss, xshift = -0.2cm, yshift = 0cm, anchor=west, align=center, minimum height = 1cm, minimum width = 1.9cm] (had_enc) {Hadamard \\ Encoder};
 
  \node [block, fill=white, draw, right = of had_dec, xshift = -0.2cm, anchor=west, align=center, minimum height = 1cm, minimum width = 1.9cm] (dec_loss) {$J^{\mathrm{DEC}}$~(\ref{equ:dec_loss})};
  
  \node [block, fill=white, draw, below = of Hadamardnet.west, yshift = -0.6cm,  anchor=west, align=center, minimum height = 1cm, minimum width = 1.9cm] (err_loss) {$J^{\mathrm{error}}$~(\ref{equ:error_loss})};
  
  \coordinate (gt) at ($ (had_enc.east) + (1.0cm, 0)$);
  
  \coordinate (err_step) at ($ (err_loss.east) + (0.8cm, -0.7cm)$);
  
   \node [left = -0.1cm of gt, yshift=0.3cm, anchor = east] (label)  {$(\overline{s}_i)$};
   
   \coordinate (split_h) at ($ (sigmoid.south) + (0.8cm, 0)$);
   \filldraw (split_h) circle (2pt);  
  
  \coordinate (split_v) at ($ (gt) - (0.6cm, 0)$);
   \filldraw (split_v) circle (2pt);  
  
 \draw [myptr] (origin) -- (Hadamardnet);
 
 \draw [myptr] (sigmoid) -- (enc_loss);
 
 \draw [myptr] (had_enc) -- (enc_loss);
 
 \draw [myptr] (gt) -- (had_enc);
 
 \draw [myptr] (had_dec) -- (dec_loss);
 
 \draw [myptr] (split_h) |- (had_dec);
 
 \draw [myptr] (split_v) |- (dec_loss);
 
 \draw [myptr] (had_dec) |- (err_step) |- (err_loss);
 
    \node [right = of sigmoid.south, xshift = -1.05cm, anchor = south west] (h) {$(\mathbf{h}_i)$};
    
    \node [left = of had_enc.west, xshift = 0.2cm, anchor = south west] (H_target) {$(\overline{\mathbf{H}}_i)$};
    
     \node [right = of had_dec.east, xshift = -1.01cm, anchor = south west] (P) {$(\mathbf{P}^*_i)$};
 
  \node [right = of err_loss.east, xshift = -0.9cm, anchor = south west] (h) {$(\mathbf{e}^*_i)$};

\end{tikzpicture}
    \caption{\textbf{Training setup} for our \texttt{HadamardNet}.}
    \label{fig:training}
\end{figure}
While our final \texttt{HadamardNet} mainly relies on the encoded-space loss $J^{\mathrm{ENC}}$, we additionally investigate further training objectives in our ablation study in Supplement~\ref{supp:SemanticSegmentationAblation}. As shown in \cref{fig:training}, the \texttt{HadamardNet} can be trained by minimizing
\begin{equation}
\label{equ:overall_loss_supp}
    J^{\mathrm{cls}} = \alpha J^{\mathrm{ENC}} + \beta J^{\mathrm{DEC}} + \gamma J^{\mathrm{error}},
\end{equation}
with $\alpha,\beta,\gamma \ge 0$ being hyperparameters fulfilling $\gamma=1-\alpha-\beta$, loss $J^{\mathrm{ENC}}$ defined in the encoded Hadamard space, loss $J^{\mathrm{DEC}}$ in the decoded one-hot space, and loss $J^{\mathrm{error}}$ minimizing the $L_2$ norm of the error vector $\mathbf{e}^*_i$. The loss in the Hadamard space
\begin{equation}
\label{equ:enc_loss_supp}
    J^{\mathrm{ENC}}(\mathbf{h}_i, \overline{\mathbf{H}}_i) \in \{J^{\mathrm{BCE}},J^{\mathrm{MSE}},J^{\mathrm{MAE}}\}
\end{equation}
can be calculated as binary cross-entropy $J^{\mathrm{BCE}}$, mean squared error $J^{\mathrm{MSE}}$, or mean absolute error $J^{\mathrm{MAE}}$ between the predicted soft codewords $\mathbf{h}_i$ and the target Hadamard codeword $\overline{\mathbf{H}}_i=\mathbf{H}^{(L)}_{s=\overline{s}_i}$ (\ref{equ:hadamard_codeword_conversion}), with $\overline{s}_i$ being the class label of pixel or object $i$. The loss in the decoded one-hot space 
\begin{equation}
\label{equ:dec_loss}
    J^{\mathrm{DEC}}(\mathbf{P}_i^*, \overline{s}_i^*) = \rho \cdot J^{\mathrm{CE}}(\mathbf{P}_i^*, \overline{s}_i^*) + (1-\rho) \cdot J^{\mathrm{MAE}}(\mathbf{P}_i^*, \overline{s}_i^*)
\end{equation}
 with hyperparameter $\rho \geq 0$, is a combination of cross-entropy $J^{\mathrm{CE}}$ and mean absolute error $J^{\mathrm{MAE}}$. The loss
 \begin{equation}
 \label{equ:error_loss}
     J^{\mathrm{error}}(\mathbf{e}_i^{*}) = \|\mathbf{e}_i^{*}\|^2
 \end{equation}
 just minimizes the $L_2$ norm of the error vector $\mathbf{e}^*_i$ (\ref{equ:e_i^*}).
 Note that the minimization of $J^{\mathrm{error}}$ (\ref{equ:error_loss}) differs from the minimization during decoding in (\ref{equ:ui_min}). In Hadamard decoding (\ref{equ:ui_min}), we find the shortest error vector $\mathbf{e}^*_i$ for a given predicted soft codeword $\mathbf{h}_i$, while the loss $J^{\mathrm{error}}$ (\ref{equ:error_loss}) encourages the prediction of soft codewords $\mathbf{h}_i$ which lead to short error vectors $\mathbf{e}^{*}_i$. This effectively promotes predicted soft codewords with fewer internal inconsistencies.

\section{Details on Training Hyperparameters}
\label{supp:details_hyperparameters}
\subsubsection{Semantic Segmentation\normalfont{:}} All semantic segmentation models were trained on the Cityscapes training set \(\mathcal{D}^\mathrm{CS}_\mathrm{train}\) \cite{Cordts2016}, the BDD100K training set \(\mathcal{D}^\mathrm{BDD,Seg}_\mathrm{train}\) \cite{Yu2019}, or the ADE20K training set $\mathcal{D}^\mathrm{ADE20K}_\mathrm{train}$ \cite{Zhou2017} using a batch size of 8 for \(\mathcal{D}^\mathrm{CS}_\mathrm{train}\) and \(\mathcal{D}^\mathrm{BDD,Seg}_\mathrm{train}\) and 16 for \(\mathcal{D}^\mathrm{ADE20K}_\mathrm{train}\) on a single \texttt{Nvidia RTX Pro 6000 Blackwell} GPU. The \texttt{SegFormer MiT-B0} \cite{Xie2022segformer} was optimized with AdamW using an initial learning rate of \(6\times10^{-5}\), weight decay~0.01 and a polynomial learning rate schedule with power~1.0, decaying the learning rate to zero over a total of 160{,}000 iterations on \(\mathcal{D}^\mathrm{CS}_\mathrm{train}\), \(\mathcal{D}^\mathrm{BDD,Seg}_\mathrm{train}\) and \(\mathcal{D}^\mathrm{ADE20K}_\mathrm{train}\). The \texttt{DeepLabv3+} \cite{Chen2018a} was trained using AdamW with an initial learning rate of \(4\times10^{-5}\), weight decay~0.01, and also combined with a polynomial learning rate schedule with power~0.9. The \texttt{DeepLabv3+} was trained for 80{,}000 iterations on \(\mathcal{D}^\mathrm{CS}_\mathrm{train}\), \(\mathcal{D}^\mathrm{BDD,Seg}_\mathrm{train}\) and \(\mathcal{D}^\mathrm{ADE20K}_\mathrm{train}\). No early stopping was applied and all models were trained until completion of the respective iteration budget. Data augmentation, normalization, and preprocessing followed the standard pipelines used in \texttt{mmsegmentation} \cite{mmseg2020}, including random cropping, horizontal flipping, and per-dataset normalization.

\subsubsection{Object Detection\normalfont{:}} All object detection models were trained on the $\mathcal{D}^\mathrm{VOC}_\mathrm{train}$, $\mathcal{D}^\mathrm{BDD,OD}_\mathrm{train}$, or $\mathcal{D}^\mathrm{COCO}_\mathrm{train}$ using mainly the standard parameters of the \texttt{mmdetection} framework \cite{mmdetection} with a total batch size of 16. 
\texttt{DETR} \cite{Carion2020} was trained on a single \texttt{Nvidia RTX Pro 6000 Blackwell} GPU using AdamW \cite{Loshchilov2019} with an initial learning rate of \(1\times10^{-4}\) and weight decay \(1\times10^{-4}\). Gradient clipping with a maximum norm of \(0.1\) was employed, and the backbone parameters used a learning rate multiplier of \(0.1\). The model was trained for 150 epochs with a multi-step learning rate schedule, where the learning rate was reduced by a factor of \(0.1\) after epoch 100. \texttt{Faster R-CNN} \cite{Ren2015} was trained on a single \texttt{Nvidia A100} GPU using AdamW \cite{Loshchilov2019} with an initial learning rate of \(1\times10^{-4}\), betas \((0.9, 0.999)\), and weight decay \(0.05\). Automatic mixed precision was used during training, and gradient clipping with a maximum norm of \(35\) was applied. The model was trained for 12 epochs with a learning rate schedule consisting of a linear warm-up phase for the first 2000 iterations starting from a factor of \(1/1000\) of the base learning rate, followed by a multi-step decay where the learning rate was reduced by a factor of \(0.1\) at epochs 8 and 11. All models used the standard augmentation and preprocessing pipelines of \texttt{mmdetection} \cite{mmdetection}, including random horizontal flipping, multi-scale resizing, and per-dataset normalization. Training proceeded until completion of the full epoch budget.

\subsubsection{Regression Model\normalfont{:}} The regression model $\mathcal{F}$ is trained for 20 epochs with a total batch size of 4 for semantic segmentation and 16 for object detection using the AdamW \cite{Loshchilov2019} optimizer with an initial learning rate of \(1\times10^{-4}\), betas \((0.9, 0.999)\) and weight decay \(1\times10^{-4}\). 
We use a cosine annealing learning rate scheduler that decreases the learning rate following a cosine schedule over the full training.
As semantic segmentation models predict most pixels with high confidence ($>98\%$), we do not train on all pixels of each training sample. Instead, we use the top 1{,}000 pixels with the highest $\|\mathbf{e}_i^*\|_1$ and randomly sample an additional 4{,}000 pixels to mitigate this data imbalance.
Analogously, for object detection and the regression anomaly score (\ref{equ:anomaly_score_pred_e}), we only use objects with $H(\mathbf{P}_i^*) \ge 0.2$ in training to tackle data imbalance. In inference, we make use of the fact that $\|\mathbf{e}_i^*\|$ must be near $0$ for objects with entropy $H(\mathbf{P}_i^*) < 0.2$ by predicting $\widehat{\|\mathbf{e}_i^*\|}_1=0$ in that entropy range. Note that for object detection, the classes $S$ also include the no object class.

\section{Perturbation Types and Strengths}
\label{supp:attack_strength}
All investigated perturbation types produce a perturbation vector $\mathbf{r}_\epsilon = (r_{\epsilon, i, c}) \in [-1,1]^{C \times H \times W}$, with elements $r_{\epsilon, i, c} \in \mathbb{R}$ which is added to the original input image $\mathbf{x} \in \mathbb{I}^{C \times H \times W}$ with $\mathbb{I}=[0,1]$, to obtain the perturbed image $\mathbf{x}_\epsilon = \max(\min(\mathbf{x}+\mathbf{r}_\epsilon, 1), 0) \in [0,1]^{C \times H \times W}$. Here, $C = 3$ is the number of color channels, $H$ the image height, $W$ the image width, and $\min()$ and $\max()$ operating on pixel/object level $i$. To ensure comparable signal disturbances across perturbation types for a given strength $\epsilon > 0$, we follow the formulation of Klingner et al. \cite{Klingner2020} to achieve a consistent signal-to-noise ratio. The perturbation strength is defined as
\begin{equation}
\label{equ:perturbation_strength}
\epsilon = \sqrt{\frac{1}{HWC} \cdot \mathrm{E}(||\mathbf{r}_\epsilon||_2^2)},
\end{equation}
where $\mathrm{E}(||\mathbf{r}_\epsilon||_2^2)$ denotes the expectation of the squared $L_2$ norm of the perturbation $\mathbf{r}_{\epsilon}$.

\subsubsection{Perturbation Types for Segmentation\normalfont{:}}
For semantic segmentation, we consider Gaussian noise, salt and pepper noise, and adversarial perturbations generated using FGSM~\cite{Goodfellow2015}, PGD~\cite{Kurakin2017}, and the Metzen~\cite{Metzen2017} attack. Gaussian noise $r_{\epsilon, i, c}$ is sampled from a normal distribution $\mathcal{N}(0, \epsilon^2)$. Salt and pepper noise randomly sets a fraction of pixels to the minimum $(0)$ or maximum value $(1)$, and the resulting perturbation strength $\epsilon$ is computed using (\ref{equ:perturbation_strength}).

FGSM generates perturbations according to
\begin{equation}
\label{equ:FGSM}
\mathbf{r}_\epsilon = \epsilon \cdot \mathrm{sign}\left(\nabla_{\mathbf{x}} J^{\mathrm{CE}}(\overline{\mathbf{s}}, \mathbf{P}^*)\right),
\end{equation}
where $\nabla_{\mathbf{x}} J^{\mathrm{CE}}$ is the gradient of the cross-entropy loss with respect to the input $\mathbf{x}$, $\mathbf{P}^*$ is the predicted class posterior probability, $\overline{\mathbf{s}} = (\overline{s}_i)$ is a vector containing the ground-truth class labels, and $\mathbf{P}^*$ (\ref{equ:ui_min}) are the class posterior vectors for all pixels/objects $i$. PGD is an iterative variant of FGSM that applies multiple small steps and enforces all perturbation values to be in the range $[-\epsilon, \epsilon]$.

For the Metzen attack \cite{Metzen2017}, we employ the dynamic target variant. The adversarial objective is to change the prediction at all pixels, where the model predicted the class \texttt{car}, by redirecting them to their nearest neighbor class. Similar to PGD, the attack is an iterative variant of (\ref{equ:FGSM}), but it applies different weighting to pixels belonging to the attacked class (\texttt{car}) and to background pixels.
The adversarial attacks for semantic segmentation are reimplemented based on the descriptions provided in the respective papers \cite{Metzen2017, Kurakin2017, Goodfellow2015}.

\subsubsection{Perturbation Types for Object Detection\normalfont{:}} For object detection, in addition to universal perturbations such as Gaussian noise and salt and pepper noise, we consider the TOG attacks, specifically the untargeted, vanishing, fabrication, and mislabeling attacks proposed by~\cite{Chow2020}. The TOG attacks can be interpreted as PGD attacks adapted to object detection, differing primarily in their attack objectives. Specifically, the untargeted attack aims at maximizing the overall detection loss $J^{\mathrm{OD}}$ (\ref{equ:object_detection_loss}). The vanishing attack minimizes $J^{\mathrm{OD}}$ with respect to an empty set of target objects, thereby suppressing all detections. Conversely, the fabrication attack maximizes $J^{\mathrm{OD}}$ under the same empty target set to induce additional detections. Finally, the mislabeling attack minimizes $J^{\mathrm{OD}}$ using maliciously chosen class labels for the target objects. For the maliciously chosen class labels, we assign to each object the class with the second-highest posterior probability. The adversarial attacks for object detection are reimplemented using the published source code from \cite{Chow2020}. For the baseline by Xu et al.\ \cite{Xu2018e}, we use the Hungarian matching cost \cite{Carion2020} as the distance function between two predictions.

\section{Semantic Segmentation Ablation}
\label{supp:SemanticSegmentationAblation}
\begin{table}[t!]
  \centering
  \caption{\textbf{Ablation study on loss functions for \texttt{Hadamard} in semantic segmentation.} We report mIoU (\%) and global AuROC (\%) for different loss configurations (\ref{equ:overall_loss_supp}) of \texttt{HadamardNet} using \texttt{SegFormer MiT-B0} on $\mathcal{D}^\mathrm{CS}_\mathrm{val}$. Best results in bold, second best underlined.}
  \label{tab:ablation_losses}
    
\extrarowheight=\aboverulesep
    \addtolength{\extrarowheight}{\belowrulesep}
    \aboverulesep=0pt
    \belowrulesep=1pt
    \begin{tabular}{@{}lll rrrrr@{}}
    \toprule[.9pt]
    \makecell[l]{$\alpha J^\mathrm{ENC}$ \\ (\ref{equ:enc_loss_supp})} & 
    \makecell[l]{$\beta J^\mathrm{DEC}$ \\ (\ref{equ:dec_loss})} & 
    \makecell[l]{$\gamma J^\mathrm{error}$\\ (\ref{equ:error_loss})} & 
    {mIoU} &
    \makecell[r]{AuROC \\(FGSM)} &
    \makecell[r]{AuROC \\(Metzen)} &
    \makecell[r]{AuROC \\(PGD)} &
    \makecell[r]{AuROC \\(S\&P)} \\
    \midrule
    \multicolumn{3}{c}{One-hot baseline} & $76.4^{\pm 0.4}$ & - & - & - & - \\
    \midrule
    \midrule
    - &  $J^{\mathrm{DEC}} = J^{\mathrm{CE}}$ & - & $\underline{76.5}^{\pm 0.2}$ & $91.3^{\pm 0.7}$ & $66.3^{\pm 1.6}$ & $84.9^{\pm 2.6}$ & $64.8^{\pm 3.6}$\\
    - & $1.0\cdot J^\mathrm{DEC}$ & - & ${76.4}^{\pm 0.3}$ & $91.6^{\pm 1.3}$ & $68.7^{\pm 0.5}$ & $87.3^{\pm 1.8}$ & $65.1^{\pm 1.2}$\\
    \midrule
    $0.2\cdot J^\mathrm{BCE}$ & $0.8\cdot J^\mathrm{DEC}$ & - & $76.2^{\pm 0.2}$ & $92.0^{\pm 0.8}$ & $75.6^{\pm 2.4}$ & $92.3^{\pm 0.6}$ & $64.7^{\pm 1.1}$\\
    $0.2\cdot J^\mathrm{MSE}$ & $0.8\cdot J^\mathrm{DEC}$ & - & $76.2^{\pm 0.3}$ &$88.8^{\pm 2.7}$&$70.7^{\pm 2.9}$&$90.1^{\pm 1.6}$& $64.5^{\pm 1.7}$\\
    $0.2\cdot J^\mathrm{MAE}$ &$0.8\cdot J^\mathrm{DEC}$ & - & $76.0^{\pm 0.2}$ &$92.2^{\pm 2.3}$&$70.3^{\pm 1.5}$&$84.2^{\pm 1.1}$& $65.9^{\pm 3.1}$\\
    \midrule
    $1.0\cdot J^\mathrm{BCE}$ & - & - & $76.2^{\pm 0.1}$ & $94.5^{\pm 0.0}$ & $\underline{84.1}^{\pm 1.0}$ & $92.4^{\pm 3.3}$ & $68.4^{\pm 1.3}$\\
    $1.0\cdot J^\mathrm{MSE}$ & - & - & $76.3^{\pm 0.3}$ & $\mathbf{96.8}^{\pm 0.6}$ & $\textbf{84.5}^{\pm 0.9}$ & $\mathbf{95.2}^{\pm 1.0}$ & $\mathbf{73.3}^{\pm 0.7}$\\
    $1.0\cdot J^\mathrm{MAE}$ & - & - & $76.4^{\pm 0.5}$ &$88.4^{\pm 1.2}$&$63.8^{\pm 2.5}$&$92.8^{\pm 1.9}$& $\underline{69.8}^{\pm 1.2}$\\
    \midrule
    $0.8\cdot J^\mathrm{BCE}$ & - & $0.20\cdot J^\mathrm{MSE}$ & $76.4^{\pm 0.3}$& $94.0^{\pm 1.0}$& $83.6^{\pm 8.1}$& $93.0^{\pm 1.2}$& $68.4^{\pm 1.6}$\\
    $0.8\cdot J^\mathrm{MSE}$ & - & $0.20\cdot J^\mathrm{MSE}$ & $\mathbf{76.6}^{\pm 0.2}$& $\underline{95.7}^{\pm 0.5}$& $79.6^{\pm 1.1}$& $\underline{95.1}^{\pm 1.3}$& $65.8^{\pm 1.1}$\\
    $0.8\cdot J^\mathrm{MAE}$ & - & $0.20\cdot J^\mathrm{MSE}$ & $76.3^{\pm 0.3}$ & $83.2^{\pm 11.1}$& $66.1^{\pm 6.8}$& $87.1^{\pm 9.7}$& $66.0^{\pm 2.5}$\\
    \bottomrule[.9pt]
\end{tabular}

\end{table}
In \cref{tab:ablation_losses}, we present an ablation study on different loss function configurations for \texttt{HadamardNet} in semantic segmentation. All experiments are conducted using a \texttt{SegFormer MiT-B0} \cite{Xie2022segformer} trained on $\mathcal{D}^\mathrm{CS}_\mathrm{train}$ and evaluated on $\mathcal{D}^\mathrm{CS}_\mathrm{val}$. We report mIoU (\%) in absence of attacks or disturbances for various combinations losses and global AuROC for each perturbation across strengths $\epsilon\in \{1,2,4,8,16\}$. Results are compared against a one-hot encoded baseline.
We observe that nearly all mIoU values are statistically equivalent across all investigated loss combinations and also statistically equivalent to the one-hot baseline mIoU. However, the loss combinations have a strong impact on the perturbation detection performance (AuROC). Using only a loss in the decoded one-hot space $J^\mathrm{DEC}$ (first table segment), or assigning a large weight to $J^\mathrm{DEC}$ ($\beta = 0.8$, second table segment), leads to poor AuROC across all investigated perturbations. Using only a Hadamard space loss $J^\mathrm{ENC}$ (third table segment) yields the best AuROC performance, with $J^\mathrm{ENC}=J^\mathrm{MSE}$ achieving the highest AuROC for all investigated perturbations. Additionally, introducing a small contribution of the error loss $J^\mathrm{error}$ with weight $\gamma = 0.2$ does not provide further improvements but instead degrades perturbation detection (AuROC) across all perturbations. Therefore, we finally decided for $J^\mathrm{cls}=J^\mathrm{ENC}=J^\mathrm{MSE}$ for all following experiments \textit{both in the semantic segmentation and object detection task}.

\section{Detailed Perturbation Detection Results on Cityscapes for Semantic Segmentation}
\label{supp:SemanticSegmentationPerturbation}
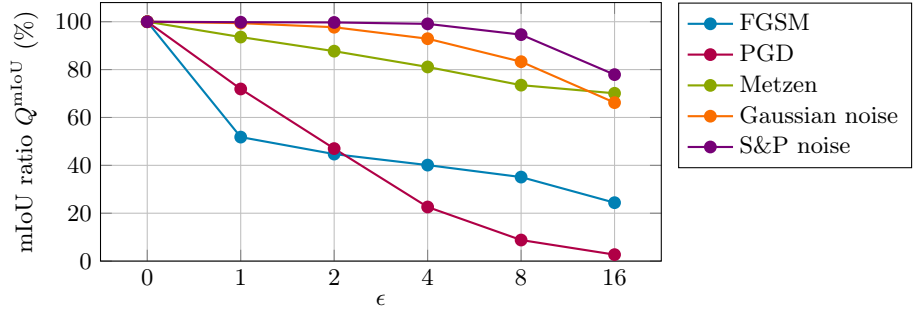
\begin{figure}[t]
    \centering
    \begin{tikzpicture}

\begin{axis}[
    width=9cm,
    height=5cm,
    xlabel={$\epsilon$},
    xlabel style={yshift=1mm}, 
    ylabel={mIoU ratio $Q^{\mathrm{mIoU}}$ (\%)},
    ylabel style={yshift=-0mm},
    xmode=log,
    log basis x=2,
    xtick={0.5,1,2,4,8,16},
    xticklabels={0,1,2,4,8,16},
    ymin=0, ymax=108,
    ytick={0,20,40,60,80,100},
    grid=both,
    clip=false,
    legend pos= outer north east,
    legend cell align={left}
]

\addplot[tu5,thick,mark=*] coordinates { (0.5,100) (1,51.8) (2,44.7) (4,40.1) (8,35.1) (16,24.4) };
\addlegendentry{FGSM}


\addplot[tu3,thick,mark=*] coordinates { (0.5,100) (1,71.9) (2,47.0) (4,22.6) (8,8.8) (16,2.7) };
\addlegendentry{PGD}


\addplot[tu8!70!black,thick,mark=*] coordinates { (0.5,100) (1,93.6) (2,87.7) (4,81.1) (8,73.5) (16,70.1) };
\addlegendentry{Metzen}


\addplot[tu2,thick,mark=*] coordinates { (0.5,100) (1,99.4) (2,97.7) (4,92.9) (8,83.3) (16,66.2) };
\addlegendentry{Gaussian noise}


\addplot[tu120,thick,mark=*] coordinates { (0.5,100) (1,99.8) (2,99.7) (4,99.1) (8,94.6) (16,77.9) };
\addlegendentry{S\&P noise}


\end{axis}
\node[anchor=north west] at (7.526cm,1.074cm) {%
    
};

\end{tikzpicture}
    \setlength{\abovecaptionskip}{0pt} 
    \caption{\textbf{Performance degradation under perturbations} in terms of mIoU ratios $Q^{\mathrm{mIoU}}$. All results are obtained using a \texttt{SegFormer MiT-B0} model~\cite{Xie2022segformer} on perturbed images from $\mathcal{D}^{\mathrm{CS}}_{\mathrm{val}}$.}
\label{fig:semseg_robustness}
\end{figure}

In \cref{fig:semseg_robustness}, we show the performance of the \texttt{SegFormer MiT-B0}~\cite{Xie2022segformer} model with Hadamard output encoding at different perturbation strengths $\epsilon$ and different perturbations in terms of the ratio $Q^{\mathrm{mIoU}} = \frac{\mathrm{mIoU}^{\mathrm{perturbed}}}{\mathrm{mIoU}^{\mathrm{clean}}}$ between perturbed and clean performance. The model is trained on $\mathcal{D}^\mathrm{CS}_\mathrm{train}$ and evaluated on perturbed images from $\mathcal{D}^\mathrm{CS}_\mathrm{val}$.
Across all attacks, segmentation performance ($Q^{\mathrm{mIoU}}$) decreases with increasing perturbation strength $\epsilon$. For small perturbation strengths ($\epsilon \le 2$), FGSM~\cite{Goodfellow2015} causes the largest mIoU degradation, whereas PGD~\cite{Kurakin2017} leads to the strongest degradation for larger perturbation strengths ($\epsilon \ge 4$). In contrast, the Metzen attack~\cite{Metzen2017} degrades performance substantially less, reaching a minimum of $Q^{\mathrm{mIoU}}=70.1\%$, compared to $Q^{\mathrm{mIoU}}=2.7\%$ for PGD. This behavior is explained by its adversarial objective, which suppresses only a single target class while preserving the remaining predictions. As a result, \textit{Metzen perturbations are particularly difficult to detect at small strengths $\epsilon$}.
For disturbances, segmentation performance remains largely unaffected at small strengths, specifically for $\epsilon \le 2$ with Gaussian noise and $\epsilon \le 4$ with salt and pepper noise. At higher strengths, both disturbances degrade performance, with Gaussian noise causing the stronger effect. Consequently, \textit{both disturbances are difficult to detect at small perturbation strengths $\epsilon$}.

In \cref{tab:perturbations_semseg}, we report the performance of our quantile regression approach (\ref{equ:anomaly_score_reg}) for all perturbations and strengths $\epsilon\in \{1,2,4,8,16\}$ using a \texttt{SegFormer MiT-B0} \cite{Xie2022segformer} trained on $\mathcal{D}^\mathrm{CS}_\mathrm{train}$ and evaluated on perturbed samples from $\mathcal{D}^\mathrm{CS}_\mathrm{val}$. We report the perturbation detection accuracy measured by AuROC (\%) and global AuROC (\%) and global TPR$_{5\%}$ $(\%)$ per perturbation across all strengths.
FGSM and PGD, which also cause the strongest mIoU degradation, achieve the highest detection performance across all perturbation strengths $\epsilon$. Our quantile regression approach (\ref{equ:anomaly_score_reg}) is also able to detect the Metzen attack, but with lower AuROC due to its smaller mIoU degradation for $\epsilon \le 4$.
As expected, for disturbances, the AuROC at $\epsilon \le 2$ is close to random guessing, as these perturbations do not degrade mIoU. For larger strengths ($\epsilon \ge 4$), where a measurable mIoU drop occurs, our method is able to detect the disturbances, with detection performance increasing with perturbation strength $\epsilon$.

\begin{table}[t!]
  \centering
  \caption{\textbf{Perturbation detection by quantile regression (\ref{equ:anomaly_score_reg}) for semantic segmentation} on $\mathcal{D}^\mathrm{CS}_\mathrm{val}$. We report detection performance across perturbation types and strengths $\epsilon \in \{1,2,4,8,16\}$ using \texttt{SegFormer MiT-B0}. Results are given as AuROC (\%) per $\epsilon$, global AuROC (\%) and global TPR$_{5\%}$ $(\%)$ across all $\epsilon$.}
  \label{tab:perturbations_semseg}
    \renewcommand{\arraystretch}{0.8}
\setlength{\tabcolsep}{3pt}

\begin{tabular}{@{}c|l|r|r|r|r|r|r|r@{}}
\toprule[.9pt]
\multicolumn{1}{c|}{} &
\multicolumn{1}{l|}{Perturbation} &
\multicolumn{5}{c|}{AuROC (\%)} &
\makecell[c]{Global} &
\makecell[c]{Global} \\
\multicolumn{1}{c|}{} &
&
{$\epsilon=1$} & {$\epsilon=2$} & {$\epsilon=4$} & {$\epsilon=8$} & {$\epsilon=16$} &
{AuROC} &
{TPR$_{5\%}$} \\
\midrule
\multirow{5}{*}{\rotatebox{90}{Cityscapes}} &
FGSM           & $90.9$ & $93.6$ & $93.7$ & $94.1$ & $96.4$ & $95.9$ & $79.3$ \\
& PGD           & $90.6$ & $99.2$ & $99.9$ & $94.1$ & $91.9$ & $96.1$ & $82.5$ \\
& Metzen        & $61.3$ & $69.6$ & $80.1$ & $87.7$ & $91.9$ & $75.9$ & $31.9$ \\
& Gauss.\ Noise & $51.4$ & $54.7$ & $61.4$ & $74.8$ & $91.4$ & $68.5$ & $32.9$ \\
& S\&P Noise    & $50.2$ & $50.9$ & $55.8$ & $70.7$ & $89.1$ & $64.4$ & $30.0$ \\
\bottomrule[.9pt]
\end{tabular}
\end{table}

\section{Additional ROC Curves on Cityscapes for Semantic Segmentation}
\label{supp:ROC_Sem_Seg}

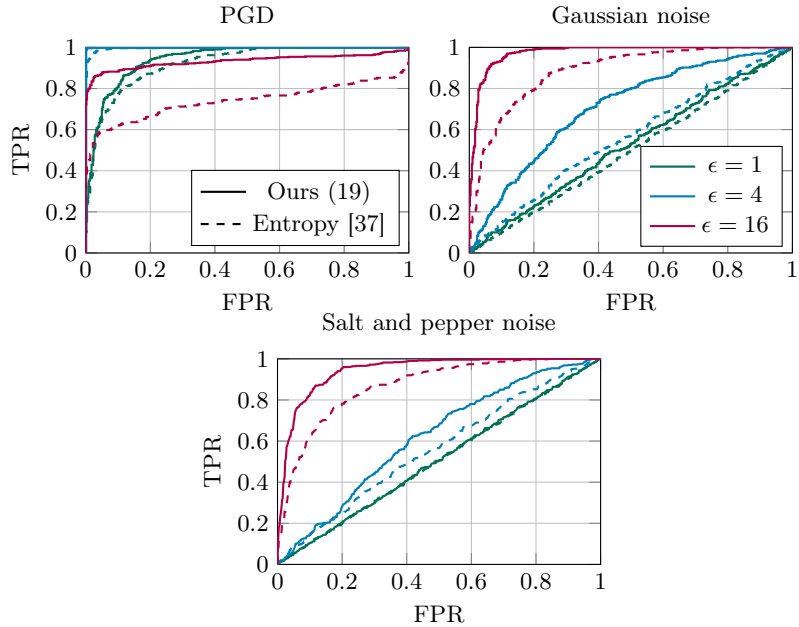
\begin{figure}[t!]
    \centering
    \begin{tikzpicture}

\begin{groupplot}[
    group style={
        group size=2 by 1,
        horizontal sep=0.8cm
    },
    width=0.48\textwidth,
    height=4.3cm,
    xmin=0, xmax=1,
    ymin=0, ymax=1,
    xlabel={FPR},
    ylabel={TPR},
    ylabel style={yshift=-0mm},
    grid=both
]

\nextgroupplot[
    title={PGD},
    legend style={at={(0.97,0.43)}, anchor=south east, font=\small}
]

\addplot[thick,tu10] table [col sep=comma, x=fpr, y=tpr] {figures/roc_Quantile_anomaly_score_pgd_ours_eps1.csv};
\addplot[thick,tu10,dashed,forget plot] table [col sep=comma, x=fpr, y=tpr] {figures/roc_entropy_pgd_baseline_eps1.csv};

\addplot[very thick,tu5] table [col sep=comma, x=fpr, y=tpr] {figures/roc_Quantile_anomaly_score_pgd_ours_eps4.csv};
\addplot[thick,tu5,dashed,forget plot] table [col sep=comma, x=fpr, y=tpr] {figures/roc_entropy_pgd_baseline_eps4.csv};

\addplot[thick,tu3] table [col sep=comma, x=fpr, y=tpr] {figures/roc_Quantile_anomaly_score_pgd_ours_eps16.csv};
\addplot[thick,tu3,dashed,forget plot] table [col sep=comma, x=fpr, y=tpr] {figures/roc_entropy_pgd_baseline_eps16.csv};


\coordinate (fgsmLegendPos) at (rel axis cs:0.97,0.03);

\nextgroupplot[
    title={Gaussian noise},
    ylabel={},
    legend style={at={(axis description cs:0.97,0.03)}, anchor=south east, font=\small}
]

\addplot[thick,tu10] table [col sep=comma, x=fpr, y=tpr] {figures/roc_Quantile_anomaly_score_gaussian_noise_ours_eps1.csv};
\addplot[thick,tu10,dashed,forget plot] table [col sep=comma, x=fpr, y=tpr] {figures/roc_entropy_gaussian_noise_baseline_eps1.csv};

\addplot[thick,tu5] table [col sep=comma, x=fpr, y=tpr] {figures/roc_Quantile_anomaly_score_gaussian_noise_ours_eps4.csv};
\addplot[thick,tu5,dashed,forget plot] table [col sep=comma, x=fpr, y=tpr] {figures/roc_entropy_gaussian_noise_baseline_eps4.csv};

\addplot[thick,tu3] table [col sep=comma, x=fpr, y=tpr] {figures/roc_Quantile_anomaly_score_gaussian_noise_ours_eps16.csv};
\addplot[thick,tu3,dashed,forget plot] table [col sep=comma, x=fpr, y=tpr] {figures/roc_entropy_gaussian_noise_baseline_eps16.csv};

\addlegendentry{$\epsilon = 1$}
\addlegendentry{$\epsilon = 4$}
\addlegendentry{$\epsilon = 16$}

\end{groupplot}

\path (group c1r1.south west) -- (group c2r1.south east) coordinate[pos=0.5] (topRowCenter);

\begin{axis}[
    at={(topRowCenter)},
    anchor=north,
    yshift=-1.4cm,
    width=0.48\textwidth,
    height=4.3cm,
    xmin=0, xmax=1,
    ymin=0, ymax=1,
    xlabel={FPR},
    ylabel={TPR},
    ylabel style={yshift=0mm},
    grid=both,
    title={Salt and pepper noise}
]

\addplot[thick,tu10] table [col sep=comma, x=fpr, y=tpr] {figures/roc_Quantile_anomaly_score_sp_noise_ours_eps1.csv};

\addplot[thick,tu10,dashed,forget plot] table [col sep=comma, x=fpr, y=tpr] {figures/roc_entropy_sp_noise_baseline_eps1.csv};

\addplot[thick,tu5] table [col sep=comma, x=fpr, y=tpr] {figures/roc_Quantile_anomaly_score_sp_noise_ours_eps4.csv};

\addplot[thick,tu5,dashed,forget plot] table [col sep=comma, x=fpr, y=tpr] {figures/roc_entropy_sp_noise_baseline_eps4.csv};

\addplot[thick,tu3] table [col sep=comma, x=fpr, y=tpr] {figures/roc_Quantile_anomaly_score_sp_noise_ours_eps16.csv};

\addplot[thick,tu3,dashed,forget plot] table [col sep=comma, x=fpr, y=tpr] {figures/roc_entropy_sp_noise_baseline_eps16.csv};
\end{axis}

\begin{axis}[
    hide axis,
    scale only axis,
    width=0pt,
    height=0pt,
    at={(0,0)},
    anchor=south west,
    xmin=0, xmax=1,
    ymin=0, ymax=1,
    legend to name=methodlegend4,
    legend style={draw=black,fill=white,font=\small}
]
\addlegendimage{black,thick}
\addlegendentry{Ours (\ref{equ:anomaly_score_reg})}
\addlegendimage{black,thick,dashed}
\addlegendentry{Entropy \cite{Smith2018}}
\end{axis}

\node at (fgsmLegendPos) [anchor=south east, xshift=3.5pt, yshift=-3pt]
    {\pgfplotslegendfromname{methodlegend4}};

\end{tikzpicture}
    \caption{\textbf{ROC for our proposed quantile detection (\ref{equ:anomaly_score_reg}) for semantic segmentation}, left for the PGD attack, right for Gaussian noise, and bottom for salt and pepper noise at perturbation strengths $\epsilon \in \{1, 4, 16\}$. All results are obtained using a \texttt{SegFormer MiT-B0} model \cite{Xie2022segformer} trained on $\mathcal{D}^\mathrm{CS}_\mathrm{train}$ and evaluated on perturbed images from $\mathcal{D}^\mathrm{CS}_\mathrm{val}$.}
    \label{fig:ROC_seg2}
\end{figure}

In \cref{fig:ROC_seg2}, we display ROC curves for the PDG attack (left), Gaussian noise (right) and salt and pepper noise (bottom) on \textit{semantic segmentation} for different perturbation strengths~$\epsilon$ for our quantile detection (\ref{equ:anomaly_score_reg}) and the entropy baseline \cite{Smith2018}. The results are obtained using the \texttt{SegFormer MiT-B0}~\cite{Xie2022segformer} model, which is trained on $\mathcal{D}^\mathrm{CS}_\mathrm{train}$ and evaluated on perturbed images from $\mathcal{D}^\mathrm{CS}_\mathrm{val}$. The ROC curves are obtained by varying the decision threshold $\theta$ applied to the anomaly score $\delta(\textbf{x})$ (\ref{equ:anomaly_score_reg}).
We observe that for disturbances ROC curves get better the stronger the disturbances are. Our quantile method (\ref{equ:anomaly_score_reg}) excels the entropy approach for both disturbances, for high strength $\epsilon \geq 4$, where a measurable mIoU drop occurs, even with high significance. For PGD, detection performance improves from $\epsilon = 1$ to $\epsilon = 4$, but decreases at $\epsilon = 16$ for both methods. Nevertheless, our quantile method (\ref{equ:anomaly_score_reg}) outperforms the entropy baseline for all perturbation strengths, at $\epsilon = 16$ even with a significant margin.

\section{Comparison with Existing Hadamard Output Encoding for Semantic Segmentation}
\label{supp:comparison_hoyos}

In \cref{tab:hadamard_comparison_hoyos}, we compare our Hadamard output encoding to the existing Hadamard-based semantic segmentation method of Hoyos et al.\ \cite{Hoyos2024} on a common \texttt{UNet} backbone \cite{Ronneberger2015}. Note that, the underlying \texttt{UNet} architectures differ across the different output representations. The \texttt{UNet} implementation for the one-hot variant and for Hadamard (ours) follows the modern \texttt{UNet-S5-D16} design equipped with an \texttt{FCN} segmentation head. Hoyos et al.\ \cite{Hoyos2024} only published on the Pix2Pix-style \texttt{UNet} generator originally designed for image-to-image translation.
With the setup of Hoyos et al.\ \cite{Hoyos2024}, the mIoU is drastically lower than that of a standard \texttt{UNet} baseline ($23.00\%$ vs.\ $69.10\%$ mIoU). In contrast, applying our \texttt{HadamardNet} to the modern \texttt{UNet} architecture achieves nearly identical performance to the one-hot baseline ($68.74\%$ mIoU). These results suggest that the setup of Hoyos et al.\ \cite{Hoyos2024} does not allow meaningful conclusions about the usefulness of Hadamard output encodings for semantic segmentation, leaving this question largely unexplored.

\begin{table}[t]
  \centering
  \caption{\textbf{Comparison of one-hot and Hadamard output encodings with prior work on} \(\mathcal{D}^\mathrm{CS}_\mathrm{val}\).}
  \label{tab:hadamard_comparison_hoyos}
  \begin{tabular}{@{}l@{\hspace{8pt}}|@{\hspace{8pt}}l r@{}}
    \toprule[.9pt]
    Model & Output representation & mIoU (\%) \\
    \midrule
    \multirow{3}{*}{\texttt{UNet}} 
      & One-hot & 69.10 \\
      & Hadamard (Hoyos \cite{Hoyos2024}) & 23.00 \\
      & Hadamard (ours) & 68.74 \\
    \bottomrule[.9pt]
  \end{tabular}
\end{table}

\section{Detailed Perturbation Detection Results on BDD100K for Semantic Segmentation}
\label{supp:additional_results_bdd_seg}

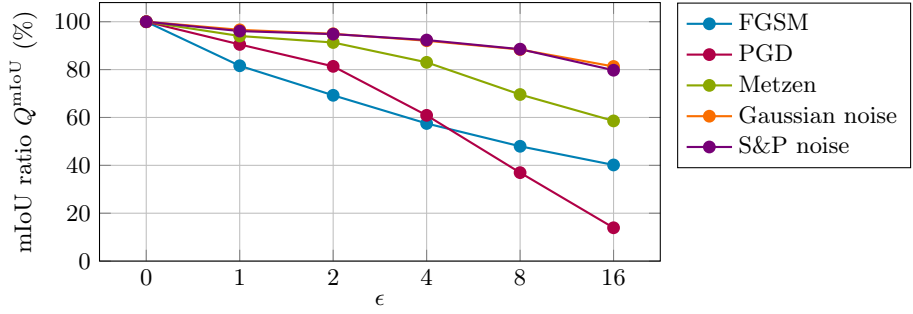
\begin{figure}[t]
    \centering
    \begin{tikzpicture}

\begin{axis}[
    width=9cm,
    height=5cm,
    xlabel={$\epsilon$},
    xlabel style={yshift=1mm},
    ylabel={mIoU ratio $Q^{\mathrm{mIoU}}$ (\%)},
    ylabel style={yshift=-0mm},
    xmode=log,
    log basis x=2,
    xtick={0.5,1,2,4,8,16},
    xticklabels={0,1,2,4,8,16},
    ymin=0, ymax=108,
    ytick={0,20,40,60,80,100},
    grid=both,
    clip=false,
    legend pos=outer north east,
    legend cell align={left}
]

\addplot[tu5,thick,mark=*] coordinates { (0.5,100) (1,81.60) (2,69.26) (4,57.49) (8,47.96) (16,40.16)};
\addlegendentry{FGSM}

\addplot[tu3,thick,mark=*] coordinates { (0.5,100) (1,90.51) (2,81.35) (4,60.89) (8,36.95) (16,13.92)};
\addlegendentry{PGD}

\addplot[tu8!70!black,thick,mark=*] coordinates { (0.5,100) (1,94.03) (2,91.32) (4,83.03) (8,69.59) (16,58.56) };
\addlegendentry{Metzen}

\addplot[tu2,thick,mark=*] coordinates { (0.5,100) (1,96.61) (2,94.95) (4,92.05) (8,88.39) (16,81.30) };
\addlegendentry{Gaussian noise}

\addplot[tu120,thick,mark=*] coordinates { (0.5,100) (1,96.09) (2,94.82) (4,92.34) (8,88.53) (16,79.74) };
\addlegendentry{S\&P noise}

\end{axis}
\node[anchor=north west] at (7.526cm,1.074cm) {%
};

\end{tikzpicture}
    \setlength{\abovecaptionskip}{0pt} 
    \caption{\textbf{Performance degradation under perturbations} in terms of mIoU ratios $Q^{\mathrm{mIoU}}$. All results are obtained using a \texttt{SegFormer MiT-B0} model~\cite{Xie2022segformer} on perturbed images from $\mathcal{D}^{\mathrm{BDD,Seg}}_{\mathrm{val}}$.}
\label{fig:semseg_robustness_bdd}
\end{figure}

In \cref{fig:semseg_robustness_bdd}, we show the performance of the \texttt{SegFormer MiT-B0}~\cite{Xie2022segformer} model with Hadamard output encoding at different perturbation strengths $\epsilon$ and different perturbations in terms of the ratio $Q^{\mathrm{mIoU}} = \frac{\mathrm{mIoU}^{\mathrm{perturbed}}}{\mathrm{mIoU}^{\mathrm{clean}}}$ between perturbed and clean performance. The model is trained on $\mathcal{D}^\mathrm{BDD,Seg}_\mathrm{train}$ and evaluated on perturbed images from $\mathcal{D}^\mathrm{BDD,Seg}_\mathrm{val}$.
Segmentation performance ($Q^{\mathrm{mIoU}}$) again decreases with increasing perturbation strength $\epsilon$. For small perturbations ($\epsilon \leq 4$), FGSM~\cite{Goodfellow2015} causes the largest mIoU degradation, whereas PGD~\cite{Kurakin2017} leads to the strongest degradation for higher strengths ($\epsilon \geq 8$). The Metzen attack~\cite{Metzen2017} degrades performance much less, as its adversarial objective suppresses only a single target class while preserving the remaining predictions, which also makes it harder to detect.
For disturbances, the behavior is similar to that observed on $\mathcal{D}^\mathrm{CS}_\mathrm{val}$. For small strengths ($\epsilon \leq 4$), mIoU degradation is negligible, while for larger strengths ($\epsilon \geq 8$) only a minor degradation occurs. Consequently, disturbances are particularly difficult to detect at small perturbation strengths $\epsilon$.

In \cref{tab:perturbations_semseg_bdd}, we report the performance of our error vector anomaly score (\ref{equ:anomaly_simple}) for all perturbations and strengths $\epsilon \in \{1,2,4,8,16\}$ using a \texttt{SegFormer MiT-B0} \cite{Xie2022segformer} trained on $\mathcal{D}^\mathrm{BDD,Seg}_\mathrm{train}$ and evaluated on perturbed samples from $\mathcal{D}^\mathrm{BDD,Seg}_\mathrm{val}$. We report the perturbation detection accuracy measured by AuROC (\%) and global AuROC (\%) as well as global TPR$_{5\%}$ (\%) per perturbation across all strengths.
FGSM and PGD, which also cause the strongest mIoU degradation, achieve the highest detection performance and the AuROC increases consistently with perturbation strength $\epsilon$, reaching $92.6\%$ for FGSM and $96.9\%$ for PGD at $\epsilon=16$. Our error anomaly score (\ref{equ:anomaly_simple}) is also able to detect the Metzen attack, but with lower AuROC due to its smaller mIoU degradation for small strengths $\epsilon$. Again, for disturbances, the AuROC at $\epsilon \le 2$ is close to random guessing, as these perturbations do not degrade mIoU. For larger strengths ($\epsilon \ge 4$), where a measurable mIoU drop occurs, our method is able to detect the disturbances, with detection performance increasing with perturbation strength $\epsilon$.

\begin{table}[t!]
  \centering
  \caption{\textbf{Perturbation detection by error vector anomaly score (\ref{equ:anomaly_simple}) for semantic segmentation} on $\mathcal{D}^\mathrm{BDD,Seg}_\mathrm{val}$. We report detection performance across perturbation types and strengths $\epsilon \in \{1,2,4,8,16\}$ using \texttt{SegFormer MiT-B0}. Results are given as AuROC (\%) per $\epsilon$, global AuROC (\%) and global TPR$_{5\%}$ (\%) across all $\epsilon$.}
  \label{tab:perturbations_semseg_bdd}
  \renewcommand{\arraystretch}{0.8}
\setlength{\tabcolsep}{3pt}

\begin{tabular}{@{}c|l|r|r|r|r|r|r|r@{}}
\toprule[.9pt]
\multicolumn{1}{c|}{} &
\multicolumn{1}{l|}{Perturbation} &
\multicolumn{5}{c|}{AuROC (\%)} &
\makecell[c]{Global} &
\makecell[c]{Global} \\
\multicolumn{1}{c|}{} &
&
{$\epsilon=1$} & {$\epsilon=2$} & {$\epsilon=4$} & {$\epsilon=8$} & {$\epsilon=16$} &
{AuROC} &
{TPR$_{5\%}$} \\
\midrule
\multirow{5}{*}{\rotatebox{90}{BDD100K}} &
FGSM           
& $65.1$ 
& $77.7$ 
& $86.6$ 
& $91.3$ 
& $92.6$ 
& $83.7$ 
& $40.3$ \\

& PGD            
& $57.7$ 
& $68.1$ 
& $85.9$ 
& $95.7$ 
& $96.9$ 
& $81.5$ 
& $47.0$ \\

& Metzen         
& $55.5$  
& $59.3$ 
& $67.7$ 
& $80.2$ 
& $90.6$ 
& $70.7$ 
& $21.2$ \\

& Gauss.\ Noise  
& $52.5$ 
& $55.7$ 
& $59.4$ 
& $62.0$ 
& $67.8$ 
& $62.5$ 
& $8.3$ \\

& S\&P Noise     
& $50.0$ 
& $50.3$ 
& $53.0$ 
& $62.2$ 
& $73.0$ 
& $61.3$ 
& $6.7$ \\
\bottomrule[.9pt]
\end{tabular}
\end{table}
\section{Results on Large-Vocabulary Datasets}
\label{supp:large_vocabulary_datasets}
\subsubsection{Semantic Segmentation\normalfont{:}}
In \cref{tab:anomaly_score_func_ade}, we compare the clean mIoU performance and the perturbed detection performance of all three variants of our proposed method described in Section \ref{subsec:attack_detection} with the two baseline detection approaches entropy \cite{Smith2018} and max posterior \cite{Hendrycks2017} for semantic segmentation across multiple perturbation types and strengths $\epsilon$. All methods employ a \texttt{SegFormer MiT-B0} \cite{Xie2022segformer} trained on $\mathcal{D}^\mathrm{ADE20K}_\mathrm{train}$ and evaluated on perturbed samples from $\mathcal{D}^\mathrm{ADE20K}_\mathrm{val}$. We report clean mIoU (\%), the perturbation detection accuracy measured by global AuROC (\%) per perturbation across all strengths and across all perturbations and strengths.
We observe that, on clean data, we reach among all methods close-by semantic segmentation mIoU on $\mathcal{D}^\mathrm{ADE20K}_\mathrm{val}$ dataset. Regarding perturbation detection performance, none of the methods is able to reliably detect Gaussian noise. Our error vector anomaly score (\ref{equ:anomaly_simple}) and the quantile anomaly score (\ref{equ:anomaly_score_reg}) show stronger detection performance for the Metzen attack and salt and pepper noise compared to the baselines. In contrast, the entropy and max posterior baselines achieve higher AuROC on the FGSM and PGD attacks, which leads to a marginally higher global AuROC overall.
However, the global AuROC values of all methods remain relatively more close to random guessing ($50\%$) compared to $\mathcal{D}^\mathrm{BDD,Seg}_\mathrm{val}$ and $\mathcal{D}^\mathrm{BDD,Seg}_\mathrm{val}$, indicating that perturbation detection on $\mathcal{D}^\mathrm{ADE20K}_\mathrm{val}$ is substantially more challenging than on $\mathcal{D}^\mathrm{BDD,Seg}_\mathrm{val}$ or $\mathcal{D}^\mathrm{CS}_\mathrm{val}$. This difficulty can be attributed to several factors. First, ADE20K contains 150 semantic classes, which leads to inherently higher prediction uncertainty and weaker class separation compared to datasets with 19 classes such as BDD100K or Cityscapes. Second, models trained on ADE20K typically achieve lower baseline mIoU, which reduces the relative performance change caused by perturbations. Third, ADE20K contains highly diverse scenes, camera setups, and image resolutions, which increases the natural variability of predictions and makes perturbation effects harder to distinguish from normal prediction noise.
\begin{table}[t!]
\centering
\caption{\textbf{Perturbation detection for semantic segmentation} on $\mathcal{D}^\mathrm{ADE20K}_\mathrm{val}$ using \texttt{SegFormer MiT-B0}. Global AuROC (best in bold) per perturbation across all strengths $\epsilon\in\{1,2,4,8,16\}$ and across all perturbations and strengths in \% and mIoU in \%.}
\label{tab:anomaly_score_func_ade}
\setlength{\tabcolsep}{0.1pt}
{
\setlength{\tabcolsep}{1pt}
\begin{tabular}{l|l|lllll|l}
\toprule[.9pt]
\multirow{2}{*}{Method} &
\makecell[c]{\multirow{2}{*}{\begin{tabular}{c}mIoU\\clean\end{tabular}}} &
\multicolumn{5}{c|}{Global AuROC (\%)} &
\multirow{2}{*}{\begin{tabular}{c}Global\\AuROC\end{tabular}} \\

& & \makecell[c]{FGSM} & \makecell[c]{PGD} & \makecell[c]{Metzen} & \makecell[c]{Gauss} & \makecell[c]{S\&P} & \\

\midrule
Entropy \cite{Smith2018} &
$37.7^{\pm0.4}$ &
$\textbf{73.9}^{\pm 0.6}$ & $\textbf{71.2}^{\pm 0.9}$ & $47.3^{\pm 0.4}$ &
$\textbf{50.0}^{\pm 0.0}$ & $52.3^{\pm 0.3}$ &
$\textbf{58.9}^{\pm 0.2}$ \\

Max posterior \cite{Hendrycks2017} &
$37.7^{\pm0.4}$ &
${72.7}^{\pm 0.6}$ & ${70.0}^{\pm 0.9}$ & $46.9^{\pm 0.4}$ &
$\textbf{50.0}^{\pm 0.0}$ & $52.0^{\pm 0.2}$ &
${58.3}^{\pm 0.2}$ \\

Ours: Error (\ref{equ:anomaly_simple}) &
$35.2^{\pm0.3}$ &
$61.9^{\pm 2.3}$ & $64.2^{\pm 1.8}$ & $51.1^{\pm 1.3}$ &
$\textbf{50.0}^{\pm 0.0}$ & $\textbf{52.9}^{\pm 0.7}$ &
$56.0^{\pm 1.0}$ \\

Ours: Reg. (\ref{equ:anomaly_score_pred_e}) &
$35.2^{\pm0.3}$ &
$47.0^{\pm 0.5}$ & $50.7^{\pm 1.6}$ & ${52.9}^{\pm 0.3}$ &
$\textbf{50.0}^{\pm 0.0}$ & $49.7^{\pm 0.2}$ &
$50.1^{\pm 0.4}$ \\

Ours: Quantile (\ref{equ:anomaly_score_reg}) &
$35.2^{\pm0.3}$ &
$53.9^{\pm 2.3}$ & $60.4^{\pm 2.3}$ &
$\textbf{56.3}^{\pm 0.9}$ & $\textbf{50.0}^{\pm 0.0}$ &
${52.5}^{\pm 0.5}$ &
$54.6^{\pm 1.1}$ \\

\bottomrule[.9pt]
\end{tabular}
}
\end{table}


\subsubsection{Object Detection\normalfont{:}}



\begin{table}[t!]
\centering
\caption{\textbf{Perturbation detection for object detection} on $\mathcal{D}^\mathrm{COCO}_\mathrm{val}$ using \texttt{Faster R-CNN}. Global AuROC (best in bold) per perturbation (Untargeted, Fabrication, Vanishing, Mislabeling, Gaussian, S\&P) across all perturbation strengths $\epsilon\!\in\!\{0.25, 0.5,1,2,4,8,16\}$ and across all perturbations and strengths in \% and AP in \%.}
\label{tab:anomaly_score_func_coco_od}
\setlength{\tabcolsep}{0.1pt}
\resizebox{1\linewidth}{!}{
\begin{tabular}{l|l|llllll|l}
\toprule[.9pt]
\multirow{2}{*}{Method} &
\makecell[c]{\multirow{2}{*}{\begin{tabular}{c}AP\\clean\end{tabular}}} &
\multicolumn{6}{c|}{Global AuROC (\%)} &
\multirow{2}{*}{\begin{tabular}{c}Global\\AuROC\end{tabular}} \\

& & \makecell[c]{Untar.} & \makecell[c]{Fabric.} & \makecell[c]{Vanish.} & \makecell[c]{Mislab.} & \makecell[c]{Gauss} & \makecell[c]{S\&P} & \\
\midrule
Entropy \cite{Smith2018} 
&$37.2^{\pm0.3}$
&$84.1^{\pm 0.7}$ &
$87.9^{\pm 0.6}$ &
$14.7^{\pm 0.2}$ &
$84.1^{\pm 0.6}$ &
$50.8^{\pm 0.2}$ &
$46.4^{\pm 1.1}$ &
$60.7^{\pm 0.2}$ \\

Max posterior \cite{Hendrycks2017}
&$37.2^{\pm0.3}$
&$\mathbf{90.9}^{\pm 0.3}$ &
$\mathbf{93.4}^{\pm 0.2}$ &
$11.4^{\pm 0.2}$ &
$\mathbf{91.0}^{\pm 0.3}$ &
$49.5^{\pm 0.0}$ &
$44.0^{\pm 0.8}$ &
$62.5^{\pm 0.2}$ \\

Ours: Error (\ref{equ:anomaly_simple})
&$34.9^{\pm0.1}$
&$86.6^{\pm 0.4}$ &
$88.6^{\pm 0.5}$ &
$15.1^{\pm 0.5}$ &
$86.6^{\pm 0.6}$ &
$\mathbf{51.8}^{\pm 0.7}$ &
$47.8^{\pm 2.2}$ &
$62.1^{\pm 0.6}$ \\

Ours: Reg. (\ref{equ:anomaly_score_pred_e})
&$34.9^{\pm0.1}$
&$72.6^{\pm 3.0}$ &
$75.1^{\pm 2.9}$ &
$\mathbf{70.5}^{\pm 1.6}$ &
$72.6^{\pm 2.8}$ &
$50.0^{\pm 0.6}$ &
$\mathbf{49.1}^{\pm 1.2}$ &
$\mathbf{64.2}^{\pm 1.6}$ \\

Ours: Quantile (\ref{equ:anomaly_score_reg})
&$34.9^{\pm0.1}$
&$67.6^{\pm 12}$ &
$68.1^{\pm 14}$ &
$27.6^{\pm 5.6}$ &
$67.8^{\pm 12}$ &
$51.7^{\pm 3.6}$ &
$49.3^{\pm 7.5}$ &
$55.1^{\pm 5.5}$  \\
\bottomrule[.9pt]
\end{tabular}
}
\end{table}

In \cref{tab:anomaly_score_func_coco_od}, we compare the perturbation detection performance of all three variants of our proposed methods in \cref{subsec:attack_detection} with state-of-the-art single-pass detection approaches for object detection across multiple perturbation types and strengths $\epsilon$. All methods employ a \texttt{Faster R-CNN} \cite{Ren2015} model trained on $\mathcal{D}^\mathrm{COCO}_\mathrm{train}$ and evaluated on perturbed samples from $\mathcal{D}^\mathrm{COCO}_\mathrm{val}$. We report the perturbation detection accuracy measured by global AuROC (\%) per perturbation across all strengths and across all perturbations and strengths and clean AP (\%). On clean data, we reach close-by AP performance on $\mathcal{D}^\mathrm{COCO}_\mathrm{val}$ with Hadamard output encodings. For perturbation detection, we observe that the max posterior method \cite{Hendrycks2017} is the best among the baselines, however showing imbalanced performance, being overall best on the untargeted, fabrication, and  mislabeling attack, but overall worst on the vanishing attack and salt and pepper noise. Again, our quantile method (\ref{equ:anomaly_score_reg}) performs poorly and unstable due to the small number of objects compared to the high number of pixels in semantic segmentation. Our regression detector (\ref{equ:anomaly_score_pred_e}) shows the best overall performance, excelling all methods on global AuROC (64.2\%) in addition to the vanishing attack and salt and pepper noise.



\section{Object Detection Loss Weight Ablation}
\label{supp:ObjectDetectionAblation}

\begin{table}[t!]
  \centering
  \caption{\textbf{Ablation study on classification loss weight for object detection.} We report AP (\%) for different $\lambda^{\mathrm{cls}}$ (\ref{equ:object_detection_loss}) of \texttt{HadamardNet} using \texttt{DETR} or \texttt{Faster R-CNN} on $\mathcal{D}^\mathrm{VOC}_\mathrm{val}$. Best results in bold.}
  \label{tab:ObjectDetectionAblation}
    
\extrarowheight=\aboverulesep
    \addtolength{\extrarowheight}{\belowrulesep}
    \aboverulesep=0pt
    \belowrulesep=0pt
    {
    \setlength{\tabcolsep}{3pt}
    \begin{tabular}{lccccccccccc}
    \toprule[.9pt]
    $\lambda^{\mathrm{cls}}$ &0.3 &0.4 &0.5 &0.6 &1.0 &3.0 &4.0 &8.0 &12.0 &16.0 &20.0 \\
    \midrule
    {\texttt{DETR}~\cite{Carion2020}} &- &- &- &- &- &$12.2$ &$24.3$ &$44.3$ &$46.1$ &$\mathbf{46.9}$ &$46.4$ \\
    \midrule
    {\texttt{Faster R-CNN}~\cite{Ren2015}} &$40.0$ &$40.2$ &$\mathbf{40.3}$ &$40.2$ &$39.6$ &$38.5$ &$38.2$ &- &- &- &- \\
    \bottomrule[.9pt]
\end{tabular}
}

\end{table}

In \cref{tab:ObjectDetectionAblation}, we present an ablation study on the weight $\lambda^{\mathrm{cls}}$ of the classification loss $J^{\mathrm{cls}}$ in the object detection loss $J^{\mathrm{OD}}$ (\ref{equ:object_detection_loss}). We fix the weights of the bounding box regression loss $J^{\mathrm{reg}}$ to the default hyperparameter values used in the \texttt{mmdetection} framework \cite{mmdetection}. We conduct an ablation for \texttt{DETR} and \texttt{Faster R-CNN} trained on $\mathcal{D}^\mathrm{VOC}_{\mathrm{train}}$ and evaluated on $\mathcal{D}^\mathrm{VOC}_\mathrm{val}$. We observe that, due to different architectures, the optimum for the weight of the classification loss $J^{\mathrm{cls}}$ (\ref{equ:overall_loss}) differs between \texttt{Faster R-CNN} and \texttt{DETR}. For \texttt{DETR}, the optimal loss weight is higher ($\lambda^{\mathrm{cls}}=16.0$) in comparison to the one-hot setup ($\lambda^{\mathrm{cls}}=1.0$ \cite{Carion2020}), while for \texttt{Faster R-CNN} it is lower ($\lambda^{\mathrm{cls}}=0.5$) in comparison to the one-hot setup ($\lambda^{\mathrm{cls}}=1.0$ \cite{Ren2015}). For \texttt{DETR}, we choose $\lambda^{\mathrm{cls}}=16.0$, for \texttt{Faster R-CNN} we choose $\lambda^{\mathrm{cls}}=0.5$ for all object detection experiments.

\section{Detailed Perturbation Detection Results on PascalVOC for Object Detection}
\label{supp:ObjectDetectionRobustness}

\begin{figure}[t]
    \centering
    \begin{tikzpicture}

\begin{axis}[
    width=9cm,
    height=5cm,
    xlabel={$\epsilon$},
    xlabel style={yshift=1mm}, 
    ylabel={AP ratio $Q^{\mathrm{AP}}$ (\%)},
    ylabel style={yshift=-0mm},
    xmode=log,
    log basis x=2,
    xtick={0.25,0.5,1,2,4,8,16},
    xticklabels={0.25,0.5,1,2,4,8,16},
    ymin=0, ymax=108,
    ytick={0,20,40,60,80,100},
    grid=both,
    clip=false,
    legend pos= outer north east,
    legend cell align={left}
]

\addplot[tu5,thick,mark=*] coordinates {
    (0.25,73.6) (0.5,50.0) (1,28.6) (2,14.8) (4,7.6) (8,4.7) (16,3.0)
};
\addlegendentry{Untargeted}

\addplot[tu8!70!black,thick,mark=*] coordinates {
    (0.25,70.4) (0.5,46.6) (1,26.4) (2,13.8) (4,7.1) (8,4.2) (16,2.5)
};
\addlegendentry{Fabrication}

\addplot[tu3,thick,mark=*] coordinates {
    (0.25,66.3) (0.5,38.9) (1,17.7) (2,8.4) (4,4.4) (8,2.7) (16,1.5)
};
\addlegendentry{Vanishing}

\addplot[tu6,thick,mark=*] coordinates {
    (0.25,65.3) (0.5,35.2) (1,11.3) (2,2.7) (4,1.0) (8,0.5) (16,0.2)
};
\addlegendentry{Mislabeling}

\addplot[tu2,thick,mark=*] coordinates {
    (0.25,99.8) (0.5,100.0) (1,99.8) (2,98.8) (4,95.6) (8,87.4) (16,70.9)
};
\addlegendentry{Gaussian noise}

\addplot[tu120,thick,mark=*] coordinates {
    (0.25,100.0) (0.5,99.8) (1,99.3) (2,97.5) (4,85.5) (8,58.9) (16,24.4)
};
\addlegendentry{S\&P noise}

\end{axis}
\end{tikzpicture}
    \setlength{\abovecaptionskip}{0pt} 
    \caption{\textbf{Performance degradation under perturbations} in terms of AP ratios $Q^{\mathrm{AP}}$. All results are obtained using a \texttt{Faster R-CNN} model~\cite{Ren2015} on perturbed images from $\mathcal{D}^{\mathrm{VOC}}_{\mathrm{val}}$.}
\label{fig:OD_robustness}
\end{figure}
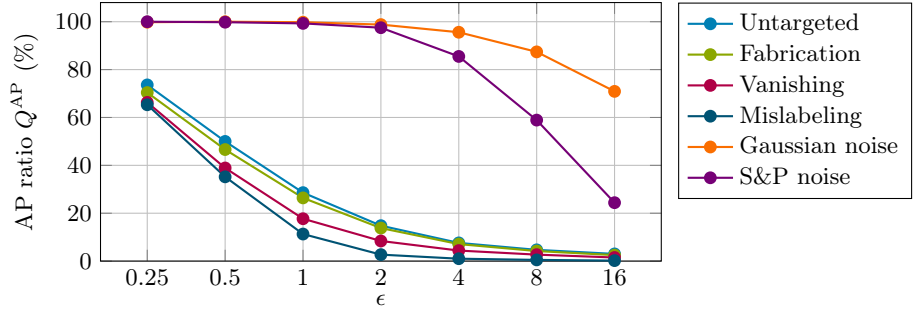

\begin{table}[t!]
  \centering
  \caption{\textbf{Perturbation detection by regression (\ref{equ:anomaly_score_pred_e}) for object detection} on $\mathcal{D}^\mathrm{VOC}_\mathrm{val}$. We report detection performance across perturbation types and strengths $\epsilon \in \{0.25, 0.5, 1,2,4,8,16\}$ using \texttt{Faster R-CNN}. Results are given as AuROC (\%) per $\epsilon$, global AuROC (\%) and global TPR$_{5\%}$ $(\%)$ across all $\epsilon$.}
  \label{tab:perturbations_det}

\renewcommand{\arraystretch}{0.8}
\setlength{\tabcolsep}{1pt}

\begin{tabular}{@{}c|r|r|r|r|r|r|r|r|r|r@{}}
\toprule[.9pt]
\multicolumn{1}{c|}{} &
\multicolumn{1}{l|}{Perturbation} &
\multicolumn{7}{c|}{AuROC (\%)} &
\makecell[c]{Global} &
\makecell[c]{Global} \\
\multicolumn{1}{c|}{} &
&
{$\epsilon=0.25$} & {$\epsilon=0.5$} &{$\epsilon=1$} & {$\epsilon=2$} & {$\epsilon=4$} & {$\epsilon=8$} & {$\epsilon=16$} &
{AuROC} &
{TPR$_{5\%}$} \\
\midrule
\multirow{6}{*}{\rotatebox{90}{Pascal VOC}} &
Untargeted
& $76.1$ & $89.9$ & $97.7$ & $99.7$ & $100.0$ & $100.0$ & $100.0$
& $93.9$ & $82.0$ \\

& Fabrication
& $78.6$ & $91.8$ & $98.4$ & $99.8$ & $100.0$ & $100.0$ & $100.0$
& $94.7$ & $83.9$ \\

& Vanishing
& $53.8$ & $63.0$ & $72.9$ & $78.4$ & $79.9$ & $79.4$ & $76.4$
& $70.6$ & $1.7$ \\

& Mislabeling
& $75.8$ & $89.7$ & $97.7$ & $99.7$ & $99.9$ & $100.0$ & $100.0$
& $93.8$ & $81.9$ \\

& Gauss.\ Noise
& $50.0$ & $50.0$ & $50.0$ & $50.1$ & $50.1$ & $50.2$ & $51.1$
& $50.2$ & $5.2$ \\

& S\&P Noise
& $50.0$ & $50.0$ & $50.0$ & $49.9$ & $53.1$ & $53.7$ & $51.9$
& $51.2$ & $5.3$ \\

\bottomrule[.9pt]
\end{tabular}

\end{table}

In \cref{fig:OD_robustness}, we show the performance of the \texttt{Faster R-CNN}~\cite{Ren2015} model with Hadamard output encodings at different perturbations and strengths $\epsilon$, reported as the AP ratio $Q^{\mathrm{AP}} = \frac{\mathrm{AP}^{\mathrm{perturbed}}}{\mathrm{AP}^{\mathrm{clean}}}$ between perturbed and clean performance. The models are trained on $\mathcal{D}^\mathrm{VOC}_\mathrm{train}$ and evaluated on perturbed images from $\mathcal{D}^\mathrm{VOC}_\mathrm{val}$. We observe that all adversarial attacks, despite their different attack objectives, lead to similar performance drops ($Q^{\mathrm{AP}}$) across perturbation strengths. Even for small attack strengths $\epsilon \le 2$, the attacks substantially degrade the object detection performance, such that for $\epsilon \ge 4$ the object detection performance ($Q^{\mathrm{AP}}$) saturates near $0\%$ for all adversarial attacks. In contrast, for disturbances, the object detection performance does not degrade at small strengths $\epsilon \le 2$. For higher strengths $\epsilon \ge 4$, both disturbances cause performance degradation, with salt and pepper noise having the stronger effect.

In \cref{tab:perturbations_det}, we report the performance of our regression approach (\ref{equ:anomaly_score_pred_e}) for all perturbations and strengths $\epsilon\in \{0.25, 0.5,1,2,4,8,16\}$ using a \texttt{Faster R-CNN} \cite{Carion2020} trained on $\mathcal{D}^\mathrm{VOC}_\mathrm{train}$ and evaluated on perturbed samples from $\mathcal{D}^\mathrm{VOC}_\mathrm{val}$. We report the perturbation detection accuracy measured by AuROC (\%), global AuROC (\%), and global TPR$_{5\%}$ $(\%)$ per perturbation across all strengths. We observe that disturbances are difficult to detect showing AuROC values close to 50\%, particularly for low perturbation strengths $\epsilon \le 2$, where they also not degrade object detection performance. Further, we observe that for the untargeted, fabrication, and mislabeling attacks, the detection performance increases with increasing strength $\epsilon$, saturating at $100 \%$ AuROC for $\epsilon \ge 4$, consistent with the saturation of $Q^{\mathrm{AP}}$ in \cref{fig:OD_robustness}. In case of the vanishing attack, the detection performance decreases slightly for strengths $\epsilon >4$, as objects are more confidently predicted as background which naturally weakens object-based detectors. Despite a reasonably high global AuROC (70.6\%) for the vanishing attack, the TPR$_{5\%}$ remains low (1.7\%), due to a delayed increase in TPR, as detailed in Supplement \ref{supp:ROC_OD}.

\section{Additional ROC Curves for Object Detection}
\label{supp:ROC_OD}

\begin{figure}[t!]
    \centering
    \begin{tikzpicture}
\begin{groupplot}[
    group style={
        group size=2 by 2,
        horizontal sep=0.8cm,
        vertical sep=1.4cm
    },
    width=0.48\textwidth,
    height=4.3cm,
    xmin=0, xmax=1,
    ymin=0, ymax=1,
    ylabel={TPR},
    ylabel style={yshift=-0mm},
    grid=both,
    legend image post style={xscale=0.7},
    legend cell align=left,
    legend style={font=\small,row sep=-2pt,
    inner xsep=3pt,
    inner ysep=1pt,}
]

\nextgroupplot[
    title={Mislabeling \cite{Chow2020}},
    legend style={at={(0.98,0.39)}, anchor=south east}
]

\addplot[thick,tu10]
table[
  col sep=comma,
  x=FPR,
  y=TPR,
  each nth point=10,
] {figures/data/mislabeling/nonlinear_eps0_25_roc_data.csv};

\addplot[thick,tu10,dashed,forget plot]
table[
  col sep=comma,
  x=FPR,
  y=TPR,
  each nth point=10,
] {figures/data/mislabeling/maxposterior_eps0_25_roc_data.csv};

\addplot[thick,tu5]
table[
  col sep=comma,
  x=FPR,
  y=TPR,
  each nth point=10,
] {figures/data/mislabeling/nonlinear_eps0_5_roc_data.csv};

\addplot[thick,tu5,dashed,forget plot]
table[
  col sep=comma,
  x=FPR,
  y=TPR,
  each nth point=10,
] {figures/data/mislabeling/maxposterior_eps0_5_roc_data.csv};

\addplot[thick,tu3]
table[
  col sep=comma,
  x=FPR,
  y=TPR,
  each nth point=10,
] {figures/data/mislabeling/nonlinear_eps1_roc_data.csv};

\addplot[thick,tu3,dashed,forget plot]
table[
  col sep=comma,
  x=FPR,
  y=TPR,
  each nth point=10,
] {figures/data/mislabeling/maxposterior_eps1_roc_data.csv};

\addlegendentry{$\epsilon = 0.25$}
\addlegendentry{$\epsilon = 0.5$}
\addlegendentry{$\epsilon = 1$}

\nextgroupplot[
    title={Vanishing \cite{Chow2020}},
    ylabel={}
]

\addplot[thick,tu10]
table[
  col sep=comma,
  x=FPR,
  y=TPR,
  each nth point=10,
] {figures/data/vanishing/nonlinear_eps0_25_roc_data.csv};

\addplot[thick,tu10,dashed,forget plot]
table[
  col sep=comma,
  x=FPR,
  y=TPR,
  each nth point=10,
] {figures/data/vanishing/maxposterior_eps0_25_roc_data.csv};

\addplot[thick,tu5]
table[
  col sep=comma,
  x=FPR,
  y=TPR,
  each nth point=10,
] {figures/data/vanishing/nonlinear_eps0_5_roc_data.csv};

\addplot[thick,tu5,dashed,forget plot]
table[
  col sep=comma,
  x=FPR,
  y=TPR,
  each nth point=10,
] {figures/data/vanishing/maxposterior_eps0_5_roc_data.csv};

\addplot[thick,tu3]
table[
  col sep=comma,
  x=FPR,
  y=TPR,
  each nth point=10,
] {figures/data/vanishing/nonlinear_eps1_roc_data.csv};

\addplot[thick,tu3,dashed,forget plot]
table[
  col sep=comma,
  x=FPR,
  y=TPR,
  each nth point=10,
] {figures/data/vanishing/maxposterior_eps1_roc_data.csv};

\nextgroupplot[
    title={Gaussian noise},
    ylabel={TPR},
    xlabel={FPR},
    legend style={at={(axis description cs:0.98,0.02)}, anchor=south east}
]

\addplot[thick,tu9]
table[
  col sep=comma,
  x=FPR,
  y=TPR,
  each nth point=10,
] {figures/data/gaussian/nonlinear_eps2_roc_data.csv};

\addplot[thick,tu9,dashed,forget plot]
table[
  col sep=comma,
  x=FPR,
  y=TPR,
  each nth point=10,
] {figures/data/gaussian/maxposterior_eps2_roc_data.csv};

\addplot[thick,tu6]
table[
  col sep=comma,
  x=FPR,
  y=TPR,
  each nth point=10,
] {figures/data/gaussian/nonlinear_eps4_roc_data.csv};

\addplot[thick,tu6,dashed,forget plot]
table[
  col sep=comma,
  x=FPR,
  y=TPR,
  each nth point=10,
] {figures/data/gaussian/maxposterior_eps4_roc_data.csv};

\addplot[thick,tu2]
table[
  col sep=comma,
  x=FPR,
  y=TPR,
  each nth point=10,
] {figures/data/gaussian/nonlinear_eps8_roc_data.csv};

\addplot[thick,tu2,dashed,forget plot]
table[
  col sep=comma,
  x=FPR,
  y=TPR,
  each nth point=10,
] {figures/data/gaussian/maxposterior_eps8_roc_data.csv};

\addlegendentry{$\epsilon = 2$}
\addlegendentry{$\epsilon = 4$}
\addlegendentry{$\epsilon = 8$}

\nextgroupplot[
    title={Salt and pepper noise},
    ylabel={},
    xlabel={FPR}
]

\addplot[thick,tu9]
table[
  col sep=comma,
  x=FPR,
  y=TPR,
  each nth point=10,
] {figures/data/sp/nonlinear_eps2_roc_data.csv};

\addplot[thick,tu9,dashed,forget plot]
table[
  col sep=comma,
  x=FPR,
  y=TPR,
  each nth point=10,
] {figures/data/sp/maxposterior_eps2_roc_data.csv};

\addplot[thick,tu6]
table[
  col sep=comma,
  x=FPR,
  y=TPR,
  each nth point=10,
] {figures/data/sp/nonlinear_eps4_roc_data.csv};

\addplot[thick,tu6,dashed,forget plot]
table[
  col sep=comma,
  x=FPR,
  y=TPR,
  each nth point=10,
] {figures/data/sp/maxposterior_eps4_roc_data.csv};

\addplot[thick,tu2]
table[
  col sep=comma,
  x=FPR,
  y=TPR,
  each nth point=10,
] {figures/data/sp/nonlinear_eps8_roc_data.csv};

\addplot[thick,tu2,dashed,forget plot]
table[
  col sep=comma,
  x=FPR,
  y=TPR,
  each nth point=10,
] {figures/data/sp/maxposterior_eps8_roc_data.csv};

\end{groupplot}

\begin{axis}[
    hide axis,
    scale only axis,
    width=0pt,
    height=0pt,
    at={(0,0)},
    anchor=south west,
    xmin=0, xmax=1,
    ymin=0, ymax=1,
    legend to name=methodlegend,
    legend style={draw=black,fill=white,font=\small,row sep=-2pt}
]
\addlegendimage{black,thick}
\addlegendentry{Ours (\ref{equ:anomaly_score_pred_e})}
\addlegendimage{black,thick,dashed}
\addlegendentry{Max posterior~\cite{Hendrycks2017}}
\end{axis}

\node at (fgsmLegendPos) 
[anchor=south east, xshift=4.5pt, yshift=-4pt]
{\pgfplotslegendfromname{methodlegend}};

\end{tikzpicture}
    \caption{\textbf{ROC of our proposed regression perturbation detection (\ref{equ:anomaly_score_pred_e}) for object detection}, in the top-segment mislabeling~\cite{Chow2020} attack (left), vanishing~\cite{Chow2020} attack (right) at specific perturbation strengths $\epsilon \in \{0.25, 0.5, 1\}$ and in the bottom-segment gaussian noise (left), salt and pepper noise (right) at specific perturbation strengths $\epsilon \in \{2,4,8\}$. All results are obtained using a \texttt{Faster R-CNN} \cite{Ren2015} model trained on $\mathcal{D}^\mathrm{VOC}_\mathrm{train}$ and evaluated on perturbed images from $\mathcal{D}^\mathrm{VOC}_\mathrm{val}$.}
\label{fig:ROC_dec_supp}
\end{figure}
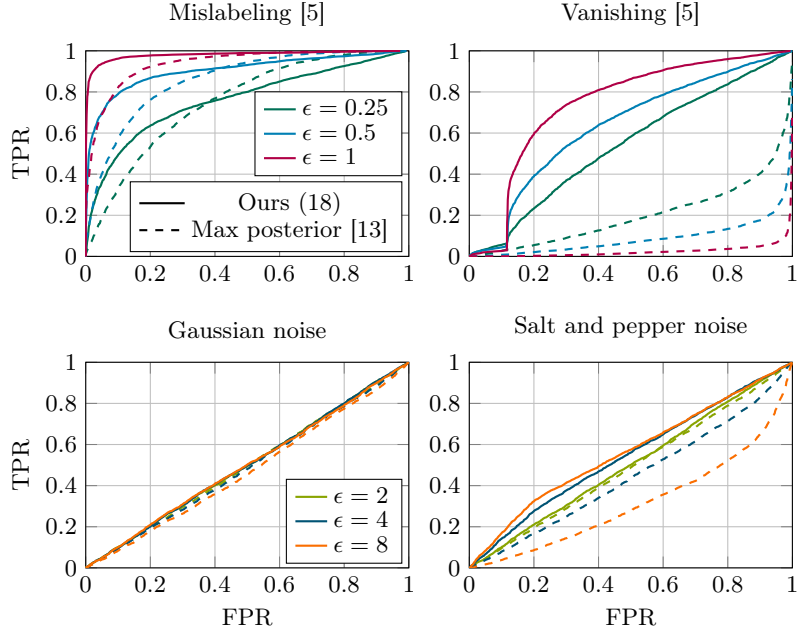

In \cref{fig:ROC_dec_supp}, we display ROC curves for the mislabeling~\cite{Chow2020} attack (top, left), vanishing~\cite{Chow2020} attack (top, right), gaussian noise (bottom, left), and salt and pepper noise (bottom, right) for different perturbation strengths~$\epsilon$ for our regression approach (\ref{equ:anomaly_score_pred_e}) and the (strong) max posterior baseline \cite{Hendrycks2017}. The results are obtained using the \texttt{Faster R-CNN}~\cite{Ren2015} model, which is trained on $\mathcal{D}^\mathrm{VOC}_\mathrm{train}$ and evaluated on perturbed images from $\mathcal{D}^\mathrm{VOC}_\mathrm{val}$. The ROC curves are obtained by varying the decision threshold $\theta$ applied to the anomaly score $\delta(\mathbf{x})$ \ref{equ:anomaly_score_pred_e}.
Again, we observe that Gaussian noise is difficult to detect for both our regression method and the strongest baseline max posterior \cite{Hendrycks2017}, with the ROC curves lying nearly on the diagonal. For the max posterior baseline \cite{Hendrycks2017}, the ROC curves deteriorate with increasing perturbation strength for the vanishing attack and salt and pepper noise. In principle, one could invert the decision rule of the max posterior baseline to move the ROC curves above the diagonal, however, this would simultaneously invert the decision for all other perturbations, thereby drastically reducing the overall performance. In contrast, for our regression method (\ref{equ:anomaly_score_pred_e}), the ROC curves improve with increasing perturbation strength for the mislabeling attack, vanishing attack, and salt and pepper noise, delivering better ROC curves for reasonably small FPRs than the max posterior baseline \cite{Hendrycks2017}. In case of the vanishing attack, we further observe a delayed increase in TPR at around 10\% FPR in contrast to all other adversarial attacks.

\section{Detailed Perturbation Detection Results on BDD100K for Object Detection}
\label{supp:additional_results_bdd_od}

\begin{figure}[t!]
    \centering
    \begin{tikzpicture}

\begin{axis}[
    width=9cm,
    height=5cm,
    xlabel={$\epsilon$},
    xlabel style={yshift=1mm}, 
    ylabel={AP ratio $Q^{\mathrm{AP}}$ (\%)},
    ylabel style={yshift=-0mm},
    xmode=log,
    log basis x=2,
    xtick={0.25,0.5,1,2,4,8,16},
    xticklabels={0.25,0.5,1,2,4,8,16},
    ymin=0, ymax=108,
    ytick={0,20,40,60,80,100},
    grid=both,
    clip=false,
    legend pos= outer north east,
    legend cell align={left}
]

\addplot[tu5,thick,mark=*] coordinates {
    (0.25,83.9) (0.5,66.1) (1,45.0) (2,28.6) (4,19.6) (8,13.4) (16,8.4)
};
\addlegendentry{Untargeted}

\addplot[tu8!70!black,thick,mark=*] coordinates {
    (0.25,87.0) (0.5,73.6) (1,55.9) (2,40.4) (4,30.1) (8,23.6) (16,17.4)
};
\addlegendentry{Fabrication}

\addplot[tu3,thick,mark=*] coordinates {
    (0.25,85.1) (0.5,68.3) (1,46.0) (2,25.2) (4,11.8) (8,5.3) (16,2.5)
};
\addlegendentry{Vanishing}

\addplot[tu6,thick,mark=*] coordinates {
    (0.25,84.2) (0.5,64.9) (1,45.3) (2,29.8) (4,18.0) (8,9.9) (16,5.3)
};
\addlegendentry{Mislabeling}

\addplot[tu2,thick,mark=*] coordinates {
    (0.25,100.0) (0.5,100.0) (1,100.3) (2,99.7) (4,97.5) (8,91.3) (16,74.2)
};
\addlegendentry{Gaussian noise}

\addplot[tu120,thick,mark=*] coordinates {
    (0.25,100.0) (0.5,100.0) (1,100.3) (2,99.7) (4,97.2) (8,87.6) (16,63.0)
};
\addlegendentry{S\&P noise}

\end{axis}
\end{tikzpicture}
    \setlength{\abovecaptionskip}{0pt} 
    \caption{\textbf{Performance degradation under perturbations} in terms of AP ratios $Q^{\mathrm{AP}}$. All results are obtained using a \texttt{Faster R-CNN} model~\cite{Ren2015} on perturbed images from $\mathcal{D}^{\mathrm{BDD,OD}}_{\mathrm{test}}$.}
\label{fig:OD_robustness_bdd}
\end{figure}
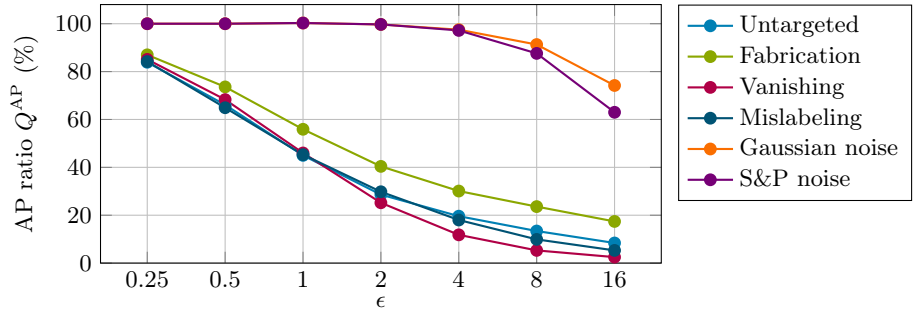

In \cref{fig:OD_robustness_bdd}, we show the performance of the \texttt{Faster R-CNN}~\cite{Ren2015} model with Hadamard output encodings at different perturbations and strengths $\epsilon$ reported as the AP ratio $Q^{\mathrm{AP}}$. The models are trained on $\mathcal{D}^\mathrm{BDD,OD}_\mathrm{train}$ and evaluated on perturbed images from $\mathcal{D}^\mathrm{BDD,OD}_\mathrm{test}$. Again, all adversarial attacks lead to similar performance drops ($Q^{\mathrm{AP}}$) consistently reducing the object detection performance with increasing strength $\epsilon$, with the fabrication attack causing slightly weaker degradation. In contrast, disturbances do not affect the object detection performance until $\epsilon=4$, beyond this point, both disturbances lead to a slight performance degradation.

\begin{table}[t!]
  \centering
  \caption{\textbf{Perturbation detection by regression (\ref{equ:anomaly_score_pred_e}) for object detection} on $\mathcal{D}^\mathrm{BDD,OD}_\mathrm{test}$. We report detection performance across perturbation types and strengths $\epsilon \in \{0.25, 0.5, 1,2,4,8,16\}$ using \texttt{Faster R-CNN}. Results are given as AuROC (\%) per $\epsilon$, global AuROC (\%) and global TPR$_{5\%}$ $(\%)$ across all $\epsilon$.}
  \label{tab:detection_comparison_od_bdd}

\renewcommand{\arraystretch}{0.8}
\setlength{\tabcolsep}{1pt}

\begin{tabular}{@{}c|l|r|r|r|r|r|r|r|r|r@{}}
\toprule[.9pt]
\multicolumn{1}{c|}{} &
\multicolumn{1}{l|}{Perturbation} &
\multicolumn{7}{c|}{AuROC (\%)} &
\makecell[c]{Global} &
\makecell[c]{Global} \\
\multicolumn{1}{c|}{} &
&
{$\epsilon=0.25$} & {$\epsilon=0.5$} &{$\epsilon=1$} & {$\epsilon=2$} & {$\epsilon=4$} & {$\epsilon=8$} & {$\epsilon=16$} &
{AuROC} &
{TPR$_{5\%}$} \\
\midrule
\multirow{6}{*}{\rotatebox{90}{BDD100K}} &
Untargeted
& $51.2$ & $61.0$ & $76.9$ & $91.1$ & $96.0$ & $98.2$ & $98.4$
& $78.6$ & $53.7$ \\

& Fabrication
& $55.4$ & $67.7$ & $83.7$ & $94.2$ & $97.6$ & $98.7$ & $98.8$
& $83.0$ & $62.3$ \\

& Vanishing
& $51.1$ & $51.6$ & $52.8$ & $55.1$ & $59.0$ & $64.5$ & $69.6$
& $57.1$ & $1.9$ \\

& Mislabeling
& $49.7$ & $57.3$ & $72.3$ & $86.2$ & $93.9$ & $96.9$ & $97.4$
& $75.4$ & $47.6$ \\

& Gauss.\ Noise
& $50.0$ & $50.0$ & $50.1$ & $50.1$ & $50.3$ & $50.3$ & $49.9$
& $50.1$ & $4.7$ \\

& S\&P Noise
& $50.0$ & $50.0$ & $50.1$ & $50.3$ & $50.2$ & $50.7$ & $53.8$
& $50.7$ & $5.1$ \\

\bottomrule[.9pt]
\end{tabular}

\end{table}

In \cref{tab:detection_comparison_od_bdd}, we report the performance of our regression approach (\ref{equ:anomaly_score_pred_e}) for all perturbations and strengths $\epsilon\in \{0.25, 0.5, 1,2,4,8,16\}$ using a \texttt{Faster R-CNN} \cite{Carion2020} trained on $\mathcal{D}^\mathrm{BDD,OD}_\mathrm{train}$ and evaluated on perturbed samples from $\mathcal{D}^\mathrm{BDD,OD}_\mathrm{test}$. We report the perturbation detection accuracy measured by AuROC (\%) and global AuROC (\%) and global TPR$_{5\%}$ $(\%)$ per perturbation across all strengths. We observe that for all adversarial attacks the detection performance increases with increasing strength $\epsilon$. As before, we observe a low global TPR$_{5\%}$ on the vanishing attack, due to the delayed increase in TPR as described in Supplement \ref{supp:ROC_OD}. We again observe that disturbances are difficult to detect with AuROC values close to 50\%, particularly for low perturbation strengths $\epsilon \le 4$, where they do not degrade object detection performance.

\end{document}